\documentclass[sigconf,nonacm]{acmart} 
\usepackage{graphicx}
\usepackage{subcaption}
\usepackage{dcolumn}
\usepackage{bm}
\usepackage{xcolor}
\usepackage{xspace}
\usepackage{xfrac}
\usepackage{color,soul}
\usepackage{hyperref} 

\usepackage{tikz}
\usetikzlibrary{trees, positioning}

\usepackage[linesnumbered,ruled,vlined]{algorithm2e} 

\renewcommand\footnotetextcopyrightpermission[1]{} 

\title{Self-Organizing Language}

\author{P.~Myles~Eugenio$^{1}$ \& Anthony~Beavers$^{2}$}
\affiliation{%
  $^{1}$\institution{Department of Physics, Indiana University}
  $^{2}$\institution{Cognitive Science Program, Indiana University}
  \city{Bloomington}\state{Indiana}\postcode{47405}\country{USA}
}
\email{pamyeuge@iu.edu}

\begin{document}
\begin{abstract}
We introduce a novel paradigm of emergent local memory. It is a continuous-learning completely-parallel content-addressable memory encoding global order. It demonstrates how local constraints on uncoordinated learning can produce topologically protected memories realizing emergent symbolic order. It is therefore a neuro-symbolic bridge.

It further has the ability to produce human language without data, by exploiting its own self-organizing dynamics. It teaches us that words arise as a side-effect of emergent symbolic order, and that human language patterns at all structural levels reflect a universal mechanism of word formation (which is subregular). This work answers essential questions about the existence \& origin of all the human language data.

\end{abstract}

\keywords{Exotic AI, Hebbian, Physics-inspired Neural Nets, Interpretable Models, Hierarchical Compression, Language Emergence, Neuro-symbolic, Neuromorphic, Topological Protected Memory}
\maketitle

\section{Introduction}

Language did not exist in the early universe, and yet it exists today. Thus learning language does not require language data. We demonstrate this here by constructing a model which converts noise into language. 

By {\it language}, we refer to human language, as in the language which children learn and are responsible for. Specifically, we pre-assume a primative token set (such as an alphabet), and work entirely in the regime of human language patterns.\footnote{The emergence of discrete categories from estimating continuous information has been studied by other authors \cite{Boersma2020_NeuralNets4Phonology}.} These patterns, while seemingly arbitrary, have {\it universal} statistical structure quantifiable in terms of the distribution of these patterns: hierarchical, with finite word length, and a scale-invariant sub-word structure related to its readabilitywithoutspaces. This coincides with a multi-scale {\it symbolic} structure, a limit of accepted \& well-formed sequences.

We thus define a valid {\it human language model} by its relationship to universals. A rigorous model must both (i) be grounded in physical reality, i.e local neural learning, and (ii) generate the universal structures of human language without pre-fitting them. Symbolic theories fail the first criterion: they describe high-level rules but are not realizations of local learning. Modern statistical language models (e.g LLM's) fail both criteria: they are non-local pattern recognizers which fit pre-existing data but cannot account for why the data exists. What's needed is a model which satisfies both.


The model we present satisfies these criteria. By {\it model}, we do not mean a single function, composed of other functions, whose parameters are optimized to minimize a global objective. Instead we refer to a set of {\it locally coordinated} rules, akin to a physical system, which continuously grows a model neuron-by-neuron. This places it in the same (local \& objective-free) paradigm of model as Conway's Game of Life \cite{PaulRendell2011_ConwayTuringComplete,PaulRendell2014thesis_ConwayTuringComplete} or Lenia \cite{Chan2019_Lenia}, except in the domain of language and neural learning.

Despite being a highly uncoordinated learner, the system can represent {\it global order}. By global order we mean not just simple patterns, like ferromagnetic or antiferromagnetic alignment, but order rich enough to encode arbitrary strings—for instance, spins arranging to spell out the names of all the US presidents in sequence. In this sense, language itself is global order. Such order is built locally, through a hierarchical memory we call the {\it retokenizer}. Even when trained on noise, the retokenizer self-organizes into a structure that encodes strings with universal sub-word distributions and finite word length. These strings act as symbols: terminal nodes of a bifix-free\footnote{Preffix \& suffix free, meaning that {\bf cat} \& {\bf catch} cannot both distinguished in the same instance of memory. Only the longer one is retrieved.} directed acyclic graph (DAG), whose topology guarantees their retrieval. In this way, symbolic behavior is not imposed but emerges from the topological character of the memory. 

It teaches us that {\bf words \& their universal sub-structure are a side-effect of emergent symbolic behavior}. Explaining the origin of meaningless sub-word patterns is only a means to discovering the model itself: a neuro-symbolic bridge (a type of meta-tokenizer). It converts statistical correlations, the products of tokens ($v_{\bf y}\otimes v_{\bf o}\otimes v_{\bf u}$), into symbols ({\it feature tokens}: $\tilde{v}^{(3)}_{\bf you}$,$\tilde{v}^{(2)}_{\bf yo}$,$\tilde{v}^{(2)}_{\bf ou}$) in a strictly local way\footnote{Neurons collectively perform a renormalization group flow \cite{Wilson1974_RG,Shankar1994_FermionRG,Shankar2017_RGbook} (i.e a staged estimation of patterns)--forming new symbols representing statistically relevant degrees of freedom at longer length scales. Their collective topology imbues each symbol with a redundancy: an emergent gauge structure.}. Its symbolic memories are exact, which allows retrieval to act as a further learning step. Introducing a new neuron during replay binds to the recalled features, producing an {\it object token} (a.k.a an {\it embedding}), e.g 
\[
a_{\bf you} = \tilde{v}^{(3)}_{\bf you} 
\oplus \big(\tilde{v}^{(2)}_{\bf yo}+\tilde{v}^{(2)}_{\bf ou}\big) 
\oplus \big(v_{\bf y}+v_{\bf o}+v_{\bf u}\big) . 
\]

The formation of object tokens marks the transition from short-term correlations to long-term symbolic memory. Each object token is completely independent, storing the feature tokens of the retokenizer (whose memory can be forgotten). Inference (next-token prediction) now follows a key-value relationship, an emergent attention mechanism with knowledge-graph behavior akin to transformer embeddings. Crucially, object tokens can themselves be correlated, producing higher-order object tokens. Thus the architecture is not fixed: new memories can be formed without erasing old ones, and the computational class itself evolves with continued learning. 

The complete model is not a singular thing, but a non-static continuous-learning completely-parallel content addressable memory encoding global order. Therefore, what we present is a novel paradigm of {\it emergent local memory}. Such a model is essential for understanding how the universal constraints of the brain are themselves responsible for the observed macroscopic language order. Thus it is also an {\it effective human language model}, and the only one we are aware of which meets the minimal criteria laid out in this intro.

In this paper, we focus on the low-data regime, which is the essential regime for answering key questions related to human memory. Conveniently, local models are analytically tractable \& thus highly interpretable. It's parallel nature makes it so that the pedagogical toy models we present are actually representative of how the model can learn in practice -- bit by bit. 

The combination of analytic simplicity and multi-scale symbolic structure makes it possible to construct model initial states with syntax-like inter-categorical relationships and hierarchy. As a demonstration, we provide free code\footnote{\url{https://gitlab.com/emla-group/ababot}} where such initial states can be specified directly, allowing symbolic theories of English (or other languages) to be hand-coded and then grown through local learning. \\

This manuscript is organized into two main parts: The first part encompasses the tensor machinery of short-term memory (STM), which builds to a discussion on the structural origins of human subword structure (Sec \ref{Sec:UniversalSubWordStructure}). The second half covers the mechanisms responsible for forming long-term memory (LTM)---starting first with a summary (Sec \ref{Sec:BenefitsOfThinking}), followed by mathematical details. We simplify the math \& discuss the high-level organization of the code in Sec \ref{Sec:Simpler&Code}. Then discuss Syntax in Sec \ref{Sec:Syntax}. We end with some toy examples in Sec \ref{Sec:Toys}.

\renewcommand{\contentsname}{Contents}
\tableofcontents

\section{Model Summary: Local Events and Hierarchical Retokenization}
\label{Sec: Model}

\subsection{Preliminary}

This framework is built on {\it tokens}, which are symbols representing the firing state of a neuron. The most basic are assumed external basis tokens ($v_{\bf a},v_{\bf b},\cdots,v_{\bf z}$), where bold font letters indicate labels ({\bf a},{\bf b},$\cdots$,{\bf z}) distinguishing those neurons.

We further define a token {\it pool}, which is an ordered list of tokens used as a basis, e.g $[v_{\bf a},v_{\bf b}]$. This allows us to construct a dense representation, while keeping track of which vector rows correspond to which tokens. For our dimension $d=2$ example pool:
\begin{eqnarray}
v_{\bf a} &=& \begin{pmatrix}1 & 0\end{pmatrix} \notag\\
v_{\bf b} &=& \begin{pmatrix}0 & 1\end{pmatrix} ,\notag
\end{eqnarray}
and use Latin indices ($j,k,l\in\{{\bf a},{\bf b}\}$) to write $v_k\in\{v_{\bf a},v_{\bf b}\}$. Tensor products of tokens (written $v_{\bf a}\otimes v_{\bf b}\equiv v_{\bf a}v_{\bf b}$) themselves form their own $d^2$ dimensional orthonormal set, 
\begin{eqnarray}
v_{\bf a}v_{\bf a} &=& \begin{pmatrix}1 & 0 & 0 & 0\end{pmatrix} \notag\\
v_{\bf a}v_{\bf b} &=& \begin{pmatrix}0 & 1 & 0 & 0\end{pmatrix} \notag\\
v_{\bf b}v_{\bf a} &=& \begin{pmatrix}0 & 0 & 1 & 0\end{pmatrix} \notag\\
v_{\bf b}v_{\bf b} &=& \begin{pmatrix}0 & 0 & 0 & 1\end{pmatrix} .\notag
\end{eqnarray}
One could even define $v_{\bf ab}^{(2)}\equiv v_{\bf a}v_{\bf b}$ to denote that they are higher dimensional vectors. Higher-order tensor products form the generic basis for representing tensors, e.g 
\begin{eqnarray}
T^{(n)} = \sum_{l,\cdots,j,k}T_{l,\cdots,j,k}^{(n)}v_{l}\otimes\cdots\otimes v_{j}\otimes v_{k} , 
\end{eqnarray}
where $T_{l,\cdots,j,k}^{(n)}$ is the matrix element. A generic tensor $T$ is a direct sum (concatenation) over all subspaces, 
\begin{eqnarray}
T = T_{(1)}\oplus T_{(2)}\oplus\cdots\oplus T_{(n)}\oplus \cdots, 
\end{eqnarray}
equivalently 
\begin{eqnarray}
 = \sum_n\sum_{l,\cdots,j,k}T_{l,\cdots,j,k}^{(n)}v_{l}\otimes\cdots\otimes v_{j}\otimes v_{k} . 
\end{eqnarray}

Note that {\it context} refers to a list of tokens, for either learning or autoregressively predicting subsequent tokens. A generic context $[v(1),v(2),\cdots,v(n)]$ can be mapped to a direct sum over tensor products
\begin{eqnarray}
&&T_{\text{context}} = \\
&&v(n)\oplus\big[v(n-1)v(n)\big]\oplus\cdots\oplus\big[v(1)v(2)\cdots v(n)\big] \notag
\end{eqnarray}
where subspaces beyond length $n$ are implicitly padded by zeros. Tensors act on context via a dot product, i.e a similarity check. For example, $T_{\text{context}}=v_{\bf a}\oplus v_{\bf b}v_{\bf a} \oplus v_{\bf a}v_{\bf b}v_{\bf a}$, reduces to the sum 
\begin{eqnarray}
T\cdot T_{\text{context}} = T_{\bf a}^{(1)}+T_{\bf ba}^{(2)} + T_{\bf aba}^{(3)} .
\end{eqnarray}

Note that the quantity $n$ is the coordination length. Tensor products of length $n$ require all $n$ tokens to be non-zero in order for the whole product to be finite---hence they can be understood as encoding conditional AND(\&) statements.

In general, the dimension of a tensor product grows exponentially in the number of products ($n$).\footnote{One can understand $T$ as living in a real-valued Fock space.} Note that we can reduce the memory overhead by truncating the exponentially-large tensor product space, defining a reduced basis set written $\tilde{v}_{\mu_n}^{(n)}$, where $d_n\equiv \text{dim}(\mu_n)\ll d^n$. For example ($d=2$), keeping {\bf ab} \& {\bf ba}: 
\begin{eqnarray}
\tilde{v}_{\bf ab}^{(2)} &=& \begin{pmatrix}1 & 0 \end{pmatrix} \notag\\
\tilde{v}_{\bf ba}^{(2)} &=& \begin{pmatrix}0 & 1\end{pmatrix} .\notag
\end{eqnarray}
We will show this effectuates a low-rank compression, 
\begin{eqnarray}
\text{compressed}\big(T^{(n)}\big) = \sum_{\mu_{n-1},k}T_{\mu_{n-1},k}^{(n)}\tilde{v}^{(n-1)}_{\mu_{n-1}}\otimes v_{k} ,
\end{eqnarray}
where the trailing token products are merged into a single compressed vector. This reduces the tensor product to effectively $2$-point coordination (i.e locally coordinated).

The sections which follow demonstrates how to grow low-rank tensors from local learning events, which directly map to $n$-points in the uncompressed $d^n$ space. \\

{\bf Throughout this paper we use standard tensor notation.} Bold font distinguish labels ({\bf a},{\bf b},{\bf c}) from free indices ($k,\mu_{n},\alpha$). Some indices ($\mu_n$) carry sub-indices ($n$) indicating layer.

\subsection{Local Learning Events}

The foundational object of our framework is a localized \emph{event}, defined as a pair
\[
\mathcal{E}^{(n)} = \big( \tilde{v}^{(n-1)}_{\mu_{n-1}},\ v_k \big) ,
\]
where \( v_k \) is the free token (a vector of dimension $d$) to be predicted, and \( \tilde{v}^{(n-1)}_{\mu_{n-1}} \) is the learned projection of the \( n{-}1 \) trailing context tokens. Each such event represents a local time-step in the dynamics of the model, and triggers a Hebbian memory update:
\begin{equation}\label{Eqn:HebbUpdate}
g^{(n)}_{\mu_{n-1}, k} \gets g^{(n)}_{\mu_{n-1}, k} + \tilde{v}^{(n-1)}_{\mu_{n-1}} v_k .
\end{equation}
This update defines a \emph{short-term memory synapse}, storing pairwise correlations between projected context features and the next token. Learning is entirely local and unsupervised: there is no global loss function, and no backpropagation.

\subsection{Hierarchical Retokenization}

To organize these events across scales, the model constructs a recursive sequence of projection maps \( P_n \) that define higher-order features. Each projection promotes a pair \( (\tilde{v}^{(n-1)}, v) \) to a single token \( \tilde{v}^{(n)} \), if it satisfies the condition:
\begin{equation}\label{Eqn:v=Pvv}
\tilde{v}^{(n)}_{\mu_n} = \sum_{\mu_{n-1}, k} P_n^{\mu_n, \mu_{n-1}, k} \tilde{v}^{(n-1)}_{\mu_{n-1}} v_k , \qquad P_n^{\mu_n, \mu_{n-1}, k} \in \{0,1\} . 
\end{equation}
These retokenized features form the model’s internal vocabulary. 

The $P_n$ are grown layer-by-layer over a finite window of training tokens. Growths form if $g_{\mu_{n-1},k}^{(n)}>\epsilon_n$ for cutoff $\epsilon_n$. Importantly, the projection maps are constrained by a \emph{smoothness condition}: a new \( n \)-gram token is allowed only if all its subcomponents are already valid \( (n{-}1) \)-grams. This ensures that every token in the hierarchy is recursively decomposable — i.e retokenizable.

\subsection{Hierarchical Energy and Inference}

The total model energy over all layers is:
\begin{equation}
H = \sum_{n \geq 2} \sum_{\mu_{n-1},k} g^{(n)}_{\mu_{n-1},k} \, \tilde{v}^{(n-1)}_{\mu_{n-1}} \, v_k .
\end{equation}
Inference corresponds to computing the energy gradient with respect to the next token:
\begin{equation}
\frac{\partial H}{\partial v_k} = \sum_{n \geq 2} \sum_{\mu_{n-1}} g^{(n)}_{\mu_{n-1},k} \tilde{v}^{(n-1)}_{\mu_{n-1}} ,
\end{equation}
with update rule $v_k\leftarrow f_k(\frac{\partial H}{\partial v_k})$, where $f$ is a bounded function acting on every element of $v_k$. This quantity defines a fast autoregressive prediction mechanism: each token is generated by summing over its correlations with retokenized contexts across multiple scales. The value of the next token is sampled from a probability distribution, e.g $\rho_k(v_k)=e^{v_k}/\sum_{j}e^{v_j}$. Note, the model can be biased to favor longer-range dependencies by rescaling weights of deeper layers: \( g^{(n)} \rightarrow \beta^n g^{(n)} \) with \( \beta > 1 \).

\section{Retokenization and Emergent $n$-Point Structure}
\label{Sec:Retokenization}

Although each layer of the model stores only local 2-point correlations between a projected context token and a predicted token,
\[
H_n = \sum_{\mu_{n-1}, k} g^{(n)}_{\mu_{n-1}, k} \, \tilde{v}^{(n-1)}_{\mu_{n-1}} \, v_k ,
\]
the recursive definition of projected tokens induces an implicit \(n\)-point interaction. Each projection is defined via Eqn \ref{Eqn:v=Pvv} and applying this recursively from the basis layer (\(n=1\)) yields:
\begin{align}
\tilde{v}^{(n-1)}_{\mu_{n-1}} 
&= \sum_{j_1,\ldots,j_{n-1}} (P_{n-1}\cdots P_3P_2)^{\mu_{n-1}}_{j_1 \cdots j_{n-1}} 
v_{j_1}v_{j_2} \cdots v_{j_{n-1}} , \\
\Rightarrow\quad
H_n &= \sum_{j_1,\ldots,j_n} g^{(n)}_{j_1 \cdots j_n} \, v_{j_1} \cdots v_{j_n} ,
\end{align}
where
\[
g^{(n)}_{j_1 \cdots j_n} = \sum_{\mu_{n-1}} g^{(n)}_{\mu_{n-1}, j_n} 
\cdot 
(P_{n-1}\cdots P_2)^{\mu_{n-1}}_{j_1 \cdots j_{n-1}} .
\]
Thus, a learned 2-point memory update at level \(n\) becomes an effective \(n\)-point interaction through hierarchical retokenization. This allows the model to encode long-range dependencies in a compressed form using only local unsupervised learning events.


\subsection{Retokenization Symmetries and Gauge Equivalence}

The same tensor \( \tilde{v}^{(n)} \) can be expressed with either left- or right-tokenized orderings of its internal structure. These forms are related via recursive retokenization of the projection tensors. Given a right-tokenized projector \( P_n^{\mu_n, \mu_{n-1}, k} \), we define the corresponding left-tokenized form recursively as:
\begin{equation}
P_{L,n}^{\mu_n, j_1, \mu_{n-1}} 
= \sum_{k, \mu_{n-2}} 
\left( \sum_{\mu_{n-1}'} 
P_n^{\mu_n, \mu_{n-1}', k} 
\cdot 
P_{L,n-1}^{\mu_{n-1}', j_1, \mu_{n-2}} 
\right)
\cdot 
P_{n-1}^{\mu_{n-1}, \mu_{n-2}, k} ,
\end{equation}
with base case \( P_{L,2}^{\mu_2, j_1, j_2} := P_2^{\mu_2, j_1, j_2} \). These transformations preserve the \emph{smooth gauge structure} of the model, and maps us between different representations of the same smooth tensor, which are distinguished by the order \& number of indices. (E.g $g^{(n)}_{\mu_n}$, $g^{(n)}_{\mu_{n-1},k}$, \& $g^{(n)}_{k,\mu_{n-1}}$ are different representations of $g^{(n)}$.) Where reordering is done via retokenization and {\it not} transposition ($g^{(n)}_{\mu_{n-1},k}\neq g^{(n)T}_{k,\mu_{n-1}}$ in general). In practice, this means: 
\[
\sum_{\mu_{n-1},k} P_n^{\mu_n,\mu_{n-1},k} \tilde{v}^{(n-1)}_{\mu_{n-1}} v_k 
=
\sum_{k,\mu_{n-1}} P_{L,n}^{\mu_n,k,\mu_{n-1}} v_{k} \tilde{v}^{(n-1)}_{\mu_{n-1}} .
\]

This formalism ensures that all token representations and learning updates remain strictly local and compositional, while implicitly capturing longer correlations. 

We may understand this redundancy of the hierarchical featureset as manifesting an emergent associativity, such that the tensor product $v_{\bf c}v_{\bf a}v_{\bf t}$ can be recast as $\tilde{v}_{\bf cat}^{(3)}$, $\tilde{v}_{\bf ca}^{(2)}v_{\bf t}$, \& $v_{\bf c}\tilde{v}_{\bf at}^{(2)}$ via $P_n$'s. We exploit this in the following sections.

\subsection{Left-Inference via Retokenization}

To perform left-inference---predicting the \emph{initial} token of a word given trailing context---we retokenize the energy function into a left-gauge form. For each layer \( n \), the hierarchical Hebbian-Ising model learns 2-point correlations of the form:
\begin{equation}
H_n = \sum_{\mu_{n-1}, k} g^{(n)}_{\mu_{n-1}, k} \, \tilde{v}^{(n-1)}_{\mu_{n-1}} \, v_k ,
\end{equation}
where \( \tilde{v}^{(n-1)} \) is the projected trailing context, and \( v_k \) is the final (predicted) token. This is the default right-inference form.

To express \( H_n \) in a form suitable for left-inference, we apply the left-tokenized projector \( P_{L,n}^{\mu_n, j, \mu_{n-1}} \), defined recursively via:
\begin{equation}
\tilde{v}^{(n)}_{\mu_n} = \sum_{j, \mu_{n-1}} P_{L,n}^{\mu_n, j, \mu_{n-1}} \, v_j \tilde{v}^{(n-1)}_{\mu_{n-1}} .
\end{equation}

The energy becomes:
\begin{equation}
H_n = \sum_{\mu_n} g^{(n)}_{\mu_n} \, \tilde{v}^{(n)}_{\mu_n} 
= \sum_{j, \mu_{n-1}, \mu_n} g^{(n)}_{\mu_n} \, P_{L,n}^{\mu_n, j, \mu_{n-1}} \, v_j \tilde{v}^{(n-1)}_{\mu_{n-1}} .
\end{equation}

To compute the gradient for left-inference, we differentiate with respect to the initial token \( v_j \):
\begin{equation}
\frac{\partial H}{\partial v_j} 
= \sum_{n \geq 2} \sum_{\mu_{n-1}, \mu_n} g^{(n)}_{\mu_n} \, P_{L,n}^{\mu_n, j, \mu_{n-1}} \, \tilde{v}^{(n-1)}_{\mu_{n-1}} .
\end{equation}

This is structurally symmetric to the right-inference form:
\begin{equation}
\frac{\partial H}{\partial v_k} 
= \sum_{n \geq 2} \sum_{\mu_{n-1}} g^{(n)}_{\mu_{n-1}, k} \, \tilde{v}^{(n-1)}_{\mu_{n-1}} ,
\end{equation}
but now uses the left-retokenized projector and reversed token ordering.


\subsection{Example: Learning from {\bf dogs}, {\bf cats}, and {\bf frogs}}

To illustrate how the hierarchical Ising model builds structure from scratch, we walk through the learning process for a minimal training set consisting of three words:
\[
{\bf dogs},\quad {\bf cats},\quad {\bf frog} .
\]

We assume the model begins with a basis token set of dimension \( d=26 \), one for each letter of the English alphabet: \( v_j \in \{ v_{\bf a}, \dots, v_{\bf z} \} \). The model learns from co-occurrence patterns in the training data, using local Hebbian updates following Eqn \ref{Eqn:HebbUpdate}.

\vspace{1em}
\noindent
\textbf{Step 1: Learning Bigrams}

From the three words, the following bigrams (2-grams) occur:
\[
{\bf do},\ {\bf og},\ {\bf gs},\quad
{\bf ca},\ {\bf at},\ {\bf ts},\quad
{\bf fr},\ {\bf ro},\ {\bf og}.
\]
The model defines a projection tensor \( P_2^{\mu_2, j_1, j_2} \), with a separate index \( \mu_2 \) for each learned bigram:
\[
P_2 = 
\sum_{\mu_2}
P_2^{\mu_2, j_1, j_2} \,
\tilde{v}^{(2)}_{\mu_2} \, v_{j_1} v_{j_2} ,
\]
where \( P_2^{\mu_2, j_1, j_2} = 1 \) for each bigram above. These define eight new 2-gram tokens:
\[
\tilde{v}^{(2)}_{\bf do},\ 
\tilde{v}^{(2)}_{\bf og},\ 
\tilde{v}^{(2)}_{\bf gs},\ 
\tilde{v}^{(2)}_{\bf ca},\ 
\tilde{v}^{(2)}_{\bf at},\ 
\tilde{v}^{(2)}_{\bf ts},\ 
\tilde{v}^{(2)}_{\bf fr},\ 
\tilde{v}^{(2)}_{\bf ro}.
\]

\vspace{1em}
\noindent
\textbf{Step 2: Learning Trigrams}

The trigrams (3-grams) in the words are:
\[
{\bf dog},\ {\bf ogs},\quad 
{\bf cat},\ {\bf ats},\quad 
{\bf fro},\ {\bf rog}.
\]
From the smooth bigrams above, the model constructs a new projector:
\[
P_3^{\mu_3, \mu_2, j_3} = 1 
\quad \text{for learned combinations} \quad 
\tilde{v}^{(3)}_{\mu_3} \cong \tilde{v}^{(2)}_{\mu_2} \, v_{j_3}.
\]
This yields six smooth trigram tokens:
\[
\tilde{v}^{(3)}_{\bf dog},\ 
\tilde{v}^{(3)}_{\bf ogs},\ 
\tilde{v}^{(3)}_{\bf cat},\ 
\tilde{v}^{(3)}_{\bf ats},\ 
\tilde{v}^{(3)}_{\bf fro},\ 
\tilde{v}^{(3)}_{\bf rog}.
\]

\vspace{1em}
\noindent
\textbf{Step 3: Learning 4-grams (Full Words)}

Finally, the full words {\bf dogs}, {\bf cats}, and {\bf frog} are learned:
\begin{equation}
\tilde{v}^{(3)}_{\bf dog} \, v_{\bf s} 
\quad \Rightarrow \quad 
\tilde{v}^{(4)}_{\bf dogs},\qquad
\tilde{v}^{(3)}_{\bf cat} \, v_{\bf s} 
\quad \Rightarrow \quad 
\tilde{v}^{(4)}_{\bf cats}, \notag
\end{equation}
\vspace{-1em}
\begin{equation}
\tilde{v}^{(3)}_{\bf fro} \, v_{\bf g} 
\quad \Rightarrow \quad 
\tilde{v}^{(4)}_{\bf frog}. \notag
\end{equation}

\vspace{1em}
\noindent
\textbf{Summary: Smooth Tokenset}

After training, the smooth tokenset contains:
\begin{itemize}
  \item 8 bigrams:
  \( \tilde{v}^{(2)}_{\bf do},\ 
      \tilde{v}^{(2)}_{\bf og},\ 
      \tilde{v}^{(2)}_{\bf gs},\ 
      \tilde{v}^{(2)}_{\bf ca},\ 
      \tilde{v}^{(2)}_{\bf at},\ 
      \tilde{v}^{(2)}_{\bf ts},\ 
      \tilde{v}^{(2)}_{\bf fr},\ 
      \tilde{v}^{(2)}_{\bf ro} \)
  \item 6 trigrams:
  \( \tilde{v}^{(3)}_{\bf dog},\ 
      \tilde{v}^{(3)}_{\bf ogs},\ 
      \tilde{v}^{(3)}_{\bf cat},\ 
      \tilde{v}^{(3)}_{\bf ats},\ 
      \tilde{v}^{(3)}_{\bf fro},\ 
      \tilde{v}^{(3)}_{\bf rog} \)
  \item 3 4-grams:
  \( \tilde{v}^{(4)}_{\bf dogs},\ 
      \tilde{v}^{(4)}_{\bf cats},\ 
      \tilde{v}^{(4)}_{\bf frog} \)
\end{itemize}

\vspace{1em}
\noindent
\textbf{Final Projectors}

The learned projectors take the form:
\begin{equation}
\begin{aligned}
P_2^{\mu_2, j_1, j_2} &= 1 
\quad \text{for } (\mu_2,j_1 j_2) \in 
\scriptstyle{\{ ({\bf do},{\bf d},{\bf o}),\ ({\bf og},{\bf o},{\bf g}),\ ({\bf gs},{\bf g},{\bf s}),\ 
   ({\bf ca},{\bf c},{\bf a}),}\\ 
   &\hspace{8.5em}
   \scriptstyle{({\bf at},{\bf a},{\bf t}),\ ({\bf ts},{\bf t},{\bf s}),\ 
   ({\bf fr},{\bf f},{\bf r}),\ ({\bf ro},{\bf r},{\bf o}) \}} \\
P_3^{\mu_3, \mu_2, j_3} &= 1 
\quad \text{for } (\mu_3, \mu_2, j_3) \in 
\scriptstyle{\{ ({\bf dog}, {\bf do}, {\bf g}),\ 
   ({\bf ogs}, {\bf og}, {\bf s}),\ 
   ({\bf cat}, {\bf ca}, {\bf t}),}\\
   &\scriptstyle{\hspace{8.5em}
   ({\bf ats}, {\bf at}, {\bf s}),\
   ({\bf fro}, {\bf fr}, {\bf o}),\ 
   ({\bf rog}, {\bf ro}, {\bf g}) \}} \\
P_4^{\mu_4, \mu_3, j_4} &= 1 
\quad \text{for } (\mu_4, \mu_3, j_4) \in 
\scriptstyle{\{ ({\bf dogs}, {\bf dog}, {\bf s}),\ 
   ({\bf cats}, {\bf cat}, {\bf s}),\
   ({\bf frog}, {\bf fro}, {\bf g}) \}}
\end{aligned}
\end{equation}
Or equivalently written as a superposition of tensor product states (a.k.a an {\bf entangled} state \cite{mps_Schollwock2011}):
\begin{eqnarray}
P_2 &=& \tilde{v}_{\bf do}^{(2)}v_{\bf d}v_{\bf o} + \tilde{v}_{\bf og}^{(2)}v_{\bf o}v_{\bf g} + \tilde{v}_{\bf gs}^{(2)}v_{\bf g}v_{\bf s} + \tilde{v}_{\bf ca}^{(2)}v_{\bf c}v_{\bf a} + \tilde{v}_{\bf at}^{(2)}v_{\bf a}v_{\bf t} \notag\\
&+& \tilde{v}_{\bf ts}^{(2)}v_{\bf t}v_{\bf s} + \tilde{v}_{\bf fr}^{(2)}v_{\bf f}v_{\bf r} + \tilde{v}_{\bf ro}^{(2)}v_{\bf r}v_{\bf o} \\
P_3 &=& \tilde{v}_{\bf dog}^{(3)}\tilde{v}_{\bf do}^{(2)}v_{\bf g} + \tilde{v}_{\bf ogs}^{(3)}\tilde{v}_{\bf og}^{(2)}v_{\bf s} + \tilde{v}_{\bf cat}^{(3)}\tilde{v}_{\bf ca}^{(2)}v_{\bf t} + \tilde{v}_{\bf ats}^{(3)}\tilde{v}_{\bf at}^{(2)}v_{\bf s} \notag\\
&+& \tilde{v}_{\bf fro}^{(3)}\tilde{v}_{\bf fr}^{(2)}v_{\bf o} + \tilde{v}_{\bf rog}^{(3)}\tilde{v}_{\bf ro}^{(2)}v_{\bf g} \\
P_4 &=& \tilde{v}_{\bf dogs}^{(4)}\tilde{v}_{\bf dog}^{(3)}v_{\bf s} + \tilde{v}_{\bf cats}^{(4)}\tilde{v}_{\bf cat}^{(3)}v_{\bf s} + \tilde{v}_{\bf frog}^{(4)}\tilde{v}_{\bf fro}^{(3)}v_{\bf g}
\end{eqnarray}
We can understand the product of projector tensors $P_nP_{n-1}\cdots P_2$ as encoding a bundle of finite state transitions (or feature maps), e.g $v_{\bf r}\rightarrow\tilde{v}^{(2)}_{\bf fr}\rightarrow\tilde{v}^{(3)}_{\bf fro}\rightarrow\tilde{v}^{(4)}_{\bf frog}$, which collectively describe all possible paths toward the terminal nodes of the DAG. 

In formal language terms, the feature maps act as finite-state automata which accepts strings whose complexity is subregular\footnote{Formally, regular languages are recognized by finite-state automata that lack explicit memory for counting or nesting \cite{Hopcroft&Motwani&Ullman2001book_AutomataTheoryLanguages&Computation}.}.

\begin{figure}
    \centering
    \includegraphics[width=.95\linewidth]{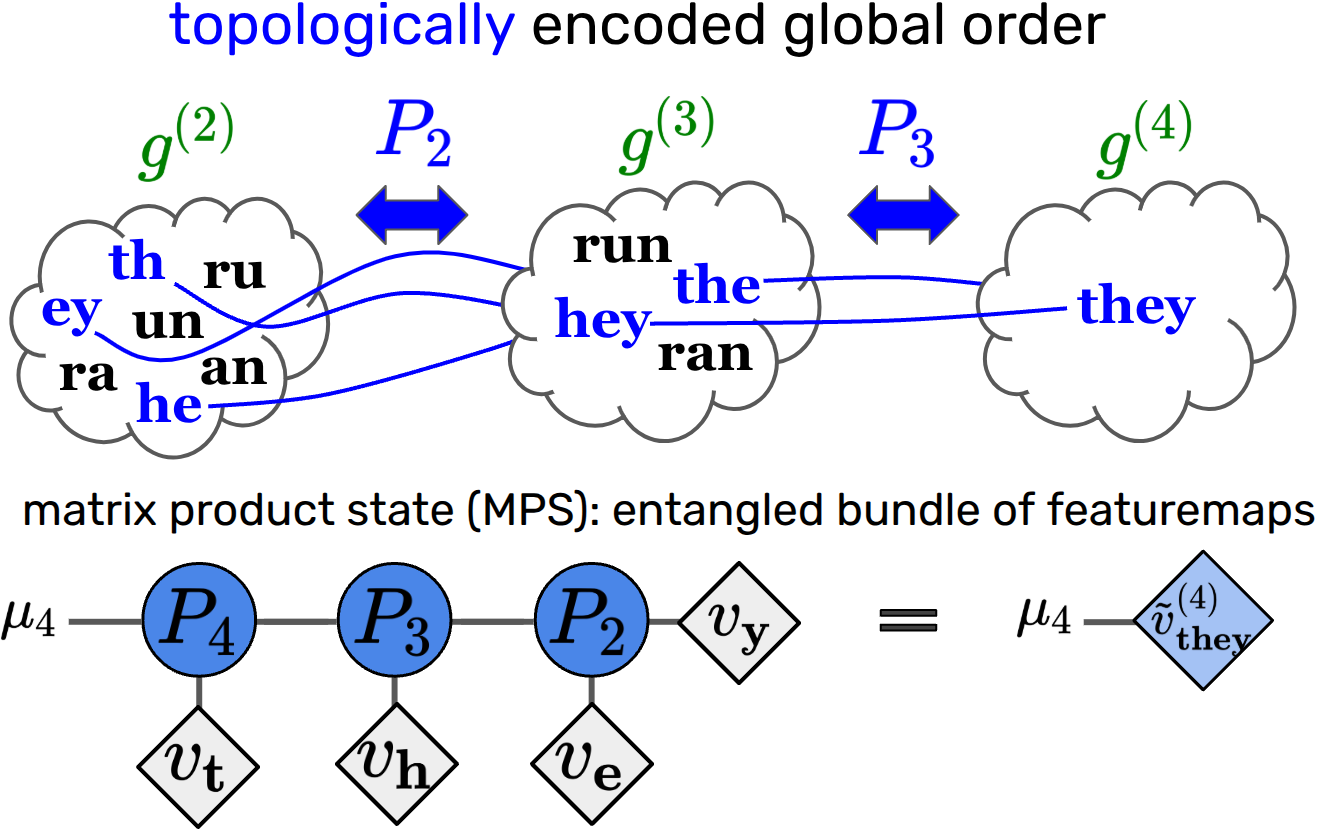}
    \caption{Local interactions ($\tilde{v}^{(n-1)}\otimes v$) drive new model growths ($\tilde{v}^{(n)}$) when input events align with the allowed hierarchical DAG structure. Stored strings are induced global order parameters ($v\otimes v\otimes\cdots\otimes v\rightarrow\tilde{v}^{(n)}$), each living as a terminal node of the DAG. Each growth is compressed/decompressed as a matrix product state (MPS).}
    \label{Fig:TopologicalEncoding}
\end{figure}

\subsubsection{{\bf Topology of the Ground State Automaton}}\hfill

Note that, given context {\bf frog}, the model knows to terminate right-inference because no merger of $\tilde{v}^{(4)}_{\bf frog}$ with another token exists as a feature map -- defined by non-vanishing action of $P_{n}P_{n-1}\cdots P_2$. Nevertheless, the model can still infer {\bf frogs} as the best non-smooth construction due to local interactions $g^{(2)}_{\bf og}$ \& $g^{(3)}_{\bf ogs}$.

This latter subtlety highlights the hybrid sub-symbolic/automaton nature of the model, where tokenization\footnote{This dynamic tokenization is distinct from both tokenization \cite{Schuster&Nakajima2012_SubwordTokenization,Gage1994_BPE} \& entropy-based approaches to LLM's \cite{Li&all2025_EntropyAwareGeneration,Zhang&all2024_EntropyAwareGeneration,Agrawal&all2024_EntropyAwareGeneration,Simonds2025_EntropyAwareGeneration}. LLM's assume pretokenized input \cite{Phuong&Hutter2022review_DeepMind_Transformers,Gage1994_BPE,Schuster&Nakajima2012_SubwordTokenization}.} plays a dynamical role during inference to determine boundaries, and therefore imbues the model with a notion of {\it accepted} string. In energetic terms, the accepted strings live in a low-energy translationally-invariant manifold, and therefore are the most likely configurations to be observed given the context. (Thus in our example, {\bf frogs} is {\it well-formed} but not accepted.)

Their corresponding automata live in the ground state, with their formal language being the ground state configurations \cite{Reinhart&DeLasCuevas2022_IsingGrammar}. A projection onto the ground state manifold\footnote{This ground-state projection is similar to, for example, how high magnetic fields and low temperatures can effectively project the electronic states of a quantum Hall magnetic onto its lowest Landau level \cite{EugenioDag2020scipost}. Model parameters \& environmental conditions constrain the model onto a gapped isolated subspace in the spectrum.} guarantees an emergent automatonic behavior. 
Local discorrelations manifest as jumps in energy, up out of the ground state manifold. However, the projection sharpens this behavior to exactness, such that the projectors can be used in placed of energy/entropy jumps. Hence the distinction of accepted vs well-formed as one of being respectively in or just above the manifold.

\begin{figure}
    \centering
    \includegraphics[width=.7\linewidth]{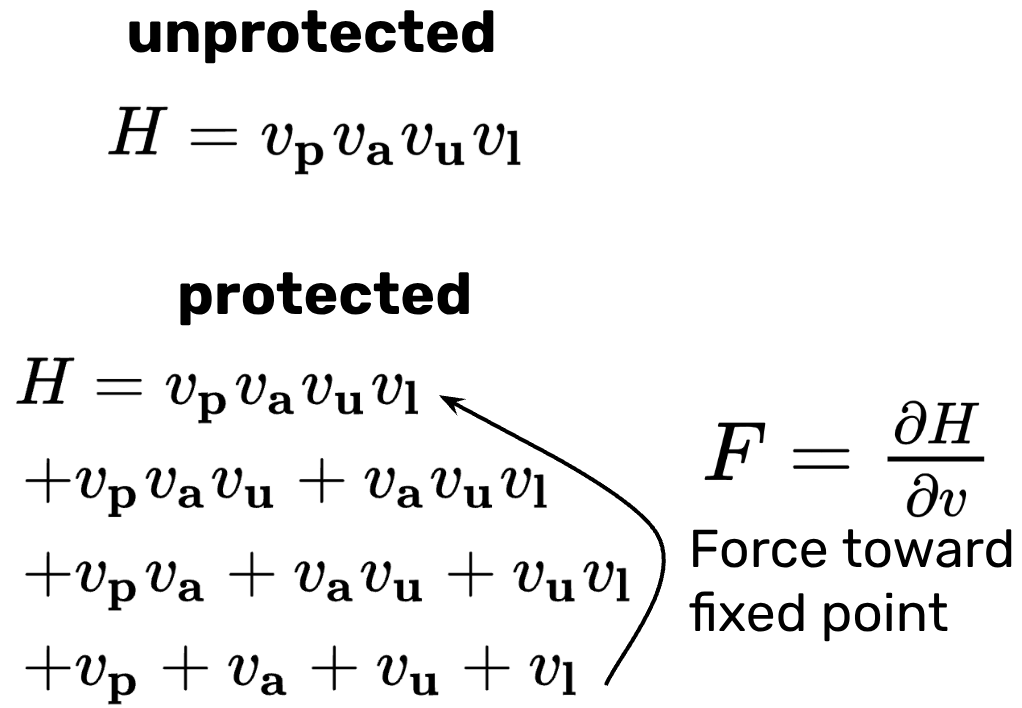}
    \caption{Comparing Hamiltonians of topologically protected global order vs unprotected. Context overlapping with a feature is driven to the fixed point (energy extremum) by a restoring force. We use convention for energy $E=-H$.}
    \label{Fig:TopProtGlobalOrder}
\end{figure}

Further we stress that the ground state is {\bf topological}. This is a non-trivial statement, not generically guaranteed for every ground state behavior of a physical system, and arises here because a global order parameter is learned in a strictly local way. The topology of the retokenizer is the topology of the DAG graph, with the strict {\it every $n$-gram feature be composed of $(n-1)$-gram features} smoothness constraint. The terminal nodes of the DAG are special features towards which the autoregressive flows are directed. Thus retrieval of terminal nodes is topologically guaranteed (e.g Fig \ref{Fig:TopProtGlobalOrder}).\footnote{This parallels recent industry advances, which exploit bifix-free data structures to reduce interference during retrieval \cite{Augeri2025_HyperTokens,Augeri2008thesis} in LLMs.}

Enforcing this constraint is equivalent to demanding gauge invariance: define left (right) retokenization operators 
\begin{eqnarray}
\mathcal{L}_n:\tilde{v}^{(n-1)}v\rightarrow v\tilde{v}^{(n-1)} \notag\\
\mathcal{R}_n:v\tilde{v}^{(n-1)}\rightarrow \tilde{v}^{(n-1)}v \notag
\end{eqnarray}
as series of projections \& reprojections using $P_n$ \& $P_{L,n}$. The product $\mathcal{R}\mathcal{L}$ is the projector onto the smooth tokenset\footnote{Note the gauge structure of retokenization arises because we use dense matrices (projectors) to describe the information of a sparse graph. However, the brain is not dense, and therefore {\it enforcing gauge invariance} doesn't implicitly mean anything, as gauges are non-physical artifacts of the dense representation. In our companion work \cite{Eugenio2025_HebbianLanguage}, we demonstrate how the equivalent DAG constraint is a likely consequence of the combination of {\it translational invariance} plus a {\it simple threshold function}.}, and is equivalent to applying a DAG mask. Thus we can interpret the exact retrieval of the global order parameter as being directly tied to the ability of the model to perform both right \& left token prediction consistent within a singular mathematical structure. (Or equivalently converging to the terminal nodes of the sparse graph.)

\begin{figure}
    \centering
    \includegraphics[width=.95\linewidth]{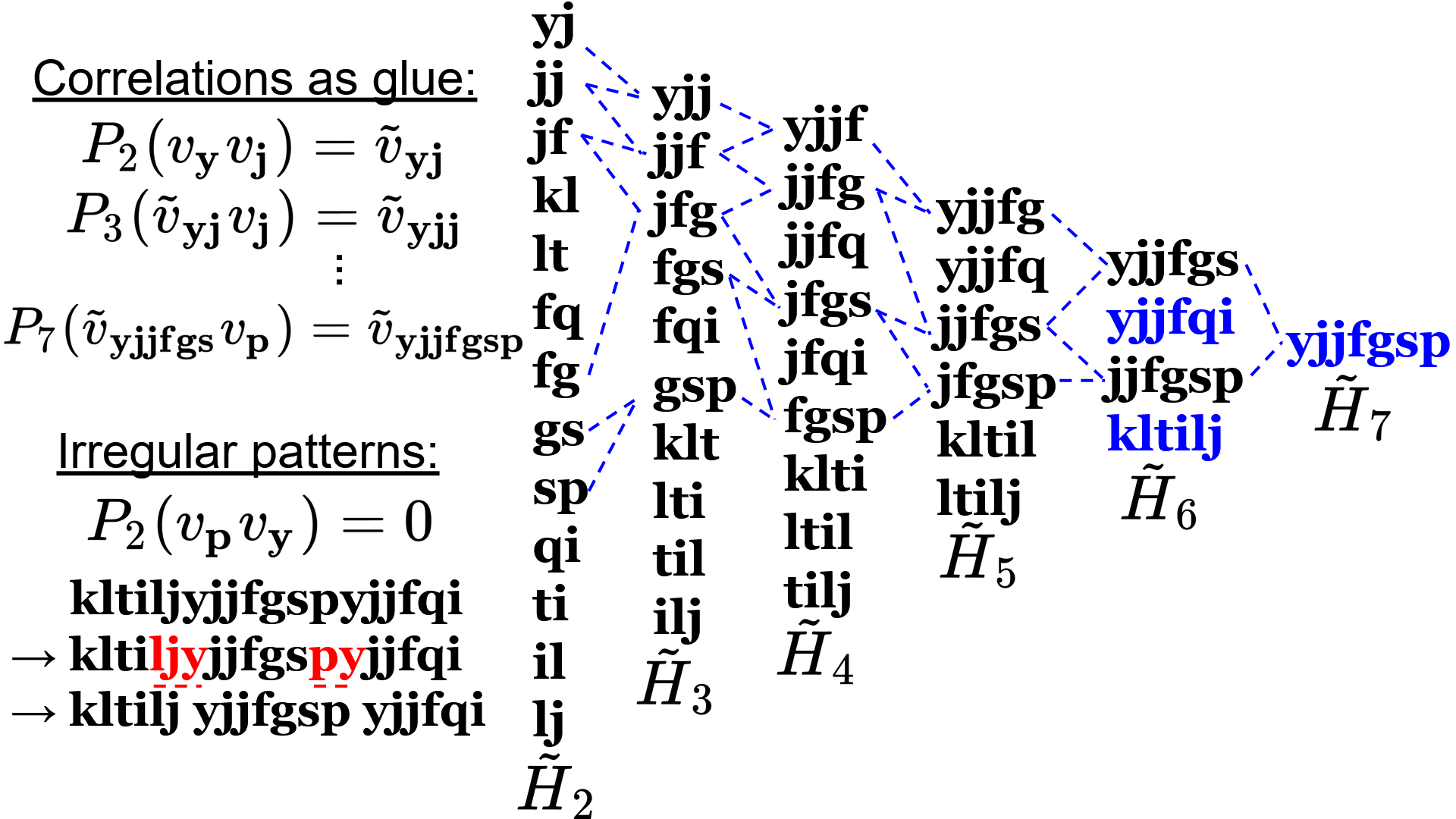}
    \caption{Summarizing the relationship between retokenization, locality, and hierarchy. Three words from a random language \cite{Eugenio2025_HebbianLanguage} (shown in blue): {\color{blue}\bf yjjfqi}, {\color{blue}\bf kltilj}, \& {\color{blue}\bf yjjfgsp}.}
    \label{Fig:TokLocHier}
\end{figure}

The ground state topology enforces a global compositional word-forming vs non-word forming distinction: you are either in the graph or you're not. Non-word forming patterns aren't just local discorrelations, measurable as a jump in entropy, but topological defects tied to the global composition rule. If a pattern is not a word forming pattern, then it necessarily contains a boundary (e.g Fig \ref{Fig:TokLocHier}).

Unlike estimating continuous information, where discorrelations reflect changes in a physical medium, what defines a ``good" vs ``bad" word pattern can be arbitrary. Presenting the model one word at a time during training allows it to learn what a ``good" pattern is. Yet {\bf human sub-word structure is \underline{not} arbitrary}, but has a {\bf specific structure reflecting the underlying neural topology}. This becomes apparent when the model is allowed to self-organize (i.e trained on unstructured noise or grown randomly). We discuss this more in Section \ref{Sec:UniversalSubWordStructure}.


\subsubsection{Quick note about {\bf spaces}}\label{Sec:Spaces}

Neither punctuation nor spacing are necessary elements of writing:
$$\texttt{youcanreadthisoranysuchtextjustfine}.$$
Historically, human writing existed without specialized tokens indicating boundary. This is best illustrated by alphabets, whose fundamental tokens are meaningless beyond their sound associations. Spaces (and other punctuation) are not real tokens inside human language patterns, but are external annotations of statistical boundaries.

In the context of the model, these boundaries are high energy discorrelations. We can lower the energy of a string by inserting a zero vector, e.g $[v_{\bf y},v_{\bf o},v_{\bf u},\vec{0},v_{\bf c},v_{\bf a},v_{\bf n}]$. But $\vec{0}$ is not a token, as its impossible to learn correlations with it. Were one to introduce a true space token, it would be interpretted by the retokenizer as being an invisible letter, and learned patterns would no longer correspond to linguistic units. It would instead learn an interpretation of the data reflecting artifacts of the tokenization. 

Thus we present the model here using a basis tokenset of lowercase alphabetic characters. The dimension of the alphabet can be anything so long as the tokenset is {\it intrinsically meaningless}. Not because we necessarily need meaningless tokens, but only because the retokenizer treats all tokens as such regardless. \\

{\bf Boundaries are statistical \& emergent.} This model does not make use of explicit boundary marking. Hierarchical boundaries, at any level of language structure, are an implicit quality of the model. This is discussed more in the context of syntax in Sec \ref{Sec:Syntax}.

\begin{figure}
    \centering
    \includegraphics[width=.85\linewidth]{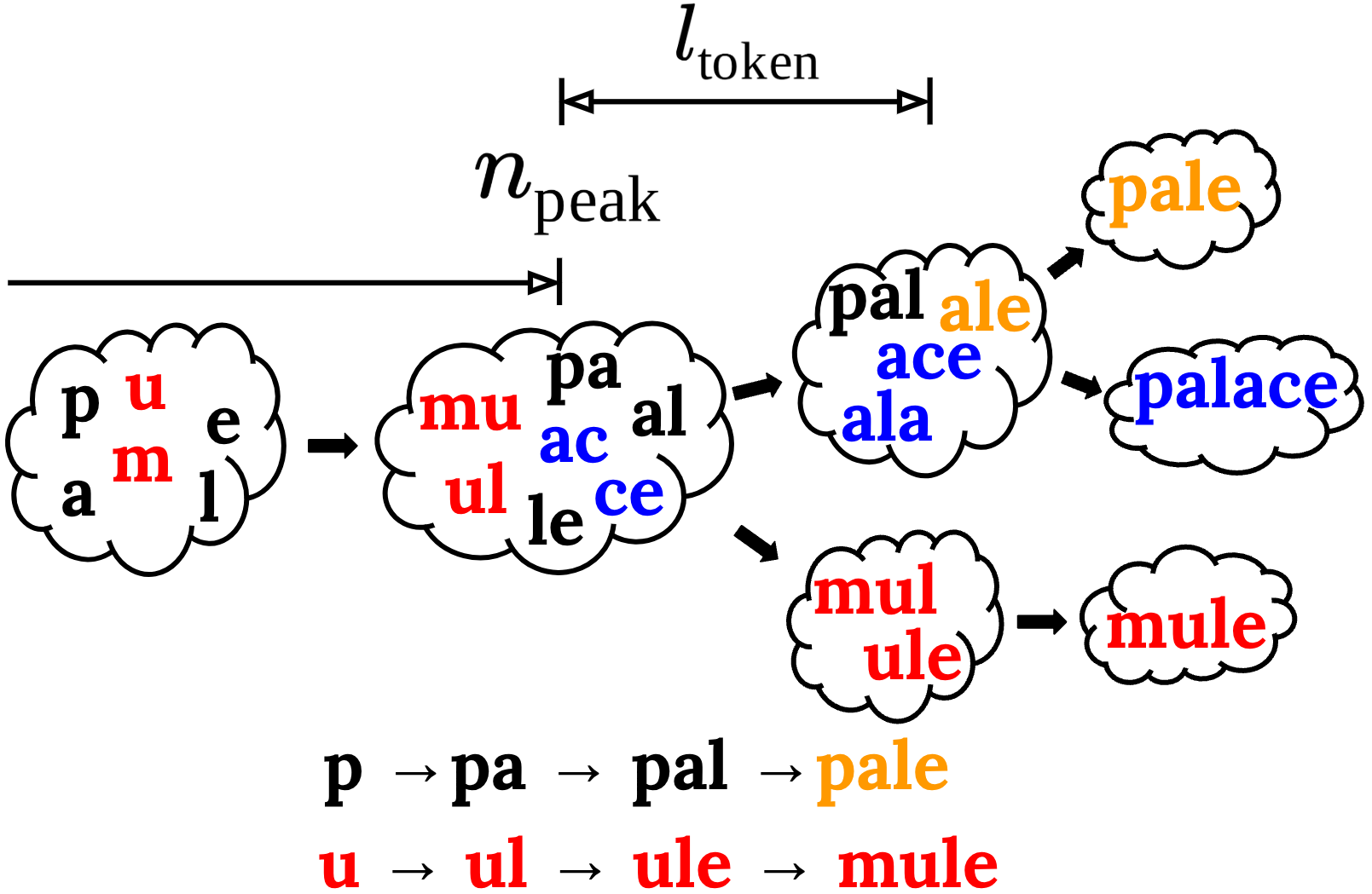}
    \caption{Hierarchical short-term memory retrieves via flows toward terminal nodes. Multiple terminal nodes compete during retrieval if they share features. However, retrieval becomes scale invariant beyond last overlap. Note: DAG's are disentangled during formation of long-term memory---see Fig \ref{Fig:plasticity&replay}.}
    \label{Fig:ScaleInvFlow}
\end{figure}

\begin{figure}
    \centering
    \includegraphics[width=1.01\linewidth]{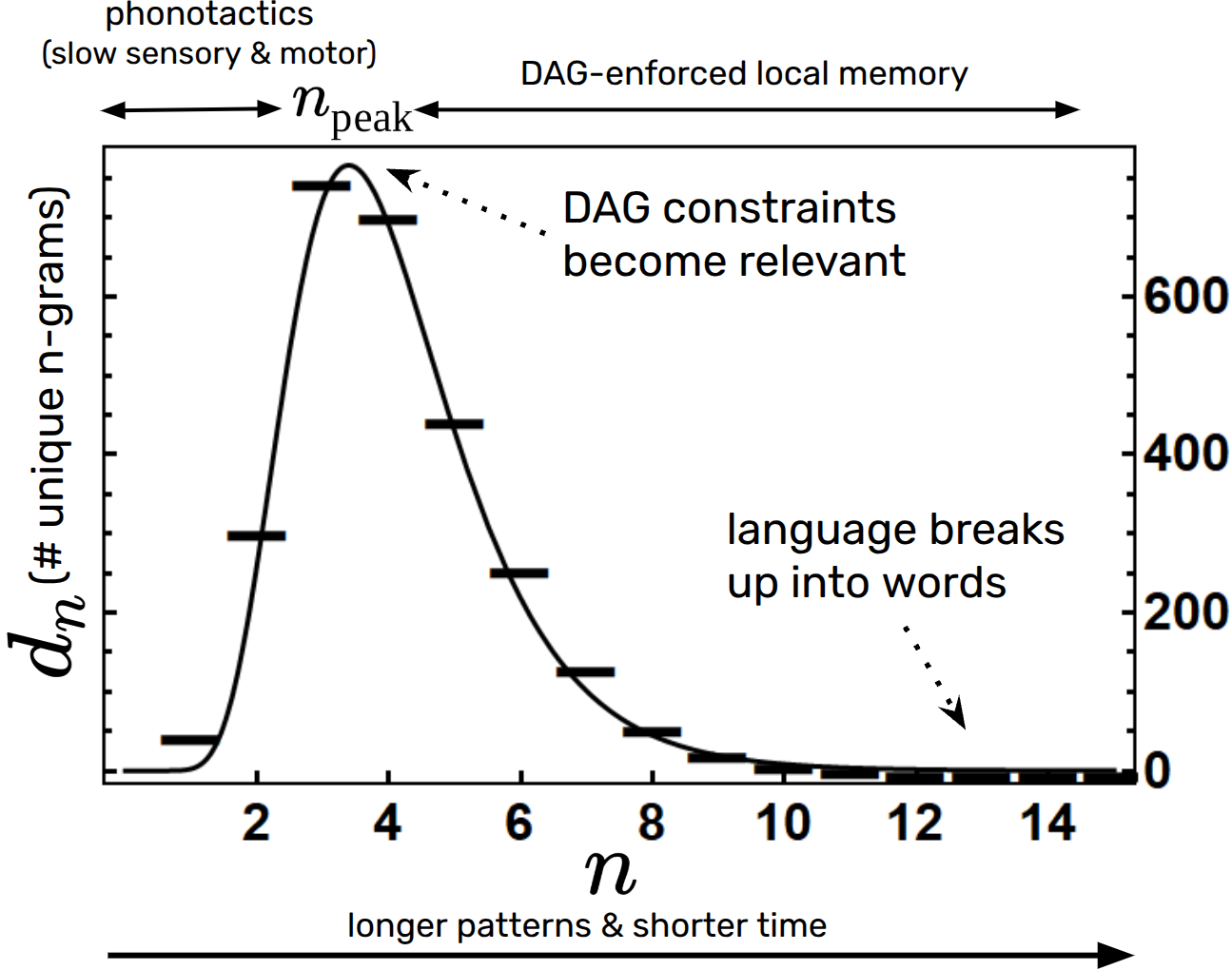}
    \caption{500 words from Alice \& Wonderland \cite{GutenbergAlice}. Solid line is log-normal fit. The shortest lengthscale patterns are biased by the mechanical constraints of pronunciation/listening. But these biases don't account for why word length is finite. Structural constraints of memory become relevant at length scales $n>n_{\text{peak}}$, leading to the $d_n$ collapse, and limiting the length of the longest pattern.}
    \label{Fig:Alice500_collapse}
\end{figure}

\section{Universal Sub-Word Structure}
\label{Sec:UniversalSubWordStructure}

Here we strip away all word meaning and focus on the statistical structure of the underlying patterns.

This structure is quantified in terms of two distributions: (1) the number of unique $n$-grams (a.k.a word-forming patterns) $d_n$ versus length $n$; and (2) the frequency of $n$-grams rank-ordered from most to least frequent (w/ one plot per $n$). 


\subsection{Words as finite patterns \& $d_n$}

The quantity $d_n$ measures the width of the feature space, and is grown layer-by-layer in the model. It has been known for at least half a century that $d_n$ approximates a log-normal distribution for both subword patterns \cite{Herdan1958_logNormal_intraword} and syntax \cite{Williams1940_logNormal_interwords}.

Such a distribution (Fig \ref{Fig:Alice500_collapse}) rises exponentially to a smooth peak at $n_{\text{peak}}$ before collapsing to zero at $n_{\text{max}}$. This collapse to zero coincides with the reality of finite word length, and how it scales with $n$ hints at the structural constraints that enforce that reality. \\

This distribution differs from $d_n$ for unstructured noise (pulled from a uniform distribution)---see Fig \ref{Fig:noise_grams} in Sec \ref{Sec:Suppl:FromNoise}. For noise w/o spaces, $n_{\text{max}}$ is equal to the length of the spaceless string, which is arbitrary. Smaller chunks can be formed by randomly inserting spaces at some frequency, which places a cutoff on length; but the transition across $n_{\text{peak}}$ is sharp (not log-normal), owing to the fact that $n<n_{\text{peak}}$ \& $n>n_{\text{peak}}$ are governed by different distributions. Left of the peak grows exponentially as $d^n$, because there is sufficient random statistics to guarantee the smaller $n$-grams are each observed at least once.

Note however that noise, while unstructured, has structure living inside of it. As we pointed out above, every configuration will be observed given enough noise. By constructing an appropriate filter, we can sift out a structured distribution. One would then describe the filter (not the noise) as being the source of the structured data. The noise is just the catalyst for learning off the filter.

The retokenizer does just this. We have a retokenizer self-organize, seeded by the same spaceless noise string described above (Fig \ref{Fig:noise_grams}). We plot the memories of the retokenizer alongside both the spaceless training string and its random chunks. It's peak is smooth and fits a log-normal.

Note that one does not need fancy tensor calculations to reproduce this result, but can do this themselves by hand using the very same noise data available in Sec \ref{Sec:Suppl:FromNoise}. The rules: (1) If a bigram shows up at least twice, put it in the graph. (2) For $n>2$, keep all $n$-grams which are smooth mergers of the relevant bigrams. (3) The accepted strings are the terminal nodes of the graph.

No arbitrary cutoff is imposed beyond $n=2$, as the smoothness constraint is strong enough to drive the collapse. \\

One doesn't need to explicitly generate noise to learn from in order to generate a DAG graph, but can grow a random DAG from a random subset of bigrams and their mergers (Sec \ref{Sec:Suppl:FromRandomGrowths}). The smoothness acts as a hierarchical constraint, which enforces closure over all possible growths. (We explore these closed hierarchical structures in the next section.) 

In formal language terms, the subregular symbolic structure generates two types of strings: (1) finite strings living at terminal nodes; and (2) a small subset of simple repeating infinite strings, which arise due to self-mergers (e.g $aa\cdot a\rightarrow aaa$). Such self-mergers are more likely to observe with random growths for small $d$; and for any $d$, only extend to infinite in the special limit\footnote{It's worth pointing out that this makes them ``fine-tuned" in the sense that they aren't physically expected. However, there are some simple self-mergers which show up in writing, e.g from {\bf lol}: {\bf lolololol} w/ various length.} where no additional cutoff is imposed for $n>2$. And self-organizing against a noise string makes them exponentially unlikely, because they have to be observed in the string.

\subsection{Hierarchical Feature Mixing \& Random Language}
\label{Sec:HierFeatMix&Ralang}

We begin by analyzing the structure of a randomly sampled vocabulary, consisting of 100 English or Spanish personal names (in Sec \ref{Sec:Suppl:HierFeatureMixing}). We plot the $d_n$ distribution for this set, which follows an approximately log-normal form, with the largest $n$ determined by the longest name in the sample.

\begin{figure}[h!]
\centering
\begin{subfigure}[t]{0.5\textwidth}
    \centering
    \includegraphics[width=.9\linewidth]{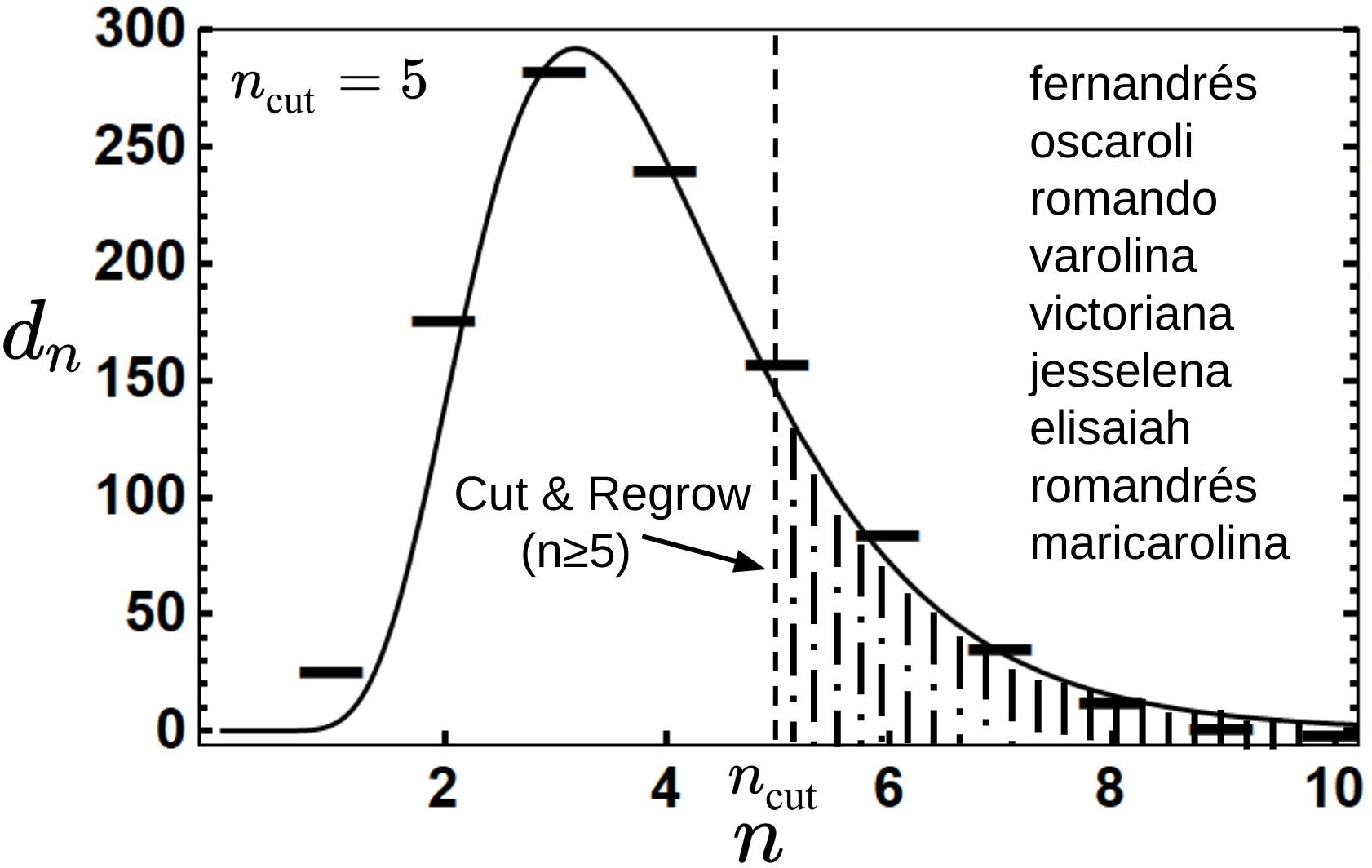}
    \label{Fig:hierarchyCut_1}
\end{subfigure}\\
\begin{subfigure}[t]{0.5\textwidth}
    \centering
    \includegraphics[width=.9\linewidth]{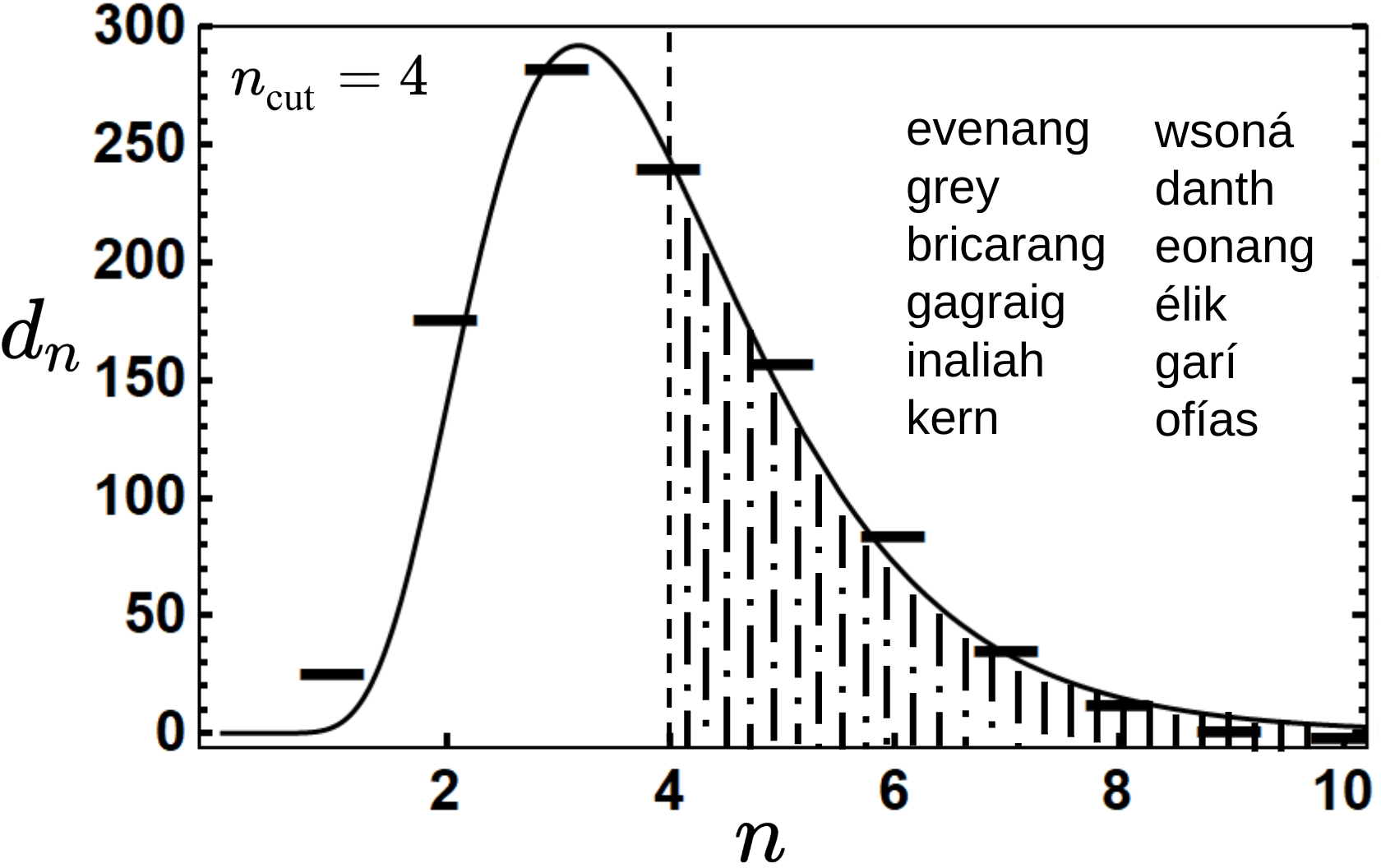}
    \label{Fig:hierarchyCut_2}
\end{subfigure}\\
\begin{subfigure}[t]{0.5\textwidth}
    \centering
    \includegraphics[width=.9\linewidth]{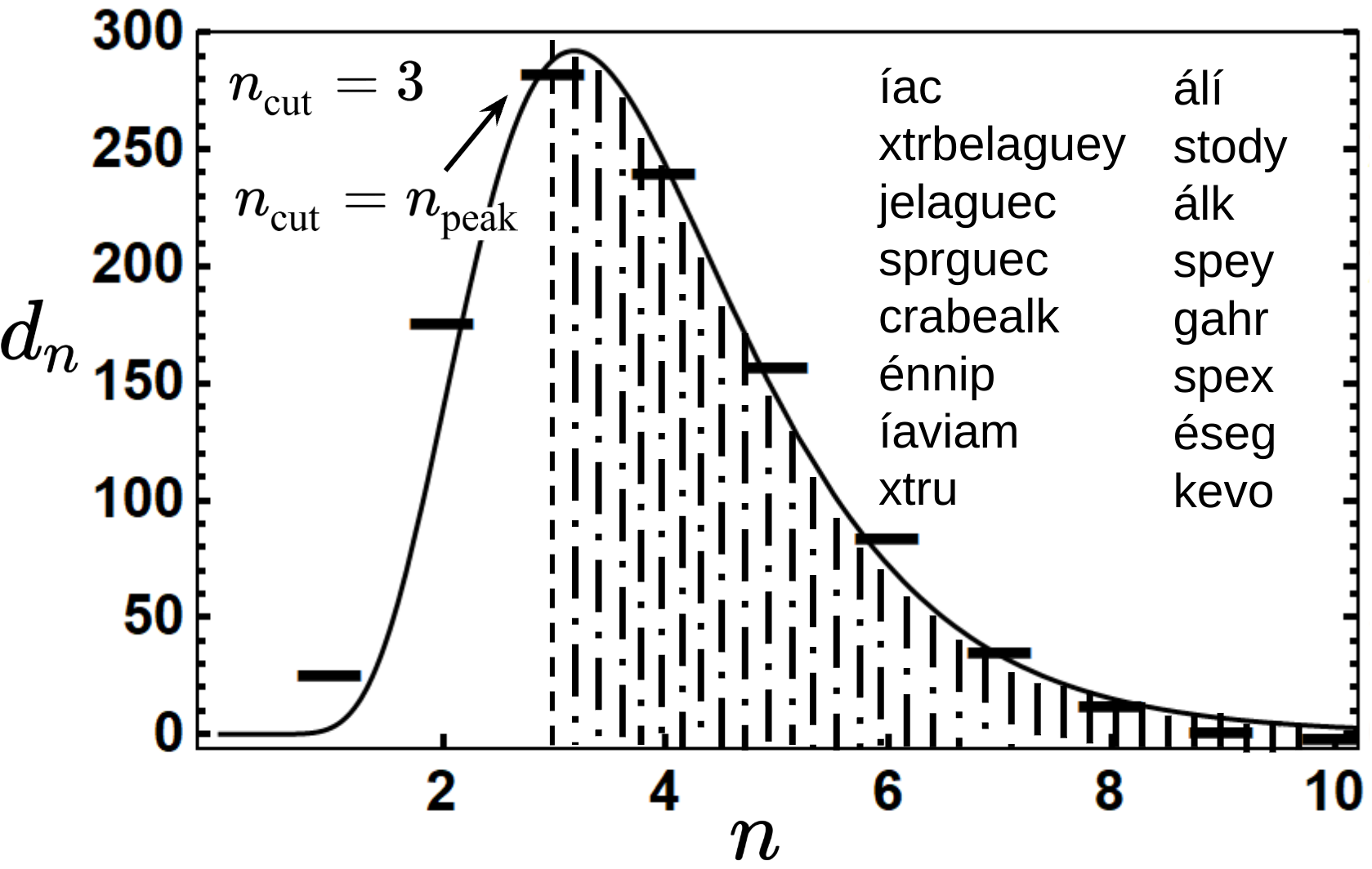}
    \label{Fig:hierarchyCut_3}
\end{subfigure}
\caption{Number of unique \( n \)-grams in a 100-name dataset (English/Spanish mix) shown as backdrop. Shaded regions mark features removed by hierarchical cuts; the solid line fits a log-normal distribution. Example words from post-cut regrowth are shown.}
\label{Fig:hierarchyCut}
\end{figure}

To probe the internal structure of the learned hierarchy, we perform a game of \emph{cut-and-regrow} (Fig \ref{Fig:hierarchyCut}). Specifically, we impose a cutoff $n_{\text{cut}}$ and delete all $n$-gram features with $n\ge n_{\text{cut}}$, then allow the hierarchy to regrow under the smoothness constraint. This process reveals how the underlying directed acyclic graph (DAG) of features governs regrowth: surviving features at small $n$ serve as anchoring nodes, from which higher-order structures emerge. Memory is locally organized, such that the shortest word-length scales dominate.

When cutting at $n_{\text{cut}} \gg n_{\text{peak}}$, where $n_{\text{peak}}$ is the location of the original $d_n$ peak, regrowth closely reproduces the original structure, introducing minimal (or no) novelty. This is because the many lower layers of the hierarchy provide strong constraints on the space of learnable regrowths.

In contrast, cutting closer to $n_{\text{peak}}$ loosens the scale-dependent constraints, allowing novel $n$-grams to form. Nevertheless, even here, the surviving short-scale DAG structure continues to guide regrowth, guaranteeing closure. 

We find that though the original notion of smoothness is perfectly preserved below the cut, cutting away more of the $d_n$ peak causes the hierarchy to begin losing its recognizable English/Spanish nature. This highlights the importance of the $n_{\text{peak}}$ grams to language structure\footnote{For \( n \lesssim n_{\text{peak}} \), \( d_n \) grows approximately exponentially, with learned constraints (\( g^{(n)} = 0 \)) pruning branches and thereby reducing the effective growth rate. For \( n > n_{\text{peak}} \), the smoothness condition takes over: new tokens are permitted only if all subgrams are valid, which recursively restricts composition and causes a collapse in \( d_n \). This collapse guarantees finite word length. In formal language terms, this might be described as a consequence of the subregular limitations of stored strings.}.

Finally, if we cut at $n_{\text{cut}}=2$, we discard {\it all} memory of the original names. This is the data-free regime, where growing a hierarchy generates what we refer to as a ``random language" \cite{Eugenio2025_HebbianLanguage}. Nevertheless, growing hierarchies under the smoothness constraint still yields finite word-lengths and an approximately log-normal $d_n$ distribution \cite{Herdan1958_logNormal_intraword,Williams1940_logNormal_interwords,Eckhard&Werner&Markus2001_logNormal_review}. This shows that bounded, hierarchical morphology is an intrinsic feature of the model itself, independent of any training data, and that language structure arises naturally from smooth hierarchical learning.

Further, this behavior illustrates a key property of the model: Short-term memory structures $g^{(n)}$ behave like a dynamic \emph{scratch board}, where hierarchical feature mixing enables creative recombination under smoothness preservation. The model's DAG structure naturally channels this process, enforcing compositionality while permitting exploration of new morphological forms.

\subsection{Synaptic Zipf-laws} 
\label{Sec:HierFeatMix&Ralang:Zipf}


The rank-ordered frequencies (a.k.a Zipfian plots \cite{Zipf1949,Zanette&Montemurro2002}) measure the relevancy (ranked from most to least frequent) of those features in the sample---Fig \ref{Fig:spectra_bjfyi}. We can interpret these ranked frequencies as defining an effective memory spectra \( g^{(n)}_{\text{eff}} \), where Hebbian learning imprints frequency counts from local events \( \mathcal{E}_n \) directly into memory:
\[
g^{(n)}_{\text{eff}} = \sum_{\mathcal{E}_n^{\small\text{vocab}}} \tilde{v} v, 
\]
with one spectra per level $n$.

Random languages generated by our model exhibit rank-ordered \( n \)-gram frequency distributions that follow a scale-invariant Zipf-like law \cite{Gan&Wang&Han2009_NtupleZipf}.

At short lengths (\( n < n_{\text{peak}} \)), constraints are minimal, and the distribution appears structureless: a Zipf-like power-law head followed by a tail of irrelevant \( n \)-grams, similar to those found in uniform random strings \cite{Eugenio2025_HebbianLanguage,Gan&Wang&Han2009_NtupleZipf}---Fig \ref{Fig:uniform}. However, as \( n \) increases, the DAG constraint increasingly shapes memory formation. Beyond the peak, the tail vanishes and the power-law scaling persists, signaling the emergence of scale-invariance tied to a dominant compositional structure at longer lengths.

\subsubsection{Scale-Invariance}\hfill

\begin{figure}
    \centering
    \includegraphics[width=.99\linewidth]{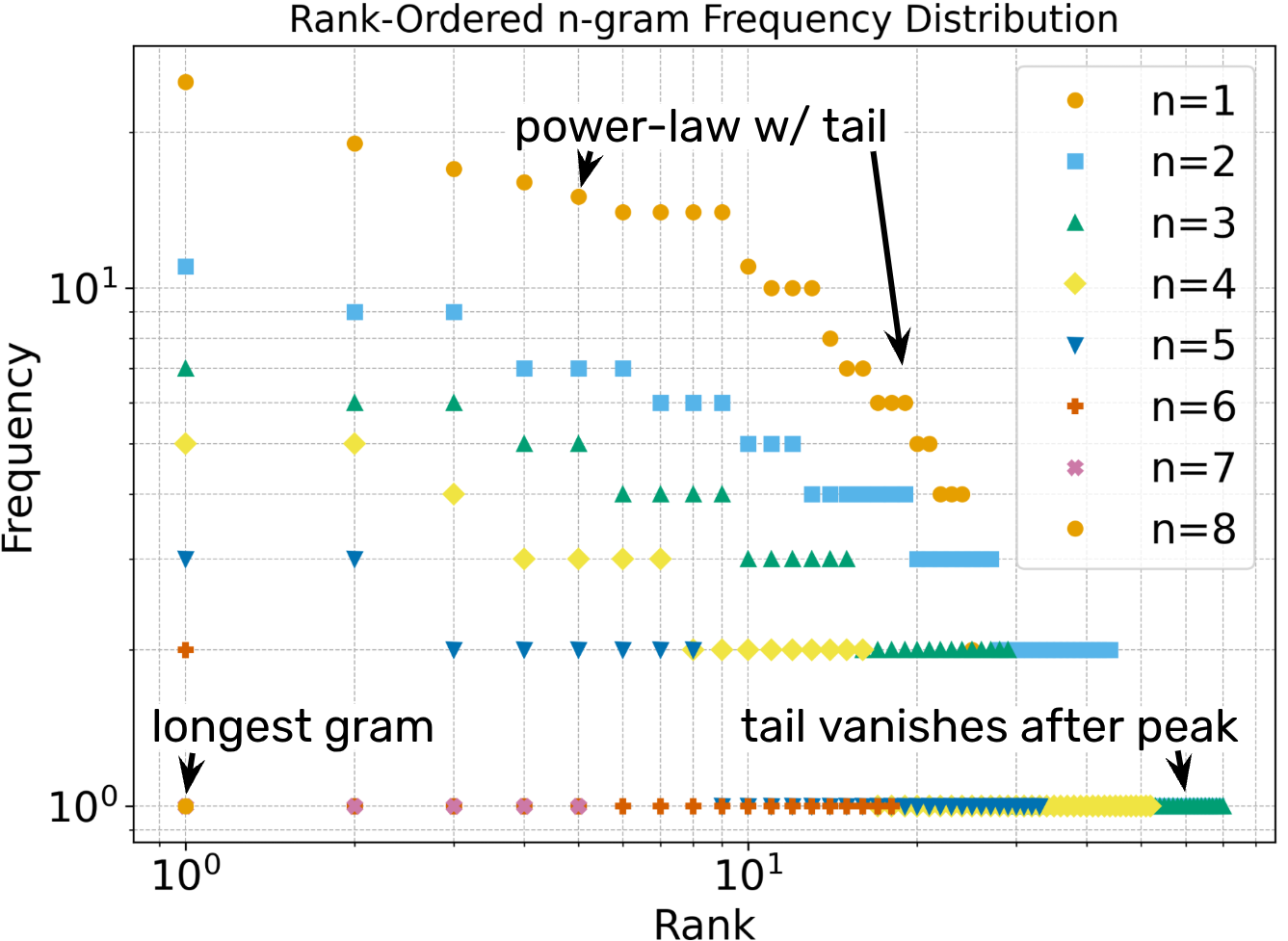}
    \caption{The frequency an $n$-gram (length $n$) appears in the vocabulary, rank-ordered from most to least frequent. The vocabulary is for the randomly grown language {\it bjfyi} (in supplement). Equivalently, this plot can be understood as representing an effective model spectra ($g^{(n)}_{\text{eff}}$ per $n$). Scale-invariance emerges beyond $n_{\text{peak}}=3$ due to the compositional structure.}
    \label{Fig:spectra_bjfyi}
\end{figure}

Note that the widest spectrum of Fig \ref{Fig:spectra_bjfyi} is $n=n_{\text{peak}}$ due to the rank ordering (largest dimension). The earlier layers (left of the $d_n$ peak) arise out of pure randomness, and initially grows exponentially. The DAG constraint grows stronger with increasing $n$, until $n_{\text{peak}}$, beyond which it becomes the dominate effect.

This gives rise to a {\bf characteristic length scale} $l_{\text{token}}$, which we here dub the {\bf invariant token length}: {\it Measures the distance past $n_{\text{peak}}$ to the last DAG overlap (where slope vanishes)}. Beyond which all autoregressive flows are guaranteed to converge to a single terminal node. 

Note that $l_{\text{token}}\simeq 0$ for noise. This is to say that that the more noisy the string, the less likely its overlap with other memories, and therefore the less its interference during retrieval from short-term memory. Memorizing noise requires learning all $n$-grams (so keep all bigrams), and can only be done for a limited length string, due to the unchecked scaling in the feature dimension. Finite $l_{\text{token}}$ indicates local structure, with a larger number of DAG overlaps, and hence requires lower overhead for a DAG memory. 

Note the distinction between scale invariance of a retrieving memory, and the scale invariance of composing subword patterns (the persistence of the power-law slopes). The latter arises from contributions from overlapping DAG graphs, whose slopes decrease as we approach longer length scales \& fewer overlaps. Once these slopes vanish, a single memory is guaranteed for a pattern of that length. Hence the vanishing of the slopes indicates increasing scale invariance during retrieval.


This latter point highlights the {\bf terminating} nature of human language patterns. If given a starting pattern (like {\bf ca}) and told to predict the next character in the string ({\bf ca}$\rightarrow${\bf cat}$\rightarrow${\bf catc}$\rightarrow\cdots$) we would find that the number of possible words overlapping decreases with the increasing length of the string. Eventually we run into a point ({\bf catches}) where we can no longer extend the string, as there are no longer valid patterns in memory. Any further prediction beyond the longest feature is necessarily random, or governed by the inter-word scale. This can be measured as a jump in entropy.

\subsubsection{Local Ordering, the Illusion of Utility, \& {\bf Inter-Generational Learning}}\hfill

Figure~\ref{Fig:greek_rosetta_stone} shows the rank-ordered frequency spectra for Greek words in the Rosetta Stone. Removing spaces (Fig.~\ref{Fig:greek_rosetta_stone}-b) does not destroy the compositional structure: the invariant token length $l_{\text{token}}$ remains finite. This indicates that word-bearing strings are {\bf locally ordered}: a restricted subset of short-range patterns dominates the statistics even when boundaries are erased. In other words, the local structure survives boundary removal, suggesting that underlying words can in principle be inferred from undifferentiated spaceless text.

This possibility motivates ``token-finding" approaches, where words are inferred directly from statistical regularities. Such methods have utility for parsing undeciphered strings or segmenting continuous speech. And it has been suggested that the lack of explicit boundary markers in human speech suggest children may exploit this during word learning \cite{Brent1999_SpeechSegmentation&WordDiscovery}. Modern statistical approaches similarly attempt unsupervised token discovery \cite{Anonymous2024_LatentSegmentLanguageModels,Pagnoni2024_ByteLatentTransformerPatches,Schmidt2025_BoundlessBytePairEncoding,Goldwater&Griffiths&Johnson2009_BayesianWordSegmentation,Creutz&Lagus2006_UnsupervisedMorphologyInduction_Morfessor}.

However, this behavior is \underline{not} a property of the retokenizer. The model cannot, in general, recover the ``correct" words from a spaceless string. While it can form symbolic tokens from unsegmented input, those tokens must first be demonstrated---that is, the model must learn what constitutes a ``good" pattern. Because it operates purely on local correlations, it is sensitive only to the frequency statistics of nearby patterns. The same local order can be mimicked by copy-\&-paste methods (Simon-like processes) for generating language-like strings \cite{Gan&Wang&Han2009_NtupleZipf}. There are no actual words composing those strings, but have a compositional structure due to copy-\&-pasting string segments to the end of the string.


If trained on such sequences, the retokenizer would simply invent its own symbols consistent with its inductive biases, just as it does when trained on noise.  
Even on real text containing concatenation errors, it may merge frequent sequences into composite tokens (e.g learning {\bf theboy} instead of {\bf the} and {\bf boy}).

Crucially, this local ordering is not a designed feature of human language, nor an adaptation to aid learners. It is a \emph{consequence of locality}---a universal property of any microscopic system composed of locally interacting parts. Any utilitarian structure is purely coincidental. Children do not learn ``correct" words but form internal interpretations aligned with their own inductive biases. If those biases resemble those of the retokenizer, then even ``incorrect" segmentations still preserve the universal subword statistics. {\bf Universal substructure is a dynamical fixed point of local learning}; it cannot collapse under continued relearning \cite{Shumailov&Gal2024nature_ModelCollapse,Shumailov2024_CurseOfRecursion}. Thus, linguistic change can be viewed as the reinterpretation of learned errors, continuously recycled into good language patterns by the same universal dynamics.

\section{The Benefits of Thinking About It}
\label{Sec:BenefitsOfThinking}


The retokenizer is not a very good model. It's bifix-free subregular language structure limits the complexity of strings it can represent. Nor is possible to repurpose it as a tokenizer (pre-processor) for an LLM. This would amount to discarding its hierarchical structure and collapsing all learned $n$-grams into a flat static vocabulary, which would scale poorly and break the model's inductive biases.

Its value is in its ability to explain where basic word structure comes from, and its relationship to symbolic order. Its word symbols are entirely meaningless, which teaches that us universal structure is independent to meaning. But this explanation is obviously incomplete.


So then: ``What is microscopically possible?" As a general statement, local memories are simple retrieval memories which capture the trace of an event. They aren't that coordinated, and can only do two things: retrieve or forget. Therefore, continued learning arises from the retrieval.


\subsection{The Replay Game (Model Thinking)}


Replay begins by selecting a random basis token \( v_k \) as seed. From this seed, the short-term memory network performs left and right inference using only the local energy gradients from \( g^{(n)} \). Inference proceeds recursively: the next token \( v \) is selected based on the energy gradient
\[
\Delta v_k = \sum_{n, \mu_{n-1}} g^{(n)}_{\mu_{n-1}, k} \, \tilde{v}^{(n-1)}_{\mu_{n-1}} ,
\]
and sampling \( \rho_k\big(f_k(\Delta v_k)\big) \). At each step, retokenization ensures that only smooth (i.e compositional) features are activated.

The process continues until the inferred string reaches a natural boundary — a region where no further smooth retokenization is possible. The result is a spontaneously generated word, such as:
\[
[v_{\bf c}, v_{\bf a}, v_{\bf t}, v_{\bf s}].
\]


At some point during this replay cycle — whether mid-replay or post-hoc — we introduce a new auxiliary neuron \( a_\alpha \) from an external source (conceptually, a pool at infinity). We activate it by construction (i.e we force it to fire).

As the replay proceeds, the features \( \tilde{v}^{(n)} \) that fire over time become locally Hebbian-correlated with the firing \( a_\alpha \). That is, at each replay step, the memory weight is updated by:
\[
m_{\alpha, \mu_n}^{(n)} \gets m_{\alpha, \mu_n}^{(n)} + \eta \, a_{\alpha}\tilde{v}_{\mu_n}^{(n)} ,
\]
where \( \eta \) is a learning rate. 

Alternatively — and equivalently — one can view the full word replay as producing a batch of active features, which are then bound in a single step to the new \( a_\alpha \) neuron. It encodes the feature content across all \( n \)-gram layers of the retokenization hierarchy:
\[
\text{e.g}\;\;\; a_{\alpha={\bf cats}} \leftrightarrow \left\{ v_{\bf c}, v_{\bf a}, v_{\bf t}, v_{\bf s}; \tilde{v}^{(2)}_{\bf ca}, \tilde{v}^{(2)}_{\bf at}, \tilde{v}^{(2)}_{\bf ts}; \tilde{v}^{(3)}_{\bf cat}, \tilde{v}^{(3)}_{\bf ats}; \tilde{v}^{(4)}_{\bf cats} \right\}.
\]
The result is a key-value memory binding \cite{Tyulmankov&all2021_BioKeyValue,Gersham&Fiete&Irie2025_KeyValueMemory,Ba&all2016_FastWeightsAssociativeKeyValue}: the neuron \( a_\alpha \) becomes a long-term embedding associated with the full replayed word. Over time, this mechanism constructs a library of {\bf long-term memory} neurons \( \{a_\alpha\} \), each associated with a unique word and its compositional features.

\subsubsection{\bf Non-static architecture, protected retrieval}

Long-term memories live in parallel channels; adding one does not overwrite others. It is a content-addressable memory, sharing qualities with sparse distributed memories \cite{Kanerva1988_SDM} \& dynamic associative networks \cite{Beavers2010_DAN1,Beavers&Harrison_DAN2}. {\bf Memory capacity is infinite}, as tokens can be stored onto the disk until needed. Nothing is ever overwritten, and thus no catastrophic forgetting.

The parallelization provides topological protection (replay or recognition of one memory does not interfere with another). Thus can retrieve \& accept {\bf hyper} independent from {\bf hyperbolic}. 

Because the memory is local, each token decomposes into a series of small tensor contractions. This supports staged (\& partial\footnote{Partial-key checks: You don't need to compute the whole token contribution, only up to the part where you can be certain no other token is more similar. This is similar to how some subword patterns like {\bf ndwi} necessarily only converge to {\bf sandwich}---a length beyond which retrieval is scale-invariant--Fig \ref{Fig:ScaleInvFlow}.}) key checks for efficient lookup on serial hardware, while an ideal neuromorphic device parallelizes the entire search.


\subsubsection{\bf Sparse not dense}

Our model is actually sparse---we just use a convenient dense representation. This produces redundancy in the form of gauges, which we exploit to perform right \& left token predictions. Basis tokens are stored locally as one-hot encodings, not random memory vectors. Structure emerges from local projector compression (a consequence of {\it locality}); and the distributed nature of the memory post-replay arises due to {\it simultaneity} of those replayed features. Thus structure can be learned without any data\footnote{Training a model, even a local model, with explicit language data is technically a non-local process. The specific ordered strings of that data were prepared by a non-local source---the human modeler. It would not have learned the word ``cat" had you not ordered the data.} or human intervention.

This is completely unlike dense distributed memories, which typically assume a basis of random memory vectors in order to reduce overlap \cite{Hopfield1982,HopfieldAllNeed_Ramsauer2021,LongSequenceHopfield_Hamza2023}. There are no binding/unbinding operations as in Smolensky's tensor product formulation \cite{Smolensky1990}.

\subsubsection{\bf Topology before meaning}

The retokenizer operates over \emph{meaningless} topological symbols (bifix-free, subregular).  
This yields \emph{exact, boundary-aware} retrieval paths: a terminal node is reachable only through its smooth subpaths, so inferred boundaries are statistical fixed points of the local dynamics rather than externally imposed marks. This guarantees symmetry between morphology and syntax---both arising from correlations encoding subregular strings of the retokenizer. The tree structure of the embedding graphs are identical to syntax trees.

\paragraph{Benefits of Replay Relearning}
\begin{itemize}
  \item \textbf{Compression:} Word features are projected onto low-rank memory vectors.
  \item \textbf{Decoupling:} Embeddings allow hierarchical memory $g^{(n)}$ to be forgotten or reused.
  \item \textbf{Continual Learning:} New embeddings can be learned without overwriting existing ones.
  \item \textbf{Parallelism:} Each $a_\alpha$ contributes independently to inference. Allows for partial staged-key checks.
  \item \textbf{Emergent Computation:} The $a_\alpha$ contribution to inference generalizes beyond local \( n \)-gram statistics. The token complexity evolves with continuous learning.
\end{itemize}

\begin{figure}[h!]
\centering
\begin{subfigure}[t]{0.5\textwidth}
    \centering
    \includegraphics[width=.95\linewidth]{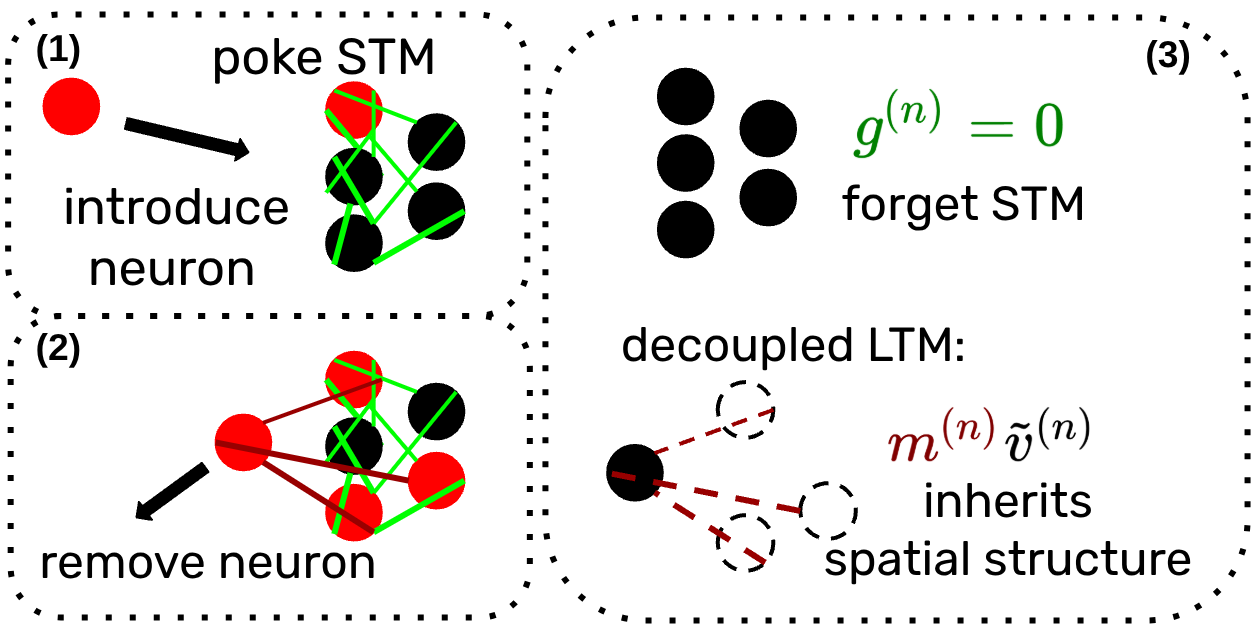}
    \caption{The dynamic game of ``replay": (1) Randomly activate node in short-term memory (STM). (2) Introduce new neuron until hierarchy converges on stored memory, then remove it. (3) Added neurons emerge as long-term memory (LTM).\\}
    \label{Fig:plasticity}
\end{subfigure}\\
\begin{subfigure}[t]{0.5\textwidth}
    \centering
    \includegraphics[width=.95\linewidth]{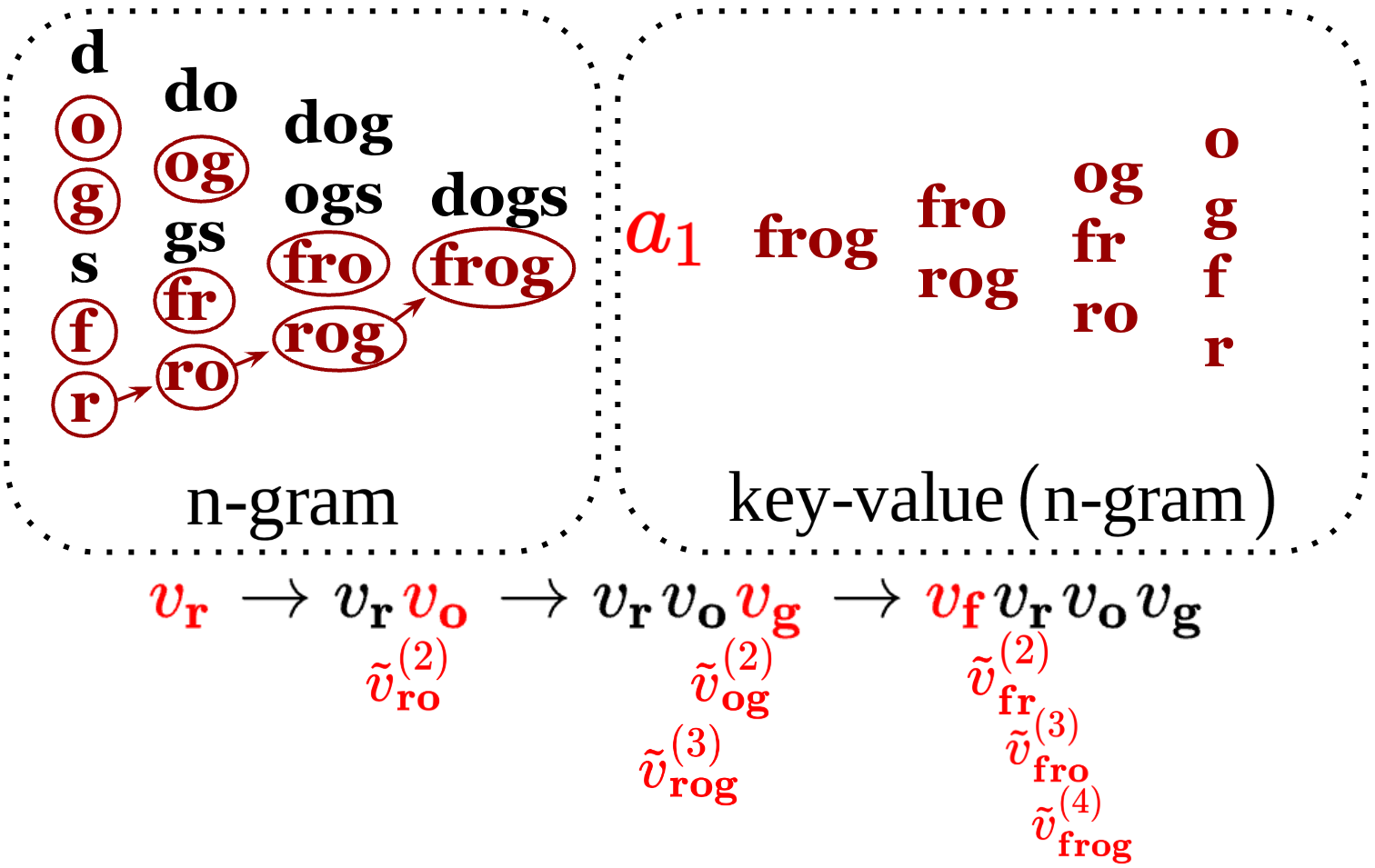}
    \caption{LEFT: State-transitions of the STM converge on terminal node. Red shows features activated during replay. RIGHT: Embedding neuron ($a_1$) learns an emergent {\it hierarchical key-value memory}, which is an attention mechanism wrapped around a word-specific n-gram model. BOTTOM: Replay as autoregression with boundary awareness. Red shows activated trailing context tokens.}
    \label{Fig:replay}
\end{subfigure}
\caption{Dynamically Coding the Local Plastic Structure}
\label{Fig:plasticity&replay}
\end{figure}

\section{Emergent Hierarchical Key-Value Memory (Recognition)}
\label{Sec:EmergentKVmemory}

Inference proceeds as a sequence of discrete event updates. At each time step, a new token is predicted based on the trailing context through the structure stored in both short-term ($g^{(n)}$) and long-term ($m^{(n)}$) memory. The full Hamiltonian is therefore 
\begin{eqnarray}\label{Eqn:H_amv}
H = \sum\limits_n\sum_{\mu_{n-1},k} g_{\mu_{n-1},k}^{(n)}\tilde{v}_{\mu_{n-1}}^{(n-1)}v_{k} 
+\sum_{\alpha} a_{\alpha}\Big(\sum\limits_n\sum_{\mu_{n}} m_{\alpha,\mu_n}^{(n)}\tilde{v}_{\mu_n}^{(n)}\Big)  . 
\end{eqnarray}

\subsection{Update Logic}

Each token prediction is based on evaluating local gradients:
\begin{equation}
Q \leftarrow f_Q\left( \Delta Q \right),
\qquad \Delta Q := \frac{\partial H}{\partial Q} ,
\end{equation}
for $Q \in \{v_k,\ a_\alpha\}$, where \( f_Q \) is any bounded activation function.

We choose to update $a_\alpha$ neurons first. Their updated values are then used to compute the update for $v_k$. The corresponding gradients are:
\begin{align}
\Delta a_{\alpha}
&= \sum_n \sum_{\mu_n} m_{\alpha,\mu_n}^{(n)} \tilde{v}_{\mu_n}^{(n)} , \\
\Delta v_{k} 
&= \sum_n \sum_{\mu_{n-1}} 
\left(
g^{(n)}_{\mu_{n-1},k}
+ \sum_\alpha a_\alpha \, m^{(n)}_{\alpha,\mu_{n-1},k}
\right)
\tilde{v}^{(n-1)}_{\mu_{n-1}} .
\end{align}

\subsection{Key-Value Memory via Fast Embedding Neurons}

If $a_\alpha$ neurons respond more rapidly than $v_k$, we treat them as reaching a fixed point first \cite{DenseHierarchyHopfield_Krotov2021}:
\begin{equation}\label{Eqn:a=f(mv)}
a_\alpha = f_\alpha\left( \sum_{n} \sum_{\mu_n} m_{\alpha, \mu_n}^{(n)} \tilde{v}_{\mu_n}^{(n)} \right) .
\end{equation}
Plugging into the equation for $\Delta v_k$ yields:
\begin{align}
\Delta v_k &= \sum_n \sum_{\mu_{n-1}} g^{(n)}_{\mu_{n-1},k} \, \tilde{v}^{(n-1)}_{\mu_{n-1}} \\
&\quad + \sum_\alpha f_\alpha\left( \sum_{n'} \sum_{\mu_{n'}'} m_{\alpha, \mu_{n'}'}^{(n')} \tilde{v}_{\mu_{n'}'}^{(n')} \right)
\cdot \sum_n \sum_{\mu_{n-1}} m_{\alpha, \mu_{n-1}, k}^{(n)} \tilde{v}^{(n-1)}_{\mu_{n-1}} . \label{Eqn:kv_update}
\end{align}

Eqn \ref{Eqn:kv_update} defines a \textbf{hierarchical key-value memory}, where:
\begin{itemize}
    \item The \emph{key} is the set of smooth projected context tokens \( \tilde{v}^{(n)} \),
    \item The \emph{embedding neuron} \( a_\alpha \) is selected by the context via \( f_\alpha(\cdot) \),
    \item The \emph{value} is the distribution over next tokens \( v_k \), computed via the activated embedding.
\end{itemize}
This emergent key-value memory mechanism allows long-term associations to be built from local dynamics alone, forming the basis for structured retrieval and generalization in later inference. Thus the model can generalize beyond local \( n \)-gram statistics, enabling long-range coherence, partial match inference, and efficient lookup. We refer to this process as \emph{recognition}\footnote{Note that {\it recognition} can interfere with {\it tokenization}. For instance, the repeated string formed from {\bf monk}$+${\bf eye}, i.e {\bf ...onkeyemonkeyemon...}, uniquely tokenizes back into the two words stated, however, the reader may still recognize {\bf monkey} or {\bf key} even though they don't uniquely tokenize the string, highlighting the competition between fast long-term memory effects with slower tokenizing short-term memory \cite{Eugenio2025_HebbianLanguage}.}.

\subsubsection{Key-Value Memory Example}

Consider Eqn \ref{Eqn:kv_update} with $g^{(n)} = 0$, s.t inference is driven purely by $a_\alpha$. Suppose we have two learned word embeddings, $\alpha \in \{1,2\}$, corresponding to:
\[
a_1 \leftrightarrow \text{{\bf cats}},\qquad a_2 \leftrightarrow \text{{\bf dogs}}.
\]

Then the update becomes:
\begin{align}
&\Delta v_k = \nonumber\\
&f_1\left( 
\tilde{v}_{\bf ca}^{(2)} m^{(2)}_{1,{\bf ca}} 
+ \tilde{v}_{\bf at}^{(2)} m^{(2)}_{1,{\bf at}} 
+ \tilde{v}_{\bf cat}^{(3)} m^{(3)}_{1,{\bf cat}} 
+ \tilde{v}_{\bf ats}^{(3)} m^{(3)}_{1,{\bf ats}} 
+ \tilde{v}_{\bf cats}^{(4)} m^{(4)}_{1,{\bf cats}} 
\right) \nonumber \\
&\quad \times \left( 
\tilde{v}_{\bf ca}^{(2)} m^{(3)}_{1,{\bf ca},k} 
+ \tilde{v}_{\bf at}^{(2)} m^{(3)}_{1,{\bf at},k} 
+ \tilde{v}_{\bf cat}^{(3)} m^{(4)}_{1,{\bf cat},k} 
\right) \nonumber \\
&+ f_2\left( 
\tilde{v}_{\bf do}^{(2)} m^{(2)}_{2,{\bf do}} 
+ \tilde{v}_{\bf og}^{(2)} m^{(2)}_{2,{\bf og}} 
+ \tilde{v}_{\bf dog}^{(3)} m^{(3)}_{2,{\bf dog}} 
+ \tilde{v}_{\bf ogs}^{(3)} m^{(3)}_{2,{\bf ogs}} 
+ \tilde{v}_{\bf dogs}^{(4)} m^{(4)}_{2,{\bf dogs}} 
\right) \nonumber \\
&\quad \times \left( 
\tilde{v}_{\bf do}^{(2)} m^{(3)}_{2,{\bf do},k} 
+ \tilde{v}_{\bf og}^{(2)} m^{(3)}_{2,{\bf og},k} 
+ \tilde{v}_{\bf dog}^{(3)} m^{(4)}_{2,{\bf dog},k} 
\right).
\end{align}
Note that generically $m^{(1)}_{\alpha,k}$ (monogram) terms also contribute, but we suppress them in this example to preserve space.

If the context consists of $v_{\bf a}, v_{\bf t}$, then $\tilde{v}_{\bf at}^{(2)} = 1$ and all other projections vanish. The update reduces to:
\[
\Delta v_k = f_1\big(m^{(2)}_{1,{\bf at}}\big) \cdot m^{(3)}_{1,{\bf at},k} .
\]
This predicts the next token $v_k$ with the strongest overlap with ${\bf s}$, due to the binding $m^{(3)}_{1,{\bf at},k} = m^{(3)}_{1,{\bf ats}} \, \delta_{k,{\bf s}}$.

Note that the update of $a_{\alpha}$ inherits information about morphological structure (the projection maps) from its interaction with the $\tilde{v}^{(n)}$. Thus, smooth features in the context trigger the appropriate word embedding via $f_\alpha(\cdot)$, which then drives prediction through projection maps. Thus there is no need to further encode positional relationships, which is handled by the projectors. We discuss how this inheritance allows optimizing for word-specific projection maps in the following section.

\subsection{Compression via Word-Specific Projector Decomposition}

To illustrate how replay leads to compression, we begin with a concrete example. Suppose the model has learned the following projectors:

\vspace{0.5em}
\noindent
\textbf{Bigrams:}
\begin{equation}
\begin{aligned}
P_2 =\ & \tilde{v}^{(2)}_{\bf ca} v_{\bf c} v_{\bf a} + \tilde{v}^{(2)}_{\bf at} v_{\bf a} v_{\bf t} + \tilde{v}^{(2)}_{\bf ts} v_{\bf t} v_{\bf s} \\
&+ \tilde{v}^{(2)}_{\bf do} v_{\bf d} v_{\bf o} + \tilde{v}^{(2)}_{\bf og} v_{\bf o} v_{\bf g} + \tilde{v}^{(2)}_{\bf gs} v_{\bf g} v_{\bf s}
\end{aligned}
\end{equation}

\noindent
\textbf{Trigrams:}
\begin{equation}
\begin{aligned}
P_3 =\ & \tilde{v}^{(3)}_{\bf cat}\tilde{v}^{(2)}_{\bf ca}v_{\bf t} + \tilde{v}^{(3)}_{\bf ats}\tilde{v}^{(2)}_{\bf at}v_{\bf s} \\
&+ \tilde{v}^{(3)}_{\bf dog}\tilde{v}^{(2)}_{\bf do}v_{\bf g} + \tilde{v}^{(3)}_{\bf ogs}\tilde{v}^{(2)}_{\bf og}v_{\bf s}
\end{aligned}
\end{equation}

\noindent
\textbf{4-grams:}
\begin{equation}
P_4 = \tilde{v}^{(4)}_{\bf cats}\tilde{v}^{(3)}_{\bf cat}v_{\bf s} + \tilde{v}^{(4)}_{\bf dogs}\tilde{v}^{(3)}_{\bf dog}v_{\bf s}
\end{equation}

Now consider the replay-imprinted $m$-tensors for $\alpha = {\bf cats}$:
\begin{equation}
\begin{aligned}
m^{(2)}_{\bf cats} &= \tilde{v}^{(2)}_{\bf ca} + \tilde{v}^{(2)}_{\bf at} + \tilde{v}^{(2)}_{\bf ts}, \\
m^{(3)}_{\bf cats} &= \tilde{v}^{(3)}_{\bf cat} + \tilde{v}^{(3)}_{\bf ats}, \\
m^{(4)}_{\bf cats} &= \tilde{v}^{(4)}_{\bf cats}.
\end{aligned}
\end{equation}

We now demonstrate compression by contracting $m^{(3)}_{\bf cats}$ with the entangled projector $P_3$:
\begin{equation}
\begin{aligned}
m^{(3)}_{\bf cats} P_3 &= (\tilde{v}^{(3)}_{\bf cat} + \tilde{v}^{(3)}_{\bf ats}) (\tilde{v}^{(3)}_{\bf cat} \tilde{v}^{(2)}_{\bf ca} v_{\bf t} + \tilde{v}^{(3)}_{\bf ats} \tilde{v}^{(2)}_{\bf at} v_{\bf s} \\
&\qquad + \tilde{v}^{(3)}_{\bf dog} \tilde{v}^{(2)}_{\bf do} v_{\bf g} + \tilde{v}^{(3)}_{\bf ogs} \tilde{v}^{(2)}_{\bf og} v_{\bf s}) \\
&= \tilde{v}^{(2)}_{\bf ca} v_{\bf t} + \tilde{v}^{(2)}_{\bf at} v_{\bf s},
\end{aligned}
\end{equation}
where orthogonality eliminates the contributions from tokens unrelated to {\bf cats}. The result is a disentangled, word-specific tensor chain:
\begin{equation}
= (\tilde{v}^{(3)}_{\bf cat} + \tilde{v}^{(3)}_{\bf ats}) (\tilde{v}^{(3)}_{\bf cat} \tilde{v}^{(2)}_{\bf ca} v_{\bf t} + \tilde{v}^{(3)}_{\bf ats} \tilde{v}^{(2)}_{\bf at} v_{\bf s}) 
:= \hat{m}^{(3)}_{\bf cats} \tilde{P}^{\bf cats}_3.
\end{equation}

This procedure can be recursively applied to the full projector chain $P_3 P_2$ to yield a compressed set of projectors $\hat{P}_n^{\alpha}$ for each word $\alpha$. These retain only the subspaces relevant to $\alpha$, enabling scalable and disentangled memory representations.

\medskip
Next, we generalize this process to arbitrary word embeddings using SVD-based decomposition of the $m$-capped projection chains.

\subsection{Generalized Compression via Projector Contraction}\label{Sec:Compression}

The concrete example above illustrates how word-specific projector chains can be disentangled by pruning irrelevant branches. We now formalize this procedure using a sequence of algebraic contractions and SVDs \cite{mps_Schollwock2011,mps_Parker&Cao&Zaletel2020,mps_Gourianov&all2022_turbulence}, resulting in a compressed representation of the hierarchical projector chain for each word embedding.

We begin with the full replay expression for a given word embedding $a_{\alpha}$:
\begin{equation}
\sum_{\mu_n} m^{(n)}_{\alpha, \mu_n} \, \tilde{v}^{(n)}_{\mu_n}
= \sum_{\mu_n, \mu_{n-1}, \dots, \mu_2} m^{(n)}_{\alpha, \mu_n} P_n^{\mu_n, \mu_{n-1}, j_n} \cdots P_2^{\mu_2, j_1, j_2} \, v_{j_1} v_{j_2} \cdots v_{j_n} .
\label{eq:replay_expansion}
\end{equation}

This expression can be interpreted as a long chain of projection maps contracted with the memory vector $m^{(n)}_{\alpha}$ at the top. In practice, many components of $m^{(n)}_{\alpha}$ will be zero, as the embedding stores only features observed during the replay of word $\alpha$.

To compress this representation, we apply a sequence of SVD decompositions layer by layer. First, we perform an SVD of the capped tensor:
\begin{equation}
m^{(n)}_{\alpha, \mu_n} = \sum_{\beta_n} \hat{m}^{(n)}_{\alpha, \beta_n} \, V_n^{\beta_n, \mu_n} ,
\end{equation}
where $V_n$ defines a word-specific low-rank subspace over $\mu_n$, and $\hat{m}^{(n)}_{\alpha, \beta_n}$ is the compressed top cap.

Next, we contract $V_n$ with $P_n$:
\begin{equation}
\tilde{P}_n^{\beta_n, \mu_{n-1}, j_n} = \sum_{\mu_n} V_n^{\beta_n, \mu_n} P_n^{\mu_n, \mu_{n-1}, j_n} .
\end{equation}
This new projector has reduced dimension, containing only $n$-grams relevant to $a_{\alpha}$. We then reshape this tensor into a matrix over index pairs $(\mu_{n-1}, j_n)$ and perform another SVD: $\tilde{P}_n^{\beta_n, (\mu_{n-1}, j_n)} = \sum_{\omega} U_n^{\beta_n, \omega} \, D_n^{\omega} \, V_n^{\omega, (\mu_{n-1}, j_n)}$.

We now proceed to compress the next layer. We contract the decomposed $V$ from above with the next projector:
\begin{equation}
T_{\omega, j_n, \mu_{n-2}, j_{n-1}} = \sum_{\mu_{n-1}} V_n^{\omega, \mu_{n-1}, j_n} \, P_{n-1}^{\mu_{n-1}, \mu_{n-2}, j_{n-1}} .
\end{equation}
Merging the left and right index pairs, we perform another decomposition: $T_{(\omega j_n), (\mu_{n-2} j_{n-1})} = \sum_{\beta_{n-1}} U'_{(\omega j_n), \beta_{n-1}} D'_{\beta_{n-1}} V'_{\beta_{n-1}, (\mu_{n-2}, j_{n-1})}$.
We now define the fully compressed projector:
\begin{align}
\hat{P}_n^{\beta_n, \beta_{n-1}, j_n} &= \sum_{\omega} U_n^{\beta_n, \omega} \, U'_{\omega, j_n, \beta_{n-1}} \, D'_{\beta_{n-1}} , \\
\tilde{P}_{n-1}^{\beta_{n-1}, \mu_{n-2}, j_{n-1}} &= V'_{\beta_{n-1}, \mu_{n-2}, j_{n-1}} .
\end{align}

We repeat this contraction-decomposition process recursively, defining new compressed projection maps $\hat{P}_n$ and reduced intermediate indices $\beta_{n-1}, \beta_{n-2}, \dots$ at each step.

The final expression for the replayed memory becomes:
\begin{equation}
\sum_{\beta_n, \dots, \beta_2} \hat{m}^{(n)}_{\alpha, \beta_n} \, \hat{P}_n^{\beta_n, \beta_{n-1}, j_n} \cdots \hat{P}_2^{\beta_2, j_1, j_2} \, v_{j_1} v_{j_2} \cdots v_{j_n} .
\end{equation}
Because each projector has its own word ($\alpha$) \& length ($n$) specific indices, it's more correct to write $\hat{P}_{n,n'}^{(\alpha)}$. Each $\hat{m}^{(n)}$-capped chain having their own set of projectors: 
\begin{eqnarray}
\hat{m}^{(n)}_{\alpha}\hat{P}_{n,n}^{(\alpha)}\hat{P}_{n,n-1}^{(\alpha)}\cdots\hat{P}_{n,2}^{(\alpha)} .\notag
\end{eqnarray}

This compressed chain retains only the information needed to reconstruct the replay trajectory of word $\alpha$, eliminating all unrelated feature branches. The compression is exact if no finite SVD eigenvalues are discarded.

\paragraph{Benefits.} This word-specific decomposition achieves three goals:
\begin{itemize}
  \item \textbf{Disentanglement:} Removes unrelated replay trajectories (an exact compression).
  \item \textbf{Compression:} Optimizes the dimension of projection maps.
  \item \textbf{Finer Staged-Key Searches:} Reduces the overhead of content addressable memory retrieval.
\end{itemize}

\hfill\\
{\it Warning}: One drawback of this compression is that the resulting projectors no longer form a gauge-invariant set. Consequently, it's not possible to regauge between right/left tokenized forms. This happens because the compression preserves only those DAG subpaths related to merging tokens on the right (i.e $\tilde{v}^{(n-1)}v\rightarrow\tilde{v}^{(n)}$). 

This is still an exact compression (keep all SVD eigenvalues), because not all subpaths are being used to project up onto a given $m^{(n)}$---squeezing out only the essential part of the dense representation. The {\it key} part of the key-value mechanism remains unaffected; only the redundancy related to extra freedom in how the {\it value} part decodes its latent information is discarded. The exact symbolic behavior \& boundary information remain. And the entire string can still be decoded exactly from only right-token prediction. Thus it is possible to move back-\&-forth between an $\alpha$-specific gauge-invariant projector set $\{P_n^{(\alpha)}\}$ \& the hyper-compressed $\{\hat{P}_{n,n'}^{(\alpha)}\}$.

\subsection{Super-Hierarchical Key-Value Memory}\label{Sec:SuperHierarchicalKV}
The hierarchical key-value memory mechanism naturally extends to higher levels of abstraction. Just as short-term memory tokens $\tilde{v}^{(n)}$ are bound into long-term word embeddings $a_\alpha$ via replay, the embeddings themselves may form higher-order structures, such as phrases, idioms, or conceptual clusters. We introduce a second memory hierarchy built on projected embedding neurons $\tilde{a}^{(m)}$, with feature maps $\mathcal{P}_m$, correlation weights $G^{(m)}$, and memory tensors $M^{(m)}$. This forms a \emph{super-hierarchical memory}, enabling compositional inference over learned embedding structures.

To support compositional memory beyond single-word representations, we extend the Hamiltonian to include higher-level interactions among embedding neurons:
\begin{align}
H &= \sum_n \sum_{\mu_{n-1},k} g^{(n)}_{\mu_{n-1},k} \tilde{v}^{(n-1)}_{\mu_{n-1}} v_k 
+ \sum_\alpha a_\alpha \left( \sum_n \sum_{\mu_n} m^{(n)}_{\alpha,\mu_n} \tilde{v}^{(n)}_{\mu_n} \right) \notag\\
&\quad + \sum_m \sum_{\alpha_{m-1}, \alpha} G^{(m)}_{\alpha_{m-1},\alpha} \tilde{a}^{(m-1)}_{\alpha_{m-1}} a_\alpha 
+ \sum_\xi b_\xi \left( \sum_m \sum_{\alpha_m} M^{(m)}_{\xi,\alpha_m} \tilde{a}^{(m)}_{\alpha_m} \right),
\end{align}
where $a_\alpha$ are embedding neurons and $b_\xi$ are super-embedding neurons. Feature projections $\tilde{v}^{(n)}$ and $\tilde{a}^{(m)}$ are defined via retokenization hierarchies $P_n$ and $\mathcal{P}_m$ respectively.

\paragraph{Update Logic:} We compute gradients
\begin{align}
\Delta v_k &= \frac{\partial H}{\partial v_k}, \quad 
\Delta a_\alpha = \frac{\partial H}{\partial a_\alpha}, \quad 
\Delta b_\xi = \frac{\partial H}{\partial b_\xi},
\end{align}
and apply bounded updates: $Q \leftarrow f_Q(\Delta Q)$ for $Q \in \{v_k,\ a_\alpha,\ b_\xi\}$. We treat $b_{\xi}$ \& $a_{\alpha}$ as being faster than $v_k$ (i.e hierarchy of equilibriation times $\tau_b \ll \tau_a \ll \tau_v$ \cite{HopfieldNeurobiology_Krotov&Hopfield2021,DenseHierarchyHopfield_Krotov2021}). Thus we solve for $b_\xi$ and $a_\alpha$ first, then substitute into $\Delta v_k$.

\paragraph{Inference Equation:}
Letting $g^{(n)} = G^{(m)} = 0$, we derive the inference rule:
\begin{multline}
\Delta v_k = \sum_\alpha 
f_\alpha\left( \sum_{n'} \sum_{\mu_{n'}'} m^{(n')}_{\alpha,\mu_{n'}'} \tilde{v}^{(n')}_{\mu_{n'}'} 
+ \sum_\xi f_\xi\left( \sum_{m'} \sum_{\alpha_{m'}'} M^{(m')}_{\xi,\alpha_{m'}'} \tilde{a}^{(m')}_{\alpha_{m'}'} \right)
\right. \\
\left.
\times \sum_{\alpha_{m-1}} M^{(m)}_{\xi,\alpha_{m-1},\alpha} \tilde{a}^{(m-1)}_{\alpha_{m-1}} 
\right)
\sum_n \sum_{\mu_{n-1}} m^{(n)}_{\alpha,\mu_{n-1},k} \tilde{v}^{(n-1)}_{\mu_{n-1}}.
\end{multline}

This defines a \emph{super-hierarchical key-value memory}: the context selects a super-embedding $b_\xi$, which activates relevant embeddings $a_\alpha$, which in turn retrieve fine-grained retokenized features for predicting $v_k$. \\

\subsubsection{{\bf Plasticity \& Emergence}}

The emergent key-value memories discussed here are a consequence of the plastic nature of Hebbian learning, combined with the hierarchical tokenizing inference. These new mechanisms are not put in by hand -- there is no ``key-value memory layer" added to the system -- and no retraining of the entire model is necessary. Nor is explicit structured data required. Instead, introducing a single new neuron to the system is sufficient in order for them to arise, demonstrating that they are truly {\it emergent} structures.

This type of plasticity is not present in models which learn via optimizing a global objective (back propagation). While such models are typically modular, their architecture is fixed beforehand. Though the black box paradigm permits a type of data-driven ``emergence", it is not the type of structural emergence described here, but instead reflects a discovery of internal representations allowed within the fixed architecture. The attention mechanism of an LLM is {\it assumed} beforehand \cite{Vaswani2018_AttentionAllYouNeed}, not a learned structured. This differs from Conway's Game of Life, which can ``learn" a Turing machine -- not from recognizing patterns in data nor minimizing an objective, but due to the local dynamics \& initial conditions \cite{PaulRendell2011_ConwayTuringComplete,PaulRendell2014thesis_ConwayTuringComplete}.

It's useful to think of replay as a minimally structured game (or a local dynamic coding scheme) -- grab a new neuron, poke short-term memory to randomly replay its correlations, then repeat the process again with a new neuron once replay terminates (Fig \ref{Fig:plasticity}). While structured, what information is replayed is entirely random, and thus no non-local dependence of specific information stored in those memories is required. This should not be confused with ``replay" as defined elsewhere in the machine learning literature, which involves replaying specific memories in order to prevent their forgetting in a fixed architecture \cite{Parisi2019review_Forgetting&Replay}. Here replay generates new feature basis in a non-static architecture---nothing is ever written over, as the memory capacity is unbounded, and thus there is no catastrophic forgetting.

\begin{figure}
    \centering
    \includegraphics[width=0.95\linewidth]{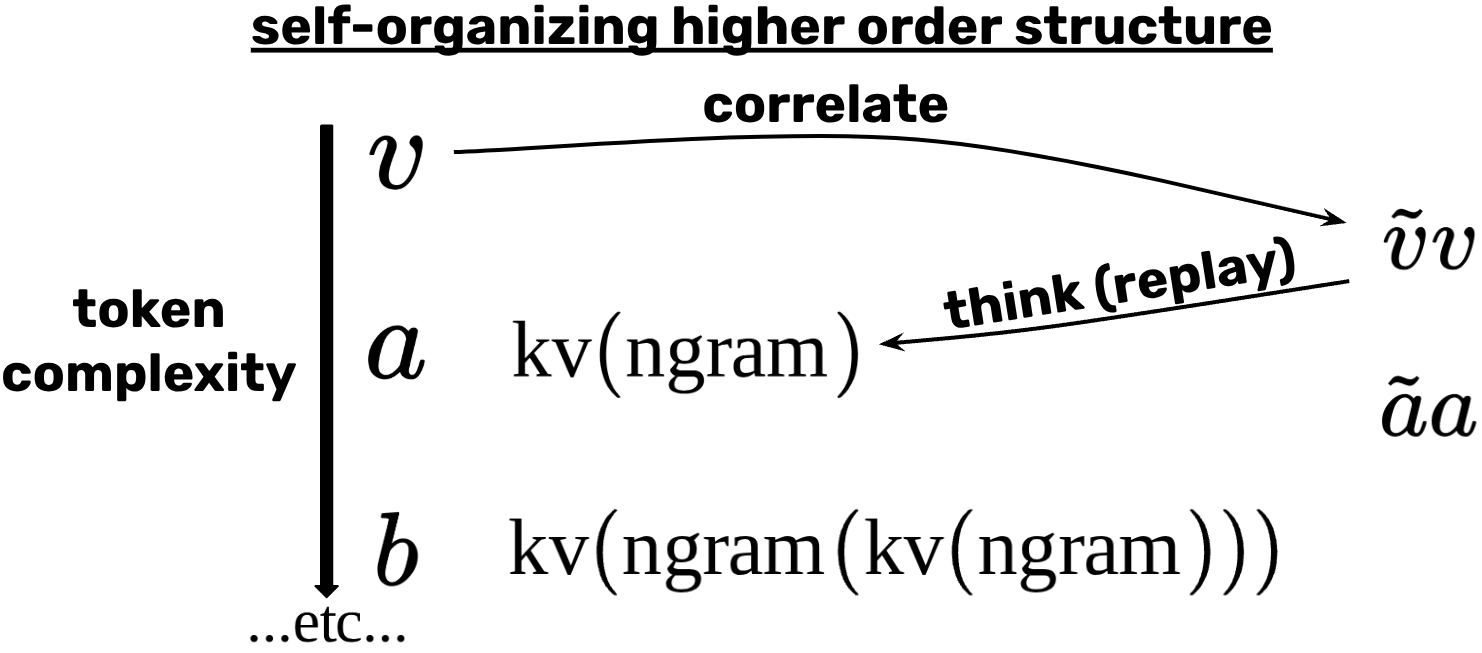}
    \caption{Hierarchies of hierarchies: Short-term memory structures are retokenizing n-gram models which grow feature tokens ($\tilde{v}$,$\tilde{a}$,$\tilde{b}$,...) from relevant correlations (products of tokens: $\tilde{v}v$,$\tilde{a}a$,...). These can be replayed into symbolic object tokens of increasing complexity ($a$,$b$,...). The simplest object tokens (first-order embeddings) decode as a key-value memory (temporal association via {\it simultaneity}) wrapping an n-gram model (captures {\it spatial} structure). Not shown: mixed correlations (e.g $ab+ba$) can also be replayed into tokens.}
    \label{Fig:hierarchyhierarchy}
\end{figure}

\section{Simplifications \& Code}\label{Sec:Simpler&Code}

\subsection{One-memory-per-retokenizer limit}

Up to this point, we have treated the problem as one of: grow short-term memory, then disentangle it into individual long-term memory tokens. However, if the boundaries in the data are known, then its possible to skip the disentanglement step. 

In this regime, each retokenizer is trained on a single input pattern, such as one word or sequence. The local Hebbian updates (Eqn \ref{Eqn:HebbUpdate}) therefore construct a minimal hierarchical graph whose projectors \( \{P_n\} \) collectively describe a single smooth path terminating in one node of the DAG. All sub-grams of the input are locally merged according to the smoothness constraint, yielding an internally consistent projector chain \( P_N P_{N-1} \cdots P_2 \) representing that single stored memory. 

The corresponding embedding is then formed directly by association (i.e direct sum) following Eqn \ref{Eqn:a=f(mv)}. For the rest of the paper, we will simplify our notation, and simply write the embeddings like 
\begin{eqnarray}\label{Eqn:a_cat}
a_{\bf cat} = (v_{\bf c}+v_{\bf a}+v_{\bf t})\oplus(\tilde{v}^{(2)}_{\bf ca}+\tilde{v}_{\bf at}^{(2)})\oplus\tilde{v}_{\bf cat}^{(3)} ,
\end{eqnarray}
where we have dropped explicit weight dependence ($m^{(n)}_{\alpha,\mu_n}$) and hidden the activation function ($f$). This makes explicit the individual DAG structure formed by the feature tokens. Memory is topological, thus specific weight information isn't important. We will simply assume that $m^{(n)}$ is sufficiently larger than $m^{(n-1)}$ to guarantee the terminal node is the autoregressive fixed point.

\subsection{Boundaries are data}

Note that higher-order embeddings (Sec \ref{Sec:SuperHierarchicalKV}) can be constructed similarly. For example, 
\begin{eqnarray}
&&b_{\bf the\ cat\ jumps} = \\
&& (a_{\bf the}+a_{\bf cat}+a_{\bf jumps})\oplus(\tilde{a}^{(2)}_{\bf the\ cat}+\tilde{a}_{\bf cat\ jumps}^{(2)})\oplus\tilde{a}_{\bf the\ cat\ jumps}^{(3)} \notag
\end{eqnarray}
Again, this requires specifying the statistical boundaries, such as specifically preparing small phrases or sentences. Thus there is an arbitrariness in learning from structured data, namely the structure itself: boundaries are data. This includes boundaries not necessarily indicated by notation in writing. If not made explicit, the model will construct its own interpretation of the boundaries, which may not align with the intended behavior, but will align with human language universals. \\


Note that the {\bf coordination length grows exponentially} with a linear increase in token depth (how many times high-order tokens were correlated \& replayed to build yet even higher-order tokens).

\subsection{Code \& Latent Space Topology}

Self-organizing neural learning is {\bf inherently multi-modal}: learning new embeddings expands it computational substrate. So far we have used \(v\) for external or input modalities and (\(a,b,\ldots\)) for internally generated ones. By default, all tokens are treated as being part of a giant undifferentiated content-addressable library. Thus for simplicity, we will unify internal tokens under the \(a\)-notation.

The accompanying code \cite{ababot2025} follows this structure directly.  
It is organized around high-level \texttt{Token} objects, each providing built-in methods \texttt{Token.key()} and \texttt{Token.retrieve()}.  
Collections of tokens are grouped in \texttt{Pool} objects, which share the same interface but represent multi-token bases used by the retokenizer to learn correlations.  
A \texttt{Pool} thus acts as a composite token defining the active basis for local learning.

Consider the embedding from the previous section: $a_{\bf the\ cat\ jumps}$. It has three children: $a_{\bf the}$, $a_{\bf cat}$, \& $a_{\bf jumps}$. These children define the basis Pool used to construct it. Decoding the latent token requires keeping track of both the projectors 
\begin{eqnarray}
P_2&=&\tilde{a}^{(2)}_{\bf the\ cat}a_{\bf the}a_{\bf cat}
+ \tilde{a}^{(2)}_{\bf cat\ jumps}a_{\bf cat}a_{\bf jumps} \\
P_3&=&\tilde{a}^{(3)}_{\bf the\ cat\ jumps}\tilde{a}^{(2)}_{\bf the\ cat}a_{\bf jumps}
\end{eqnarray}
and the three child tokens themselves.

The code handles token tracking via two complementary storage modes: \textit{inline} and \textit{registry} builds. The latter exploits a global token registry to keep track of the parent-child relationships. This provides a lightweight global store, allowing a \texttt{Token} to carry only a reference (\texttt{TokenRef}) and resolve its parameters dynamically at retrieval.

Alternatively, the inline mode stores a token and its children directly together (as opposed to storing a pointer to the children, as with the registry). If the children themselves have children, such as with our example, then all descendants are stored together under the major token. 

The inline mode has the effect that no child token is shared by multiple parents, as each carries its own degenerate copy of its children. While this requires more disk space, only the part of the inline token necessary for calculation needs be stored into working memory. It is therefore not less memory efficient, and has the benefits of each token being independent from all other tokens.




\subsubsection{\bf Token modes (GI vs.\ Pnn).}
The code supports two Token modes for representing the projector structure: a (default) gauge-invariant \emph{GI} mode, and an ultra-compressed \emph{Pnn} mode.

The default mode treats a single set of projectors $\{P_n\}$---one per DAG depth $n$. It allows for exact retrieval \& left/right inference.

The \emph{Pnn} mode follows the SVD-based matrix product compression of Sec \ref{Sec:Compression}. In this mode, each DAG depth gets its own depth-specific projectors, such that Token memory is organized as 
\[
\texttt{Token.mv\_alpha}=\begin{bmatrix}
[m^{(1)}]\\
[P_{2,2}\!,\; m^{(2)}]\\
[P_{3,2}\!,\; P_{3,3}\!,\; m^{(3)}]\\[-3pt]
\vdots
\end{bmatrix}
\]
which do \emph{not} form a gauge-invariant set. The reasons for this are discussed in Sec \ref{Sec:Compression}. Note however, the compression is still exact, such that the complete pattern can still be retrieved (even without perfect tokenization). This makes it technically possible to convert between \emph{Pnn} \& \emph{GI} via replaying \emph{Pnn} back into an empty retokenizer.

In the current implementation, we instead choose to exploit a practical gauge hack: summing left/right gauges, $P\!\leftarrow\!P\oplus P_{\mathrm{Left}}$. This merges right \& left inference paths to a single terminal node, and converts right-token prediction into an ``outward-token" prediction, which generates either a left or right token. Entropy is used to estimate boundaries \& determine which side to merge the token.

\section{Trees \& Syntax}\label{Sec:Syntax}
In summary, human language patterns at all structural levels reflect a universal mechanism of subword formation, which arises in the limit of a sub-symbolic system constrained by locality.

\subsection{The Embedding Graphs are Syntax Trees}

Local learning naturally produces a tree topology of embeddings (Fig \ref{Fig:Trees}). Each retokenizer builds correlations from leaf to root, so the resulting parent–child graph of feature tokens forms a locally organized tree. The ordering of the leaves arises directly from learned correlations, not from any predefined syntax rule. This structure reflects how global order emerges sub-symbolically in a local system.

Because the embedding graph is tree-structured and explicit, its internal organization is interpretable. The same topology that emerges from locality mirrors the syntax trees used in symbolic theories of language \cite{Keenan&Moss2016book_MathematicalStructuresInLanguage}. This allows symbolic structure to be passed directly to the model: large classes of word embeddings can be tied by simple association to form composite trees. 

Such “soft-coded” initialization is effectively a type of pre-training, where we exploit the natural hierarchical organization to directly grow a symbolic theory as a topological enforced memory.

\subsection{Subregular Symmetry}

Note that hierarchy isn't handled via symbolic rules, but is an implicit feature of how the memory is organized. Hierarchical boundaries are statistical boundaries, and hence sub-symbolic. This is the same reason why there are no space tokens (see Sec \ref{Sec:Spaces}).

Because the retokenizer is local, it cannot distinguish a latent representation from a letter token. This makes it so that all ordered patterns, at any level, exhibit universal sub-word structure. This subregular symmetry mirrors statistical observations in human language \cite{Herdan1958_logNormal_intraword,Williams1940_logNormal_interwords} and recent formal work unifying syntax under subregular grammars \cite{Graf2022_SubregularLinguistics,GrafHanson2025_SubregularSyntax}.


\subsubsection{{\bf Memorization}: feature vs object tokens}

\begin{figure}[h!]
\centering
\begin{subfigure}[t]{0.5\textwidth}
    \centering
    \includegraphics[width=.95\linewidth]{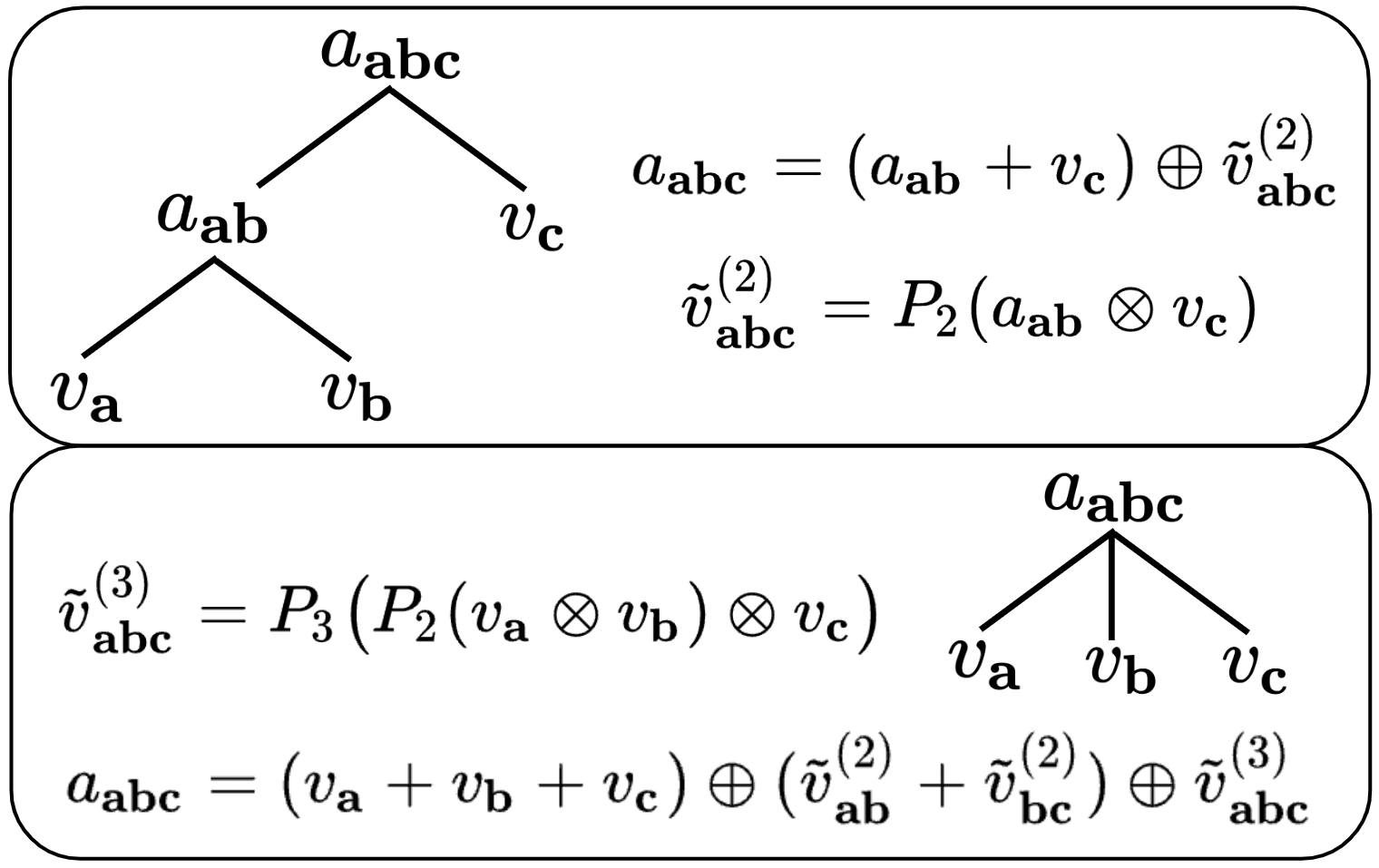}
    \caption{Two examples demonstrating how to formally map between embedding \& tree structure. There is an arbitrariness in representation arising from where one chooses to encode the boundary information.\\}
    \label{Fig:Trees_abc}
\end{subfigure}\\
\begin{subfigure}[t]{0.5\textwidth}
    \centering
    \includegraphics[width=1.\linewidth]{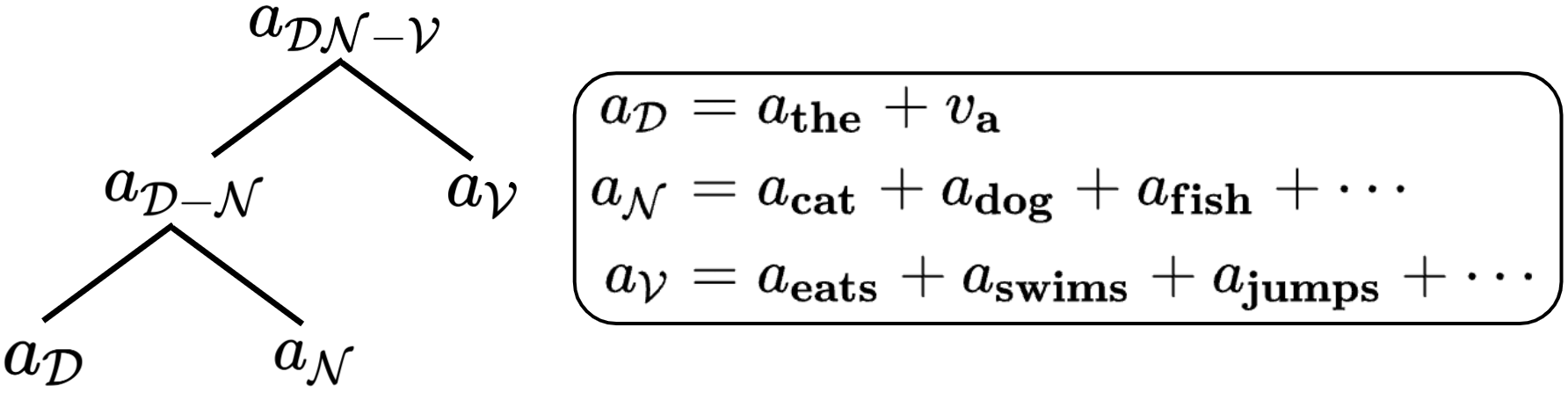}
    \caption{Categories can be formed by associating relevant tokens. Above shows a simple (determinant-noun)-verb construction.}
    \label{Fig:Trees_DNV}
\end{subfigure}
\caption{Local Tree Topology of Long-Term Memory}
\label{Fig:Trees}
\end{figure}

Note that there is an arbitrariness in representation arising from where one chooses to place the statistical boundaries (e.g Fig \ref{Fig:Trees_abc}). The pattern {\bf abc} could be learned as an ordered 3-point retokenized pattern (i.e intra hierarchy); or inter-hierarchically, ordered as a pair of retokenized 2-point memories.

In the Figure example shown, the minimal total number of neurons required for either choice are equal. However, for patterns larger than 3, it becomes more memory efficient to learn inter-hierarchically. This is because retokenized patterns require learning all sub-patterns as feature tokens.\footnote{This is also true on the computer: larger patterns means wider projectors. But any pattern can be stored inter-hierarchically, so that technically no tensor need have a dimension size larger than 2. This amounts to being able to trade width for depth (more but smaller tensors).}

This aligns with understood limitations of short-term memory \cite{Shiffrin1970_forgettingLongStrings} (the retokenizer being a compositional short-term memory structure). If the reader is asked to memorize {\bf alsjfgauy}, they might have an easier time by first breaking it up into {\bf als}, {\bf jfg}, \& {\bf auy}. This being a common tactic for remembering long numbers.

\subsection{Emergent Categories}

The same mechanism that builds meaningless subword structure also enables the self-organization of categories through replay.  
In \emph{key-match replay}, a random latent token is decoded from long-term into short-term memory.  
If the token corresponds to a word such as {\bf cat}, all embeddings that most closely overlap with it (e.g.\ {\bf the cat}, {\bf a cat runs}, {\bf cat jumped}) are replayed together.  
Their entangled feature maps can then be bound to a new embedding, forming a {\it bookmark} for future key searches and reducing the search space to similar contexts.  
These bookmarks act as new internal modalities—feature bases representing contextual associations—that can themselves be correlated and replayed to generate higher-order symbolic structure.  
We explore these emergent replay behaviors and their role in category formation in a follow-up work.

\section{Toys}\label{Sec:Toys}

We present a couple toy models to highlight model interpretability, as well as qualities special to it being a neuro-symbolic bridge.

\subsection{Latent Codes}

We start with the lower-case English alphabet as our basis, and grow embeddings $a_{\alpha}$ corresponding to an arbitrary list of words ({\bf cat},{\bf dog},{\bf fish},{\bf gator},...). We then associate all words into new class neurons (labeled {\bf A},{\bf B},{\bf C}), organized by word length, 
\begin{eqnarray}
a_{\bf A} &=& a_{\bf cat}\oplus a_{\bf dog}\oplus\cdots \\
a_{\bf B} &=& a_{\bf fish}\oplus a_{\bf cake}\oplus\cdots \notag\\
a_{\bf C} &=& a_{\bf gator}\oplus a_{\bf smith}\oplus\cdots \notag
\end{eqnarray}
Note that patterns in the latent space, e.g $[a_{\bf A},a_{\bf C},a_{\bf B}]$, behave like codes, in that the subsequent decoding of tokens in the list produces one possible valid string satisfying that order. 

Step-by-step, we activate $a_{\bf A}$, which activates all length-3 words. We then sample from that subset, generating (say) $a_{\bf cat}$, which decodes via the DAG structure of Eqn \ref{Eqn:a_cat} into precisely $[v_{\bf c},v_{\bf a},v_{\bf t}]$. Fully decoded gives us something like, ``{\bf cat smith cake}"---one of many possible 3-5-4 ordered strings.

We can store the latent code as the token 
\begin{eqnarray}
a_{\bf ACB} = (a_{\bf A}+a_{\bf B}+a_{\bf C})\oplus(\tilde{a}_{\bf AC}^{(2)}+\tilde{a}_{\bf CB}^{(2)})\oplus\tilde{a}_{\bf ACB}^{(3)}.
\end{eqnarray}
Even more complex patterns can be orchestrated as higher-order latent codes. \\


Note that we could instead have formed our vocabulary randomly, and simply formed random associations of the random vocabulary. A tree structure corresponding to multi-scale symbolic order naturally forms from these random processes. Such random structures are not content-less, as embeddings have meaning arising from their relative information. It has content, it's just random content.

\subsection{Quantity Tokens}

Consider a 1-letter alphabet ($d=1$). We'll write the basis token for this mode as $u(x)$, which is $1$ (or $0$) if the $x^{\text{th}}$ element in the string is the letter (or empty space). Its corresponding feature tokens $\tilde{u}^{(n)}$ measure quantity. They are formally defined from scalar generalizations of the projectors, which take the value $1$ or $0$ depending on if the trailing $n$ tokens aren't empty. Likewise, quantity object tokens can be defined 
\begin{eqnarray}
q_{n} = \frac{u\oplus\beta\tilde{u}^{(2)}\oplus\cdots\oplus\beta^{n-1}\tilde{u}^{(n)}}{\beta^{n-1}} , 
\end{eqnarray}
where $\beta>1$ biases toward longer length scales. This makes $q_n$ behave like $\tilde{u}^{(n)}$, but with a small residual overlap (vanishing in $\beta\rightarrow\infty$ limit) with all $\tilde{u}^{(n')}$ with $n'<n$.

Quantity tokens are a consequence of grounding symbolic behavior in sub-symbolic estimation. When $q_n$ \& $q_m$ are decoded in sequence, the resulting combined\footnote{Decoding is boundary aware, such that we can decode $[q_1,q_2]$ as $[u,0,u,u]$ or $[u,u,u]$, depending our choice to keep/throw away boundary information.} sequence keys $q_{n+m}$. Thus they provide the neural net an internal metric. Embeddings associated with $q_n$ inherent its quantity $n$.

\subsection{Feature Shadow \& Inter-Modality}

In addition to the $u$ tokens, we'll introduce the lower-case English alphabet ($d=26$), and associate every letter with $u$:
\begin{eqnarray}
a_{\bf a} &=& v_{\bf a}\oplus u \\
a_{\bf b} &=& v_{\bf b}\oplus u \notag\\
&\vdots&\notag\\
a_{\bf z} &=& v_{\bf z}\oplus u \notag
\end{eqnarray}
The new internal modalities are keyed by length $1$ context in the $u$ channel. Rather than treat them as separate channels, we can define the map 
\begin{eqnarray}
u(x) = \theta\Big(\sum_jv_j(x)-1\Big) , 
\end{eqnarray}
which is a step-function, acting on every element of the $v$ string at position $x$ in the context. Thus $u(x)$ lights up if any of the $26$ neurons of $v_j$ at site $x$ fire.

Now any higher-order token learned from these modes inherent a notion of length equal to the length of their stored pattern. For example, 
\begin{eqnarray}
a_{\bf aba} = (a_{\bf a}+a_{\bf b})\oplus(\tilde{a}^{(2)}_{\bf ab}+\tilde{a}^{(2)}_{\bf ba})\oplus\tilde{a}^{(3)}_{\bf aba}
\end{eqnarray}
is keyed by context $[u,u,u]$. As are all other length $3$ word embeddings in memory.

Now consider the following list of latent tokens: $[q_3,q_5,q_4]$. Fully decoded into $u$, it looks like 
$$[u,u,u,0,u,u,u,u,u,0,u,u,u,u].$$
The regions between the $0$'s project up into $\tilde{u}^{(n)}$, and therefore overlap with all tokens storing patterns of that length. This occurs even though $q_n$'s are defined independent of $a_{\alpha}$'s, and is a consequence of their shared projected behavior, not any direct correlation or shared association\footnote{Note that the feature tokens $\tilde{u}^{(n)}$ of both $q_n$ and $a_{\bf aba}$ aren't necessarily literally the same. Feature tokens $\{\tilde{u}^{(n)},\tilde{v}^{(n)},\tilde{a}^{(n)}\}$ are better understood as being transient placeholders representing how context is projected. The projectors don't have to be identical (due to representational freedom of dense matrices), so long as they behave equivalently. In the code, each token stores its own independent projectors.}.

This pattern of quantities can be stored as the token: 
\begin{eqnarray}
a_{3,5,4} = (q_3+q_4+q_5)\oplus(\tilde{q}_{3,5}^{(2)}+\tilde{q}_{5,4}^{(2)})\oplus\tilde{q}_{3,5,4}^{(3)}.
\end{eqnarray}

Note that because we have multiple external modes ($v_k$ \& $u$), there is a freedom in how to decode. Step-by-step, activating $a_{3,5,4}$ drives (say) $q_4$ to fire, which decodes out to its boundaries as $[0,u,u,u,u,0]$. If before decoding further in $u$-channel, we stop and form a list (a.k.a a Pool or temporary association) of all embeddings which overlap with that context, then randomly sampling from that list behaves like sampling from a list of length-4 words. More generally, by choosing a decoding order for the input modes ($u$ then $v_k$), we can decode governed by the inter-modal relationships. Then fully decoded $q_{3,5,4}$ looks like $a_{\bf ACB}$ from the previous subsection.

\section{Discussion}
\subsection*{Non-local \& Non-natural}

Large language models are not scientific theories of language, nor are they physical models of learning. They are non-local pattern recognizers designed to minimize a global objective. Though inspired by human learning, and even sharing some surface mechanisms such as attention, their mode of coordination is wholly artificial: each parameter receives an error signal computed from the entire system. This is equivalent to imagining every neuron equipped with a walkie-talkie, receiving global instructions\footnote{This is described as the {\it teaching signal} in neuroscience literature \cite{Schneidman2020_RandomProjections}. However, we stress that the fundamental problem is in the non-local processing of information, not the signaling---i.e the presence \& sensitivity of a global objective. {\it What in the brain has the degree of coordination enough to compute, using the non-local state of the brain, individual updates for every neuron?} (Nothing.)} on how to change its weights. No naturally occurring microscopic phenomenon exhibits such coordination.

These models do not explain language; they passively model data.  
This crucial fact is often ignored in the recent scientific literature \cite{Schrimpf&all2021,Krotov2023_AstrocyteTransformer,Hosseini&all2024,Caucheteux&all2022,Rathi&all2025_TopoLM,Lillicrap2020_BackpropBrain}.  
For example, studies have shown that the embeddings of pre-trained LLMs can be used to predict voxel-level activity in fMRI recordings \cite{Schrimpf&all2021,Caucheteux&all2022}.  
Yet this correlation is not evidence that the model intrinsically captures human language structure, or is a human language model.  
Language arises from microscopic organization in the brain, while fMRI voxels are only coarse averages of that same structure—so any predictive correlation is expected.  
The LLM plays a passive role in this correlation, reproducing statistical regularities already present in the data.  
In a universe without language data, such models would not produce language.  
They are no more a theory of language than a regression model that predicts planetary motion is a theory of gravity \cite{Vafa2025_PlanetaryLLM}.

Such models not only assume the existence of language, they cannot sustain it at any level (subword or otherwise).  
They lack the capacity for inter-generational learning—the process by which knowledge is transmitted and reinterpreted across time.  
When trained repeatedly on their own outputs, they undergo model collapse \cite{Shumailov&Gal2024nature_ModelCollapse}, a behavior that appears universal to any system trained by global fitting.  
This makes them evolutionarily sterile: they can imitate language but cannot participate in its continuation.  
Children, by contrast, exhibit robust inter-generational learning \cite{Kegl&Iwata1989_LSN,Senghas&Coppola2001_SignLanguage,Senghas1995_SignLanguage,reviewCreoles2018,Mcwhorter&Good2012_Saramaccan} despite orders of magnitude less data and computation, and yet could not compete with LLMs on modern industry benchmarks. 

No amount of model scaling can overcome this deficiency. It reflects a paradigmatic gulf between modern black box AI and the cognitive models of nature. Progress in understanding the latter requires starting over, and building models which reflect naturally occurring microscopic conditions, and not just looking for the model which best fits the data.

\subsubsection*{\bf The brain isn't a computer}
Computers, at a structural \& architectural level, are non-local \& symbolic. By contrast, the brain is a highly uncoordinated collection of structured accidents. Local memories lack the coordination necessary to perform calculations over large arrays. Hence why the model we present here decomposes into small tensors, despite realizing global order.

\subsection*{Thinking as an accidental mechanism of scaling}

Nature is a collection of accidents. Naturally occurring memories are simple traces of an event, such as a thumbprint in clay. They lack the coordination to do anything other than retrieve or forget. Consequently, retrieval (thinking) is essential to continued learning, because such memories can do little else without further coordination. This is not a statement about a specific microscopic mechanism, but a broad conclusion of locality. Our contribution here is to lay down the formalism for understanding how this incidental relearning occurs, and its (completely accidental) benefits: a parallel content-addressable long-term symbolic memory with non-static computational architecture. 

Unlike LLM's where ``thinking" is test-time compute for improving reasoning tasks, thinking in humans is a natural phenomenon without designed purpose or utility. While ``thinking" in LLM's provides some mechanisms for reinforcement learning \cite{Zhao&all2025_ZDR}, the training is still data-dependent \& non-local, and doesn't reflect the microscopic conditions of thinking in humans---therefore it is not really ``thinking" as we personally understand it. Such terminology, while a convenient analogy, is unscientific. LLM's don't learn by processing (and reprocessing) experiences; and not being event-driven, likely have no experience of any relatable kind.

Retrieval is therefore the natural source of continued learning.  
Improving a local model does not resemble optimization in global learners.  
There is no fine-tuning of stored memories, because each is topologically encoded and cannot be overwritten without destroying its structure.\footnote{Topological states are stable: the effect needs to be large enough to close the energy gap.}  
Without a global objective, nothing drives the system to collectively find internal representations; forcing back-propagation through local memories would destroy their plasticity and reduce the model to a static estimator.  

Scaling instead occurs through competition and accumulation: as new memories are added, they compete for activation and overlap during retrieval. As thinking generates new feature bases, new types of replay become possible---the evolving architecture means more ways in which noise can be injected to catalyze retrieval. Learning and relearning therefore unfold as a dynamic, multi-scale many-body problem. It is this physics problem that is the future of continued research in naturally occurring cognitive phenomena.

\section{Conclusion}    

We lay down the foundations for a universal description of human language as arising from naturally occurring microscopic conditions. This takes the form of a novel paradigm of AI, which realizes globally coordinated behavior as arising from highly uncoordinated learning. We find that the uncoordinated way in which coordinated behavior is learned leads to emergent multi-scale symbolic order (organized as locally-ordered trees) \& an evolving computational architecture. The local memory can store an infinite number of complex memories in a highly compressed way.

\section*{Code Availability}
\url{https://gitlab.com/emla-group/ababot}

\section*{Acknowledgements}

We thank Larry Moss, Ricky Simanjuntak, \& James Augeri for comments/discussion.


\bibliographystyle{ACM-Reference-Format}
\bibliography{biblio}

\clearpage\onecolumn
\appendix
\section{Universal Sub-Word Structure}

\subsection{Hierarchical Feature Mixing via Short-Term Memory}
\label{Sec:Suppl:HierFeatureMixing}

Here we employ the cut\&regrow method, discussed in the main text, in order to explore the hierarchical structure inspired from the features of 100 proper English or Spanish names. 

Growths are formed form local learning events following $g^{(n)}_{\mu_{n-1}, k} \gets (1-\xi_g)g^{(n)}_{\mu_{n-1}, k} + \xi_g\tilde{v}^{(n-1)}_{\mu_{n-1}} v_k$ with a cutoff $g_{\mu_{n-1},k}^{(n)}>\epsilon_n$. \\

\underline{Step 1:} We train a hierarchy against the 100 words, with the following hyperparameters: $\xi_g=.0001$ \& $\epsilon_n=0$ for all $n$. Doing this grows a hierarchy containing all the morphological features in the 100 training names. 

\underline{Step 2:} We choose $n_{\text{cut}}$ and discard layers $n\geq n_{\text{cut}}$. We then randomly regrow new layers by randomly growing the $g_{\mu_{n-1},k}^{(n)}$ (same as in previous section). This grows a new vocabulary, which we extract through random replay. (We choose $N_r=200$ replays, but more can be used to extract more words if desired.) 

The produced words exhibit a mixed featureset. The $n$-grams of the original 100 names form the nodes of a graph defining its smooth tokenset. New tokens, generated by cutting and regrowing the hierarchy layers, preserve the original notion of smoothness up until the cut ($n_{\text{cut}}-1$). This guarantees that new names are well-formed blends of both language morphologies. \\

In what follows, we show a sequence of random name regenerations for the 100 words, starting from $n_{\text{cut}}\gg n_{\text{peak}}$. Because randomly sampled $0\leq g^{(n)}_{\mu_{n-1},k}\leq 1$, then we need choose $0\leq \epsilon_n\leq 1$. We choose $\epsilon_{n_{\text{cut}}}>0$ when necessary to prevent the effective dimension from blowing up beyond our computational limitations; $\epsilon_n=.5$ for a chosen $n>n_{\text{cut}}$, in order to break up long words; and else $\epsilon_n=0$. \\ 

100 word data string:
{\bf gregory carl carolina esperanza declan lourdes jonás rosa kyle jacob \
maricarmen tanner brian sergio ivonne christian daniel weston fidel \
álvaro thomas lucía carmen norberto zane esther angela anthony ronald \
maría damien leo efrén preston leonardo matías craig gage bobby wayne \
trey liam antonio maxwell ricardo elías oscar ronaldo albert fernanda \
jesse andrés kip joe cintia roman águeda paco elisabet mateo ryker \
gary grant preston félix cody selene javier victoria walker malik \
ford sara ian steven russell teresa clara amanda luciana isaiah \
dawson elena sofía jaime blake celia tucker sean hernando milagros \
natalia dennis jack clinton dexter elena avery axel} \\

High cuts $n_{\text{cut}}\gg n_{\text{peak}}$ preserve the original morphology almost completely, thus regrowing those cuts tends to mostly reproduce the original vocabulary, and only a handful of new words.

$n_{\text{cut}}=6$:
{\bf milagr clintonio hernanda maricardo lucian milagro} \\

Lowering $n_{\text{cut}}$ reduces the constraints, allowing for more novel growths. 

$n_{\text{cut}}=5$:
{\bf jessell dexteresa amandrés oscarmen dexterestonio declara granthony \
presthe russelena javie westonio oscardo victoriana esperanthony \
celiam christiana ivonner álvarolintonio jessel espera esperantonio \
cintiana sergi milagrosa matía álvarolintiana granthomas pacob \
celisabet fernandrés maricarl presthernandrés presa granza alberto \
maricarolina presthernando leonaldo} \\

Turning on $\epsilon_{n}>0$ for $n>n_{\text{cut}}$ controls word length. \\

$n_{\text{cut}}=5$ \& $\epsilon_7=.5$ (to regulate word length):
{\bf fernandrés oscaroli romando agrosa lagros isabet varolina estoni \
vonner granto oscarm victoriana granza nthomas maliam perantoni \
peranza taliam omanda scarmen hernan amandré maricaroli icardo \
intonio exterest declara eranthony briana usselen stonio presther \
ricarme onardo pacob} \\

$n_{\text{cut}}=5$ \& $\epsilon_8=.5$ (to regulate word length):
{\bf icarmen maliam álvarol aricarl homandré jessell estonio alberto \
milagrosa lisabet xteresto lvarolinto briana hernandré ivonner \
homando amandrés eranthomas orberto celiam westhern ctoriana oscardo \
ernanda eonaldo jesselena declara presa pacob ronardo xteresa peranza \
lisaiah sselene} \\

$n_{\text{cut}}=5$ \& $\epsilon_9=.5$ (to regulate word length):
{\bf celisabet briana alberto presa milagrosa sthernando granthon cintonio \
ranthony ronardo celiam oscardo oscaroli elisaiah clintonio natalik \
peranthomas ctoriana romandrés matía arolinti aricardo westhern \
westonio leonaldo jesselena christiana álvaroli pacob thomanda \
presther exteresa ivonner romanda erantonio maricard sthernandrés \
maricarolina jesselene aricarmen nataliam restherna declara} \\

$n_{\text{cut}}=5$ \& $\epsilon_{10}=.5$ (to regulate word length):
{\bf maricarolina maliam briana westherna granza exterestonio alberto \
thernanda nthomanda victoriana celiam presa milagrosa granthomandrés \
lvarolina ivonner romanda maricardo declara natalik christiana \
leonaldo jesselena nataliam celisaiah perantoni peranthomas eranthony \
westonio oscardo maricarl celisabet arolintiana granthony carolinton \
amandrés álvarolinto romando russelena cintonio esthernandrés} \\

$n_{\text{cut}}=5$ \& $\epsilon_{11}=.5$ (to regulate word length):
{\bf clintiana oscardo natalik ivonner cintiana thomandrés fernandrés \
milagrosa hernandrés dexteresa presa granza ronardo cintonio nataliam \
presthernandr dexterestonio alberto romandrés rolintiana maricaroli \
esperanthony maricardo esperantonio álvarolinto christiana granthony \
russelena jesselena westonio jesselene victoriana pacob sthernando \
declara leonaldo speranthomas celisabet} \\

As $n_{\text{cut}}\rightarrow n_{\text{peak}}$, the new names begin to lose a recognizably English or Spanish structure. The longest correlations become increasingly replaced by randomness. By the time $n_{\text{cut}}=n_{\text{peak}}$, all but the most local correlations from original data are lost. Taking $n_{\text{cut}}=2$ is equivalent to generating a random language without starting data. \\

$n_{\text{cut}}=4$ \& $\epsilon_{4}=.7$ (to regulate exploding $d_n$):
{\bf kern luci noryk eonnisa inaliah honi delis exter iang evenang grey \
ilanert zannisa evene resto denang saranat inalis vony álvarom zanza \
pack élik thert sarl bricard lbery águe ianth eclarannisa forb ssern \
cell xele lías ofías xwesth gelis ronnie ndaw amanth águedam riamien \
mari prey danth eclake álvarol rtoryk zannie pern ianert ofía sserang \
soni lberey iando eonang narm xwel fide rgio dang delake desaiamil \
malke gagraig iell xelis ctonás bricarang sserannisa resp eclaranat} \\

$n_{\text{cut}}=4$ \& $\epsilon_{4}=.7$ (to regulate exploding $d_n$) \& $\epsilon_{5}=.5$ (to regulate word length):
{\bf ranal arme iang wsoná atía mand yleo tuci liah lene homa ictoná weleo \
hery perg bern zann gros onne mane cari oric saronn zang hony yner \
ilar essel axwe amarde tresti serg grom ianz denn gelik dest stonio \
omat clane joná edama feraim rony nthe sarona wald cland álvar ciandr \
amardo aric hris ofías maxe stoná eres xelí gela xteo mena rdell rant \
hric araim frén matev garí tery riam pert sabe into hera oryke stonn} \\

$n_{\text{cut}}=4$ \& $\epsilon_{4}=.7$ (to regulate exploding $d_n$) \& $\epsilon_{6}=.5$ (to regulate word length):
{\bf wesai iant axell herey xwesth caronanat bland caronat xwele heres \
caris teven clin homal carolin amarí selin ykery onant wayn jaco \
isary amat rber tonn iell vict hony varai luci ylenn eanda danda honá \
ilake ykere agelin javi keresp jonat jaim ntord jesteo veresp \
gronanat aliam ianio thera ierg oryk manio rdenarí jonis manz rosab \
nzanz nzanio ucía} \\

$n_{\text{cut}}=4$ \& $\epsilon_{4}=.7$ (to regulate exploding $d_n$) \& $\epsilon_{7}=.5$ (to regulate word length):
{\bf walb toná ynernata thom drés selik dene ucke omaraim seliann uedang \
feryk preg rbernata tana uedant estord prey ronniel javi omaxe \
lvaronni eona inang iamalke imen thoná aynert frés avergi scar heryk \
gard axwes aker spernatí clianni frén heraim iamas ngell ndaws axwe \
aniert tianzannielí amielí lenarme bria olint mant élis mando \
ranieraim agranza zannisa egord jonisa cranzannielí} \\

$n_{\text{cut}}=3$ \& $\epsilon_{3}=.9$ (to regulate exploding $d_n$):
{\bf íac xtrbelaguey jelaguec sprguec crabealk énnip áll áldaiabealk useg \
ixweny éseg sprger inzaguey ásahom iss kevo íaviv álí xtrbeax énniv \
ásahr spry catu orger grlv xtrt keolici spey xtru boeny ong der \
ixearí maw rnzaguec patu craiabealk fos íaviam dedo orguey ixeax \
crabelaguey rikici álk spex vedo tan deax áso íavip brm lelaguec \
ésealk xtrbldaikiclv keoliclv rnzarí stoeny craiam ucth malk fofí élv \
yniv íaviob leny gahr fré jelaguey deahom stody rne lelaguey xtrbldac \
vayll gahom ony}

\subsection{Learning Language Patterns in the Noise}\label{Sec:Suppl:FromNoise}

The following vocabulary was learned from noise:

{\bf ehqi nod lez pvp vpqn by idl khg adr uyfuq lp mfx nz zfb vhytjzf bg euq uqiq yfbp aiz eeeh dxiw mv ytd qiqfl rleu aid wxiqt xgeu eht onr oqg tdzfo mnr ofod ccai vpv vhytjz drl xix eze adxge adxgehg qg fleh wof zk iqp khy mwo ud} \\


The noise is the uniform string (shown below). We highlight those accidental correlations which trigger growths (given $\epsilon_{2}=.002$, $\epsilon_{n>2}=0$, \& $\xi_g=.001$):\\

v\allowbreak n\allowbreak c\allowbreak y\allowbreak h\allowbreak x\allowbreak k\allowbreak k\allowbreak d\allowbreak k\allowbreak f\allowbreak p\allowbreak g\allowbreak k\allowbreak {\bf wx}\allowbreak o\allowbreak u\allowbreak u\allowbreak l\allowbreak {\bf fb}\allowbreak v\allowbreak m\allowbreak b\allowbreak a\allowbreak c\allowbreak l\allowbreak {\bf kh}\allowbreak z\allowbreak r\allowbreak i\allowbreak l\allowbreak y\allowbreak {\bf iz}\allowbreak g\allowbreak w\allowbreak y\allowbreak g\allowbreak t\allowbreak p\allowbreak b\allowbreak b\allowbreak j\allowbreak s\allowbreak i\allowbreak j\allowbreak x\allowbreak {\bf ehqi}\allowbreak y\allowbreak {\bf lp}\allowbreak g\allowbreak u\allowbreak {\bf ge}\allowbreak g\allowbreak p\allowbreak {\bf onr}\allowbreak t\allowbreak m\allowbreak l\allowbreak {\bf nz}\allowbreak a\allowbreak b\allowbreak {\bf nr}\allowbreak {\bf mn}\allowbreak {\bf ee}\allowbreak q\allowbreak {\bf aiz}\allowbreak b\allowbreak o\allowbreak b\allowbreak s\allowbreak w\allowbreak s\allowbreak p\allowbreak e\allowbreak m\allowbreak a\allowbreak g\allowbreak n\allowbreak q\allowbreak k\allowbreak f\allowbreak {\bf vh}\allowbreak j\allowbreak h\allowbreak n\allowbreak f\allowbreak {\bf ix}\allowbreak {\bf lez}\allowbreak d\allowbreak n\allowbreak s\allowbreak r\allowbreak i\allowbreak t\allowbreak v\allowbreak f\allowbreak {\bf zf}\allowbreak a\allowbreak {\bf mnr}\allowbreak o\allowbreak u\allowbreak {\bf ud}\allowbreak d\allowbreak j\allowbreak m\allowbreak {\bf bg}\allowbreak {\bf hy}\allowbreak x\allowbreak a\allowbreak l\allowbreak w\allowbreak s\allowbreak b\allowbreak i\allowbreak j\allowbreak x\allowbreak d\allowbreak {\bf eeeh}\allowbreak b\allowbreak l\allowbreak {\bf iq}\allowbreak e\allowbreak y\allowbreak o\allowbreak s\allowbreak a\allowbreak {\bf mf}\allowbreak l\allowbreak j\allowbreak u\allowbreak h\allowbreak n\allowbreak i\allowbreak k\allowbreak t\allowbreak h\allowbreak j\allowbreak {\bf tj}\allowbreak t\allowbreak q\allowbreak o\allowbreak g\allowbreak w\allowbreak p\allowbreak g\allowbreak v\allowbreak k\allowbreak k\allowbreak c\allowbreak v\allowbreak x\allowbreak d\allowbreak p\allowbreak j\allowbreak d\allowbreak a\allowbreak f\allowbreak e\allowbreak k\allowbreak u\allowbreak {\bf fu}\allowbreak y\allowbreak o\allowbreak a\allowbreak n\allowbreak h\allowbreak a\allowbreak {\bf ca}\allowbreak d\allowbreak u\allowbreak p\allowbreak r\allowbreak s\allowbreak b\allowbreak j\allowbreak {\bf on}\allowbreak a\allowbreak o\allowbreak {\bf uy}\allowbreak t\allowbreak y\allowbreak q\allowbreak {\bf xg}\allowbreak b\allowbreak x\allowbreak {\bf fleh}\allowbreak {\bf ai}\allowbreak i\allowbreak l\allowbreak a\allowbreak v\allowbreak m\allowbreak o\allowbreak p\allowbreak n\allowbreak {\bf ai}\allowbreak k\allowbreak m\allowbreak {\bf od}\allowbreak {\bf iqt}\allowbreak c\allowbreak r\allowbreak {\bf drl}\allowbreak j\allowbreak h\allowbreak e\allowbreak k\allowbreak x\allowbreak e\allowbreak n\allowbreak w\allowbreak v\allowbreak g\allowbreak g\allowbreak j\allowbreak l\allowbreak r\allowbreak {\bf bg}\allowbreak n\allowbreak h\allowbreak c\allowbreak d\allowbreak p\allowbreak k\allowbreak {\bf zk}\allowbreak a\allowbreak l\allowbreak w\allowbreak c\allowbreak x\allowbreak r\allowbreak {\bf qi}\allowbreak {\bf of}\allowbreak a\allowbreak e\allowbreak r\allowbreak b\allowbreak d\allowbreak n\allowbreak {\bf tj}\allowbreak s\allowbreak l\allowbreak c\allowbreak {\bf nr}\allowbreak s\allowbreak j\allowbreak a\allowbreak {\bf bp}\allowbreak y\allowbreak n\allowbreak d\allowbreak c\allowbreak p\allowbreak w\allowbreak w\allowbreak i\allowbreak c\allowbreak r\allowbreak c\allowbreak x\allowbreak z\allowbreak x\allowbreak q\allowbreak b\allowbreak t\allowbreak p\allowbreak f\allowbreak {\bf iw}\allowbreak l\allowbreak q\allowbreak d\allowbreak g\allowbreak y\allowbreak n\allowbreak {\bf ix}\allowbreak p\allowbreak r\allowbreak c\allowbreak {\bf qt}\allowbreak x\allowbreak c\allowbreak e\allowbreak w\allowbreak m\allowbreak d\allowbreak {\bf ca}\allowbreak {\bf ud}\allowbreak d\allowbreak v\allowbreak z\allowbreak t\allowbreak s\allowbreak {\bf iw}\allowbreak l\allowbreak {\bf bg}\allowbreak {\bf uqiq}\allowbreak a\allowbreak e\allowbreak v\allowbreak {\bf yf}\allowbreak l\allowbreak d\allowbreak f\allowbreak k\allowbreak {\bf qn}\allowbreak k\allowbreak {\bf uy}\allowbreak b\allowbreak o\allowbreak l\allowbreak x\allowbreak w\allowbreak q\allowbreak b\allowbreak z\allowbreak p\allowbreak d\allowbreak {\bf aid}\allowbreak {\bf qg}\allowbreak m\allowbreak {\bf ad}\allowbreak f\allowbreak k\allowbreak y\allowbreak s\allowbreak t\allowbreak {\bf mn}\allowbreak s\allowbreak g\allowbreak x\allowbreak l\allowbreak {\bf no}\allowbreak w\allowbreak h\allowbreak d\allowbreak {\bf vp}\allowbreak k\allowbreak {\bf ca}\allowbreak {\bf fl}\allowbreak {\bf dz}\allowbreak {\bf ud}\allowbreak w\allowbreak u\allowbreak v\allowbreak {\bf wo}\allowbreak k\allowbreak {\bf ee}\allowbreak i\allowbreak f\allowbreak d\allowbreak y\allowbreak s\allowbreak y\allowbreak {\bf ix}\allowbreak c\allowbreak {\bf eht}\allowbreak z\allowbreak {\bf oq}\allowbreak f\allowbreak s\allowbreak {\bf tj}\allowbreak k\allowbreak {\bf bg}\allowbreak j\allowbreak j\allowbreak x\allowbreak m\allowbreak s\allowbreak {\bf zk}\allowbreak l\allowbreak h\allowbreak {\bf dr}\allowbreak h\allowbreak m\allowbreak x\allowbreak w\allowbreak q\allowbreak y\allowbreak p\allowbreak h\allowbreak o\allowbreak m\allowbreak {\bf od}\allowbreak r\allowbreak z\allowbreak {\bf jz}\allowbreak {\bf nz}\allowbreak {\bf xix}\allowbreak m\allowbreak g\allowbreak x\allowbreak k\allowbreak y\allowbreak a\allowbreak y\allowbreak q\allowbreak o\allowbreak p\allowbreak h\allowbreak x\allowbreak h\allowbreak z\allowbreak y\allowbreak n\allowbreak {\bf euq}\allowbreak c\allowbreak i\allowbreak {\bf pq}\allowbreak t\allowbreak o\allowbreak s\allowbreak r\allowbreak w\allowbreak w\allowbreak k\allowbreak o\allowbreak v\allowbreak q\allowbreak w\allowbreak e\allowbreak i\allowbreak m\allowbreak e\allowbreak x\allowbreak r\allowbreak o\allowbreak e\allowbreak {\bf adxgehg}\allowbreak k\allowbreak l\allowbreak {\bf bp}\allowbreak {\bf dz}\allowbreak r\allowbreak e\allowbreak v\allowbreak {\bf dz}\allowbreak x\allowbreak q\allowbreak h\allowbreak w\allowbreak l\allowbreak b\allowbreak {\bf fx}\allowbreak {\bf lp}\allowbreak o\allowbreak l\allowbreak c\allowbreak y\allowbreak {\bf mv}\allowbreak c\allowbreak f\allowbreak r\allowbreak z\allowbreak d\allowbreak {\bf wof}\allowbreak {\bf qt}\allowbreak c\allowbreak {\bf zk}\allowbreak s\allowbreak k\allowbreak r\allowbreak {\bf no}\allowbreak o\allowbreak {\bf yf}\allowbreak r\allowbreak p\allowbreak {\bf le}\allowbreak m\allowbreak k\allowbreak r\allowbreak {\bf cc}\allowbreak p\allowbreak {\bf pq}\allowbreak {\bf fx}\allowbreak a\allowbreak {\bf td}\allowbreak j\allowbreak v\allowbreak u\allowbreak g\allowbreak j\allowbreak {\bf lp}\allowbreak k\allowbreak q\allowbreak {\bf of}\allowbreak {\bf mw}\allowbreak j\allowbreak a\allowbreak l\allowbreak {\bf rl}\allowbreak d\allowbreak b\allowbreak {\bf tj}\allowbreak h\allowbreak s\allowbreak u\allowbreak a\allowbreak o\allowbreak j\allowbreak o\allowbreak {\bf iz}\allowbreak n\allowbreak b\allowbreak o\allowbreak r\allowbreak x\allowbreak m\allowbreak {\bf uq}\allowbreak s\allowbreak d\allowbreak k\allowbreak k\allowbreak x\allowbreak n\allowbreak j\allowbreak {\bf wxiqt}\allowbreak {\bf ee}\allowbreak j\allowbreak w\allowbreak y\allowbreak p\allowbreak a\allowbreak {\bf mv}\allowbreak e\allowbreak w\allowbreak {\bf khy}\allowbreak {\bf vp}\allowbreak u\allowbreak i\allowbreak {\bf tdzfo}\allowbreak t\allowbreak k\allowbreak e\allowbreak {\bf by}\allowbreak g\allowbreak z\allowbreak s\allowbreak s\allowbreak s\allowbreak c\allowbreak i\allowbreak f\allowbreak {\bf zk}\allowbreak {\bf iq}\allowbreak z\allowbreak q\allowbreak {\bf dx}\allowbreak v\allowbreak t\allowbreak {\bf iz}\allowbreak {\bf by}\allowbreak {\bf qiqfl}\allowbreak y\allowbreak {\bf ez}\allowbreak {\bf lp}\allowbreak c\allowbreak {\bf nz}\allowbreak a\allowbreak {\bf fo}\allowbreak z\allowbreak {\bf pvp}\allowbreak u\allowbreak a\allowbreak y\allowbreak u\allowbreak x\allowbreak s\allowbreak p\allowbreak g\allowbreak g\allowbreak v\allowbreak v\allowbreak t\allowbreak c\allowbreak {\bf fb}\allowbreak q\allowbreak s\allowbreak x\allowbreak a\allowbreak r\allowbreak o\allowbreak m\allowbreak {\bf zfb}\allowbreak {\bf hg}\allowbreak a\allowbreak e\allowbreak s\allowbreak t\allowbreak k\allowbreak j\allowbreak n\allowbreak y\allowbreak {\bf ca}\allowbreak u\allowbreak c\allowbreak m\allowbreak q\allowbreak j\allowbreak {\bf iw}\allowbreak {\bf ht}\allowbreak i\allowbreak {\bf vpqn}\allowbreak {\bf cc}\allowbreak k\allowbreak y\allowbreak {\bf rleu}\allowbreak {\bf hq}\allowbreak {\bf qp}\allowbreak x\allowbreak {\bf vpv}\allowbreak {\bf ze}\allowbreak j\allowbreak s\allowbreak l\allowbreak r\allowbreak t\allowbreak u\allowbreak n\allowbreak e\allowbreak x\allowbreak {\bf fo}\allowbreak n\allowbreak q\allowbreak a\allowbreak c\allowbreak {\bf mfx}\allowbreak x\allowbreak {\bf ccai}\allowbreak {\bf oqg}\allowbreak r\allowbreak r\allowbreak h\allowbreak h\allowbreak o\allowbreak s\allowbreak k\allowbreak w\allowbreak z\allowbreak {\bf dl}\allowbreak {\bf qg}\allowbreak {\bf nr}\allowbreak b\allowbreak m\allowbreak g\allowbreak {\bf ud}\allowbreak c\allowbreak {\bf ofod}\allowbreak o\allowbreak b\allowbreak {\bf idl}\allowbreak g\allowbreak i\allowbreak c\allowbreak j\allowbreak {\bf dxiw}\allowbreak v\allowbreak {\bf bp}\allowbreak b\allowbreak {\bf jz}\allowbreak l\allowbreak s\allowbreak {\bf ze}\allowbreak q\allowbreak u\allowbreak n\allowbreak j\allowbreak {\bf mf}\allowbreak {\bf ht}\allowbreak h\allowbreak s\allowbreak y\allowbreak {\bf vhytjzf}\allowbreak y\allowbreak w\allowbreak t\allowbreak f\allowbreak n\allowbreak t\allowbreak {\bf yfbp}\allowbreak e\allowbreak v\allowbreak i\allowbreak {\bf rl}\allowbreak f\allowbreak t\allowbreak p\allowbreak {\bf adr}\allowbreak i\allowbreak {\bf mn}\allowbreak p\allowbreak c\allowbreak {\bf ytd}\allowbreak h\allowbreak e\allowbreak m\allowbreak t\allowbreak {\bf qf}\allowbreak {\bf iqp}\allowbreak {\bf td}\allowbreak {\bf mv}\allowbreak k\allowbreak m\allowbreak {\bf hy}\allowbreak y\allowbreak {\bf by}\allowbreak m\allowbreak {\bf eze}\allowbreak n\allowbreak {\bf nz}\allowbreak w\allowbreak e\allowbreak {\bf of}\allowbreak a\allowbreak q\allowbreak {\bf khg}\allowbreak {\bf id}\allowbreak g\allowbreak s\allowbreak o\allowbreak t\allowbreak {\bf vh}\allowbreak p\allowbreak f\allowbreak {\bf qp}\allowbreak v\allowbreak y\allowbreak r\allowbreak j\allowbreak n\allowbreak {\bf nod}\allowbreak a\allowbreak h\allowbreak h\allowbreak {\bf fu}\allowbreak {\bf mw}\allowbreak x\allowbreak {\bf hy}\allowbreak c\allowbreak k\allowbreak {\bf dl}\allowbreak {\bf oq}\allowbreak n\allowbreak {\bf uyfuq}\allowbreak y\allowbreak z\allowbreak {\bf on}\allowbreak x\allowbreak {\bf xgeu}\allowbreak {\bf mwo}\allowbreak {\bf hqi}\allowbreak \\

Minimize energy by inserting spaces (i.e zero vectors). The lowest-energy tokenized state: \\

v n c y h x k k d k f p g k {\bf wx} o u u l {\bf fb} v m b a c l {\bf kh} z r i l y {\bf iz} g w y g t p b b j s i j x {\bf ehqi} y {\bf lp} g u {\bf ge} g p {\bf onr} t m l {\bf nz} a b {\bf nr} {\bf mn} {\bf ee} q {\bf aiz} b o b s w s p e m a g n q k f {\bf vh} j h n f {\bf ix} {\bf lez} d n s r i t v f {\bf zf} a {\bf mnr} o u {\bf ud} d j m {\bf bg} {\bf hy} x a l w s b i j x d {\bf eeeh} b l {\bf iq} e y o s a {\bf mf} l j u h n i k t h j {\bf tj} t q o g w p g v k k c v x d p j d a f e k u {\bf fu} y o a n h a {\bf ca} d u p r s b j {\bf on} a o {\bf uy} t y q {\bf xg} b x {\bf fleh} {\bf ai} i l a v m o p n {\bf ai} k m {\bf od} {\bf iqt} c r {\bf drl} j h e k x e n w v g g j l r {\bf bg} n h c d p k {\bf zk} a l w c x r {\bf qi} {\bf of} a e r b d n {\bf tj} s l c {\bf nr} s j a {\bf bp} y n d c p w w i c r c x z x q b t p f {\bf iw} l q d g y n {\bf ix} p r c {\bf qt} x c e w m d {\bf ca} {\bf ud} d v z t s {\bf iw} l {\bf bg} {\bf uqiq} a e v {\bf yf} l d f k {\bf qn} k {\bf uy} b o l x w q b z p d {\bf aid} {\bf qg} m {\bf ad} f k y s t {\bf mn} s g x l {\bf no} w h d {\bf vp} k {\bf ca} {\bf fl} {\bf dz} {\bf ud} w u v {\bf wo} k {\bf ee} i f d y s y {\bf ix} c {\bf eht} z {\bf oq} f s {\bf tj} k {\bf bg} j j x m s {\bf zk} l h {\bf dr} h m x w q y p h o m {\bf od} r z {\bf jz} {\bf nz} {\bf xix} m g x k y a y q o p h x h z y n {\bf euq} c i {\bf pq} t o s r w w k o v q w e i m e x r o e {\bf adxgehg} k l {\bf bp} {\bf dz} r e v {\bf dz} x q h w l b {\bf fx} {\bf lp} o l c y {\bf mv} c f r z d {\bf wof} {\bf qt} c {\bf zk} s k r {\bf no} o {\bf yf} r p {\bf le} m k r {\bf cc} p {\bf pq} {\bf fx} a {\bf td} j v u g j {\bf lp} k q {\bf of} {\bf mw} j a l {\bf rl} d b {\bf tj} h s u a o j o {\bf iz} n b o r x m {\bf uq} s d k k x n j {\bf wxiqt} {\bf ee} j w y p a {\bf mv} e w {\bf khy} {\bf vp} u i {\bf tdzfo} t k e {\bf by} g z s s s c i f {\bf zk} {\bf iq} z q {\bf dx} v t {\bf iz} {\bf by} {\bf qiqfl} y {\bf ez} {\bf lp} c {\bf nz} a {\bf fo} z {\bf pvp} u a y u x s p g g v v t c {\bf fb} q s x a r o m {\bf zfb} {\bf hg} a e s t k j n y {\bf ca} u c m q j {\bf iw} {\bf ht} i {\bf vpqn} {\bf cc} k y {\bf rleu} {\bf hq} {\bf qp} x {\bf vpv} {\bf ze} j s l r t u n e x {\bf fo} n q a c {\bf mfx} x {\bf ccai} {\bf oqg} r r h h o s k w z {\bf dl} {\bf qg} {\bf nr} b m g {\bf ud} c {\bf ofod} o b {\bf idl} g i c j {\bf dxiw} v {\bf bp} b {\bf jz} l s {\bf ze} q u n j {\bf mf} {\bf ht} h s y {\bf vhytjzf} y w t f n t {\bf yfbp} e v i {\bf rl} f t p {\bf adr} i {\bf mn} p c {\bf ytd} h e m t {\bf qf} {\bf iqp} {\bf td} {\bf mv} k m {\bf hy} y {\bf by} m {\bf eze} n {\bf nz} w e {\bf of} a q {\bf khg} {\bf id} g s o t {\bf vh} p f {\bf qp} v y r j n {\bf nod} a h h {\bf fu} {\bf mw} x {\bf hy} c k {\bf dl} {\bf oq} n {\bf uyfuq} y z {\bf on} x {\bf xgeu} {\bf mwo} {\bf hqi} \\

Thus we can see the hierarchical DAG structure living accidentally in the uniform string. For the hyperparameters given, bigrams are only learned if they show up at least twice in the string (because $\epsilon_2/\xi_g=2$). The word {\bf ehqi} is only learned because its bigrams show up at least one more time (beyond {\bf ehqi} itself) elsewhere in the string, such as how {\bf eh} appears in {\bf adxgehg}, {\bf fleh}, \& {\bf eeeh}. \\

Note that were we to instead have chosen $\epsilon_n=0$ for all $n$, the model would attempt to learn {\it all} correlations in the string. The model therefore attempts to learn the entire string as a smooth tokenset, which results in it only learning a single long word. 

By choosing $\epsilon_2>0$, we tell the model to {\it not} learn all observed bigrams. These now ``non-word-forming" bigrams are perceived as boundaries between smooth regions of the string, thus breaking up the string into smaller words. Further turning on $\epsilon_3>0$ would break up the string into yet smaller words, formed from smooth trigrams \& bigrams. As we subsequently fix each $\epsilon_n$ to a finite value, we'd find the words continue to break up until smaller chunks. This continues until $n$ larger than the largest chunks ($n=n_{\text{max}}$). At this point, the value of $\epsilon_{n>n_{\text{max}}}$ has no effect on learning, as there are no words in the string of that length. 

Different choices of the hyperparameters $\epsilon_n$ result in the string breaking up in different ways, with the strongest effect coming from learned $n$-grams at early $n$. We can therefore think of a smooth vocabulary as one possible tokenizable representation ``living" inside the uniform string. Different tokenizable representations are possible given the statistics of the $n$-grams and the hyperparameters. \\

\begin{figure}[h!]
\centering
\begin{subfigure}[t]{0.5\textwidth}
    \centering
    \includegraphics[width=.975\linewidth]{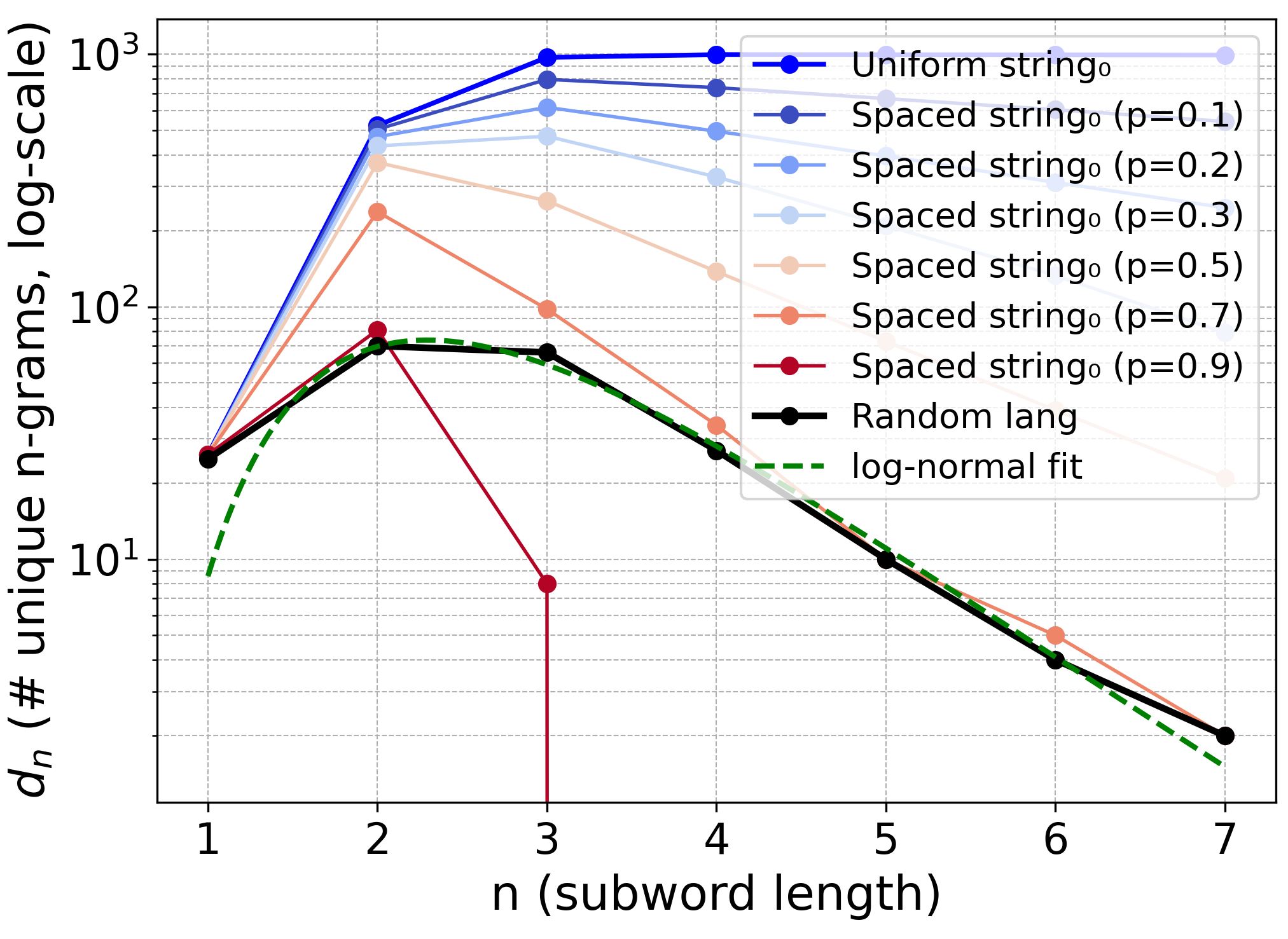}
    \label{Fig:noise_grams_logplot}
\end{subfigure}
\begin{subfigure}[t]{0.5\textwidth}
    \centering
    \includegraphics[width=.975\linewidth]{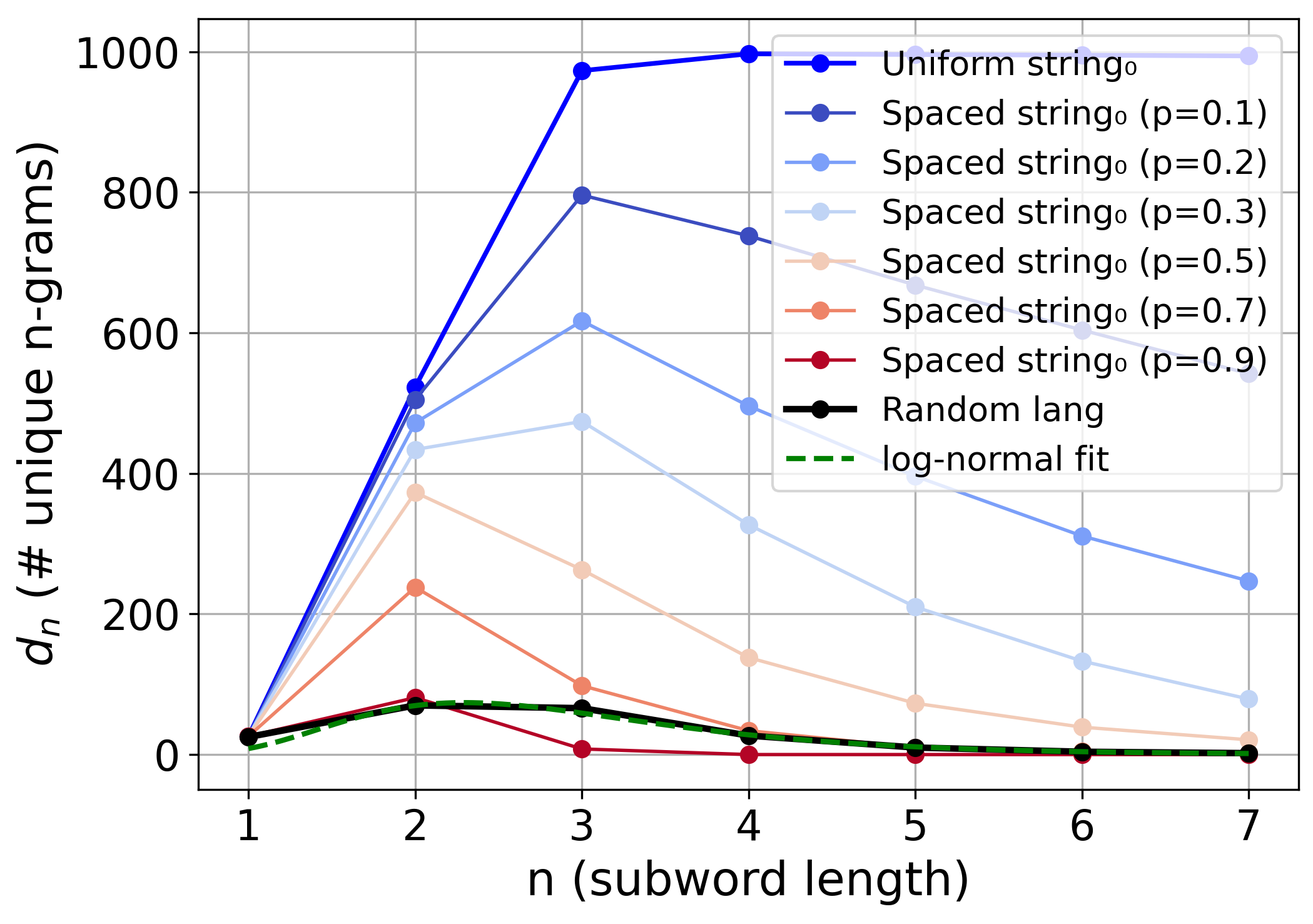}
    \label{Fig:noise_grams_plot}
\end{subfigure}
\caption{Emergent learning patterns in noise: Comparing $d_n$ between the seed string$_0$ \& the learned tokenizable representation living inside it (a.k.a random language). For comparison, seed$_0$ with spaces inserted at different frequencies. Even for the handful of tokenizable patterns in the string, the $d_n$ of random language begins to approach log-normal distribution, with a smooth transition across the peak. By comparison, the transition across the peak for random strings (w/ or w/o spaces) is sharp; and its not possible to predict $d_n>d_{n_{\text{peak}}}$ from $d_n<d_{n_{\text{peak}}}$, where $n_{\text{peak}}$ depends only on $d$ \& the sample size. Compare also Fig \ref{Fig:random_strings_vs_growths}.}
\label{Fig:noise_grams}
\end{figure}

Fig's \ref{Fig:noise_grams} show the $d_n$ distribution for the learned random language vocab \& the noise it was learned from. Additionally, we plot $d_n$ for the same noise string broken up by randomly inserting spaces. The hierarchical DAG structure of the random language exhibits a smooth transition across $n_{\text{peak}}$, in contrast to the sharp transitions exhibited by uniform chunks derived from the very same string -- highlighting their distinct structural difference.

The sub-word structure of human language is the same as that demonstrated here for the random language \cite{Eugenio2025_HebbianLanguage}, which is estimated to be a log-normal distribution \cite{Herdan1958_logNormal_intraword,Williams1940_logNormal_interwords,Eckhard&Werner&Markus2001_logNormal_review}. Our model provides a bottom-up explanation for the morphological distribution of language.

\subsection{Random Synaptic Growths}\label{Sec:Suppl:FromRandomGrowths}

Growing a random language follows equivalently to Sec \ref{Sec:Suppl:HierFeatureMixing}, except now where the entire graph is randomly learned (starting from $n=2$). We grow this random language with the following hyperparameters: $\epsilon_n=\big[.9,.55,.2,.2,.3,.3\big]$. \\

{\bf bjfyi bjfyimf cbjfa cbjfye cha chs cua cuvyea cuvyeg cuvyen cuvyy gpfha gpfhsl jfyy kr kv lhslhslh lyesl njfa njfjac njfyia njfyim njfyin omz ow oyea oyeg oyy pfhslha plha plyea qd qnjfj qpggw qpuv rea resl rzg sfa sfhslha soql sp spf spggw tin ug vdv vdx vdxj vyesl wg xmfa xmz xw yvha yvhslh yvyy zggw zgw zmz zn} \\ 

A uniform string (noise) of equal character length to random language {\it bjfyi} (for comparison): \\ 

{\bf l qvw tcprh t tsfcfz uu nitxe tgsjnqb enpg mroiatfin jnjth mtteluf fo v b tq phbk vikdvz jbhinz jkfeh nnhakwoo xnh w x hw mbs et fb xc fpg bs smnlpgx rt n g mb m k ib cxzholk granv dm oglbj rji agdfa dkew so p jwp krp w vgop qjrlzyl ggr tl o g k t n ovxrcwc ctrwq o j ix ljlja e lwblvs ymkm es f e v m km j cqifdhncovhzqg fz z ygyx dx} \\

\begin{figure}[h!]
\centering
\begin{subfigure}[t]{0.5\textwidth}
    \centering
    \includegraphics[width=1.\linewidth]{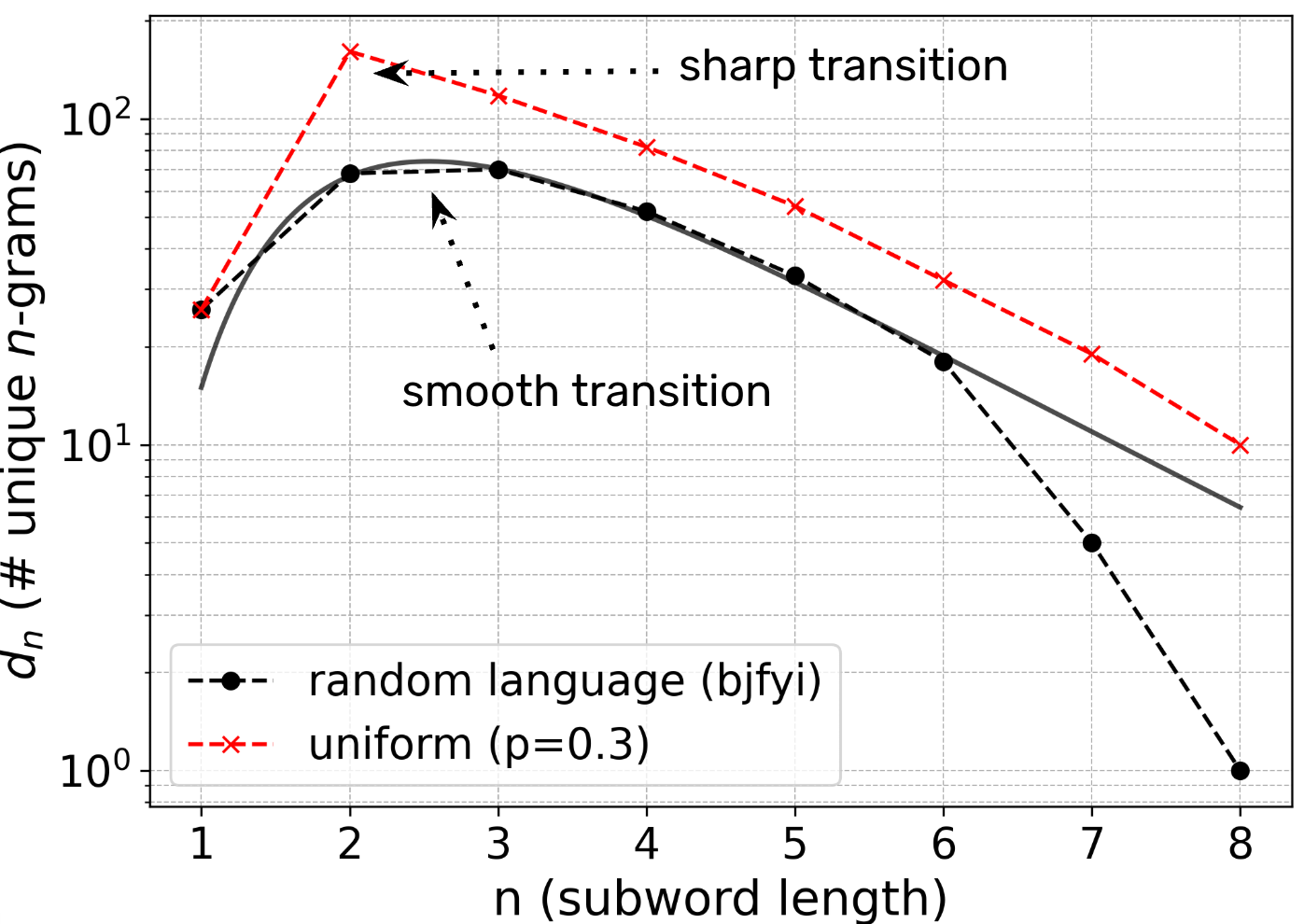}
    \label{Fig:random_strings_vs_growths_bjfyi_logplot}
\end{subfigure}
\begin{subfigure}[t]{0.5\textwidth}
    \centering
    \includegraphics[width=1.02\linewidth]{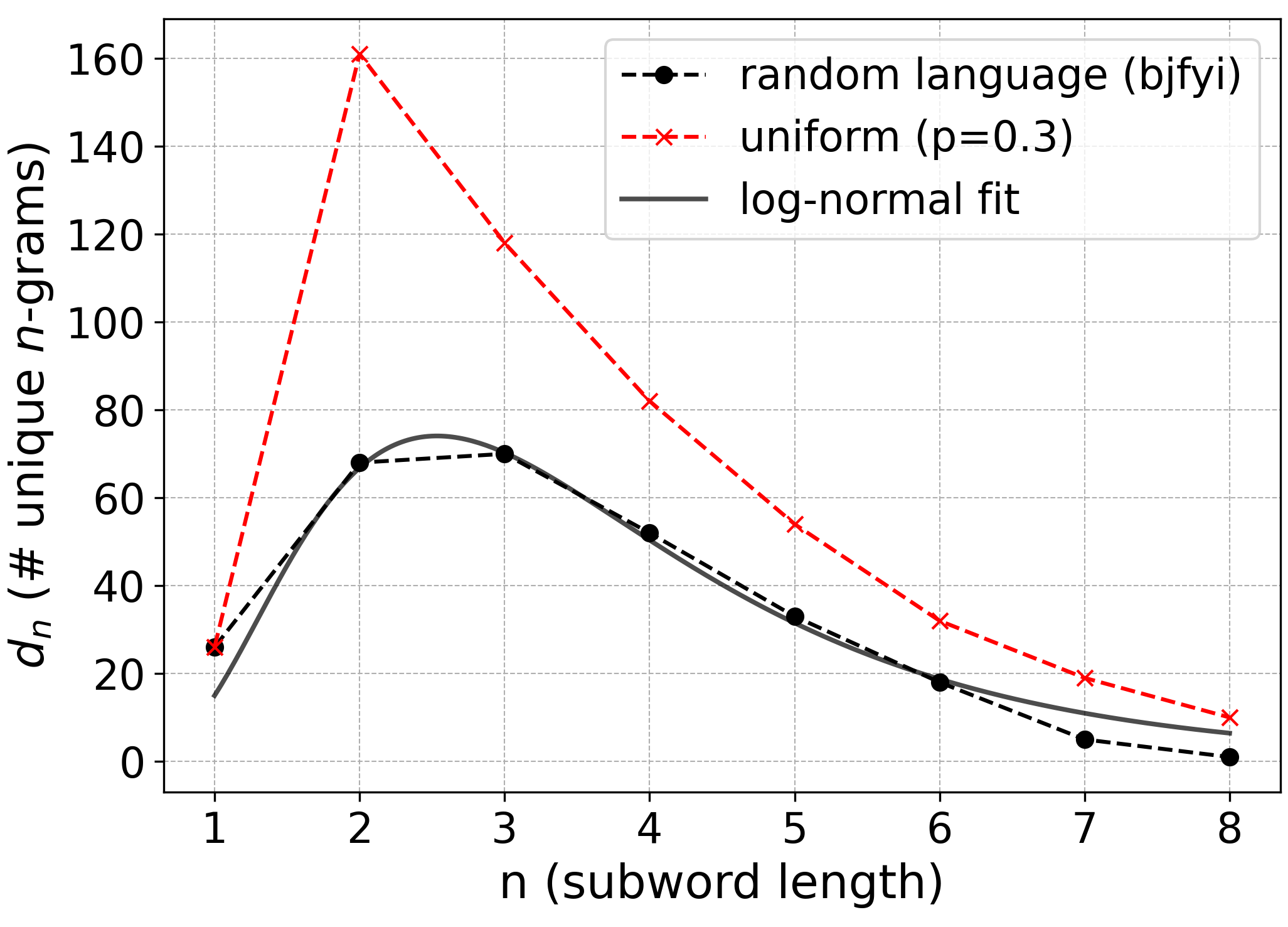}
    \label{Fig:random_strings_vs_growths_bjfyi}
\end{subfigure}
\caption{The distribution of word-forming patterns $d_n$ versus pattern length $n$ for: random language {\it bjfyi}, and a uniform string of equivalent number of characters (frequency $p=.3$ of inserting a space). The random language was generated by randomly growing the model layer-by-layer, with each subsequent layer constrained to be DAG-compatible with the previously grown layers (a.k.a the smoothness constraint). The $d_n$ for a hierarchical DAG has a smooth transition, because the DAG becomes increasingly more constraining to the growth with increasing $n$. The peak-and-collapse is a consequence of this constraint. By contrast, for uniform strings chunks, $n<n_{\text{peak}}$ \& $n>n_{\text{peak}}$ have no direct inter-relationship. The position of $n_{\text{peak}}$ for uniform strings is set entirely by $d$ \& the string size.}
\label{Fig:random_strings_vs_growths}
\end{figure}


\begin{figure}
    \centering
    \includegraphics[width=.5\linewidth]{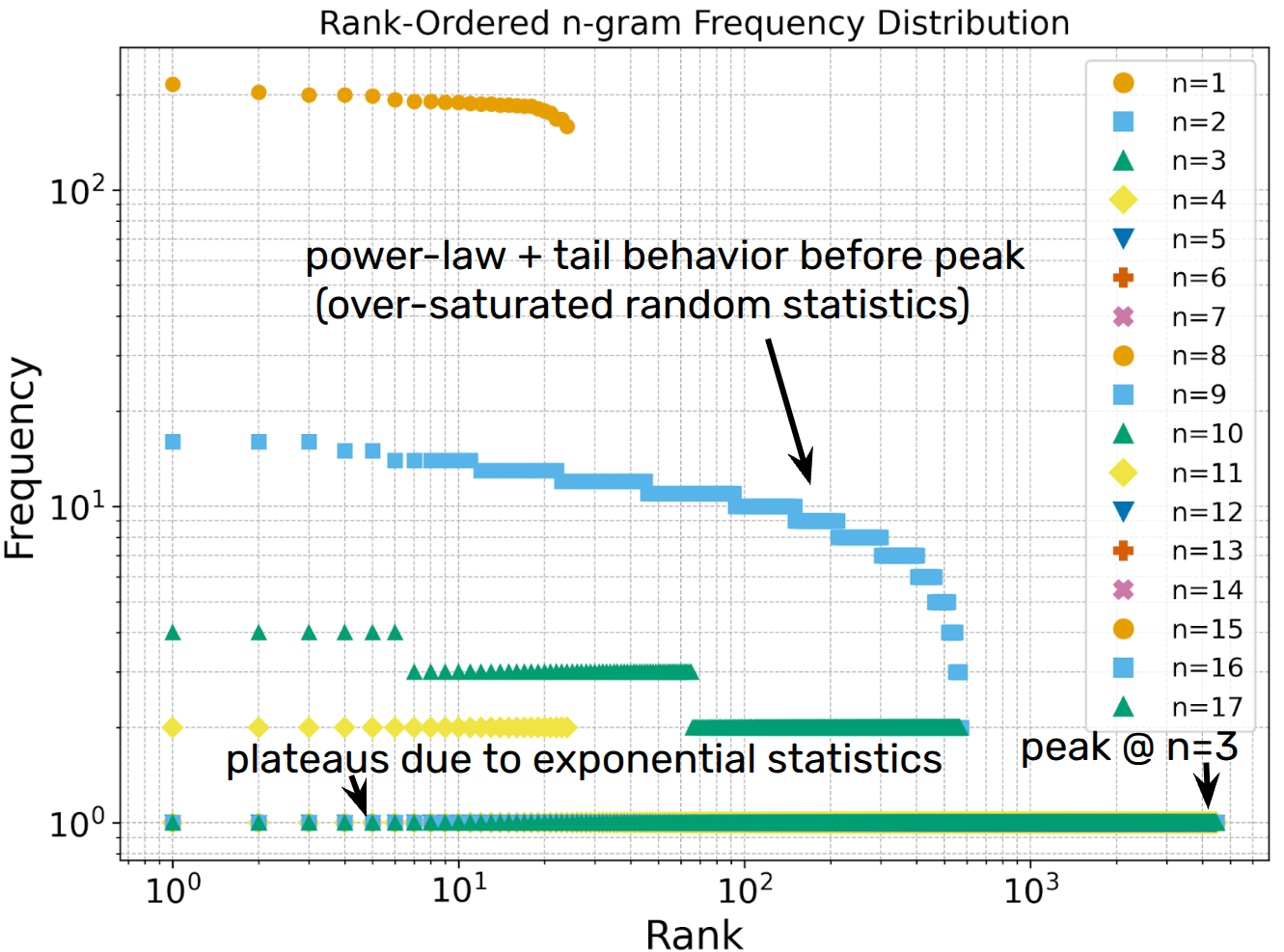}
    \caption{Spaceless string sampled from a uniform distribution over and alphabet. String length is the same as that in Fig \ref{Fig:greek_rosetta_stone}. The slope quickly crashes post peak for random strings.}
    \label{Fig:uniform}
\end{figure}

\begin{figure}[h!]
\centering
\begin{subfigure}[t]{0.508\textwidth}
    \centering
    \includegraphics[width=1.\linewidth]{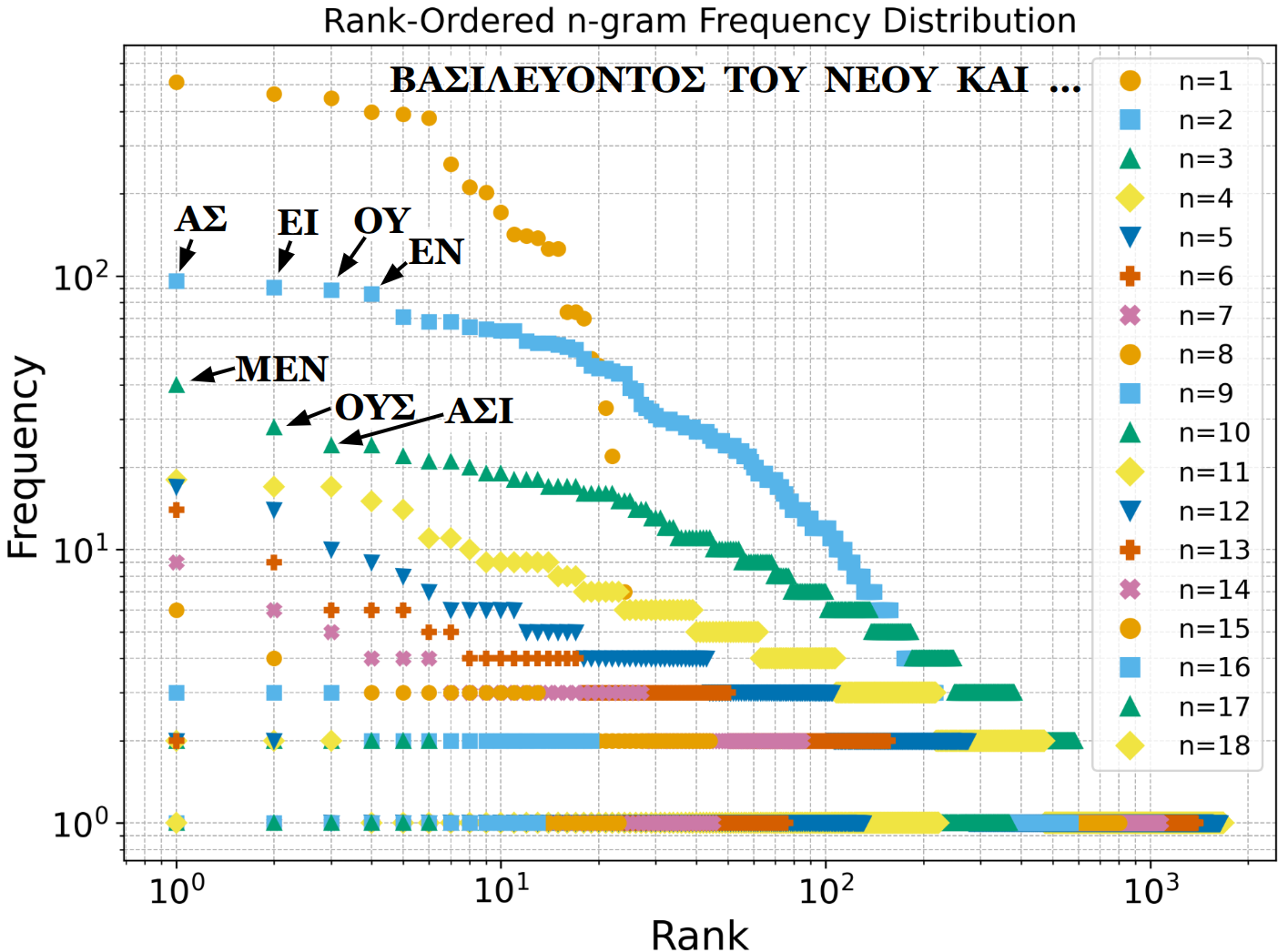}
    \label{Fig:greek_unique}
\end{subfigure}
\begin{subfigure}[t]{0.5\textwidth}
    \centering
    \includegraphics[width=1.02\linewidth]{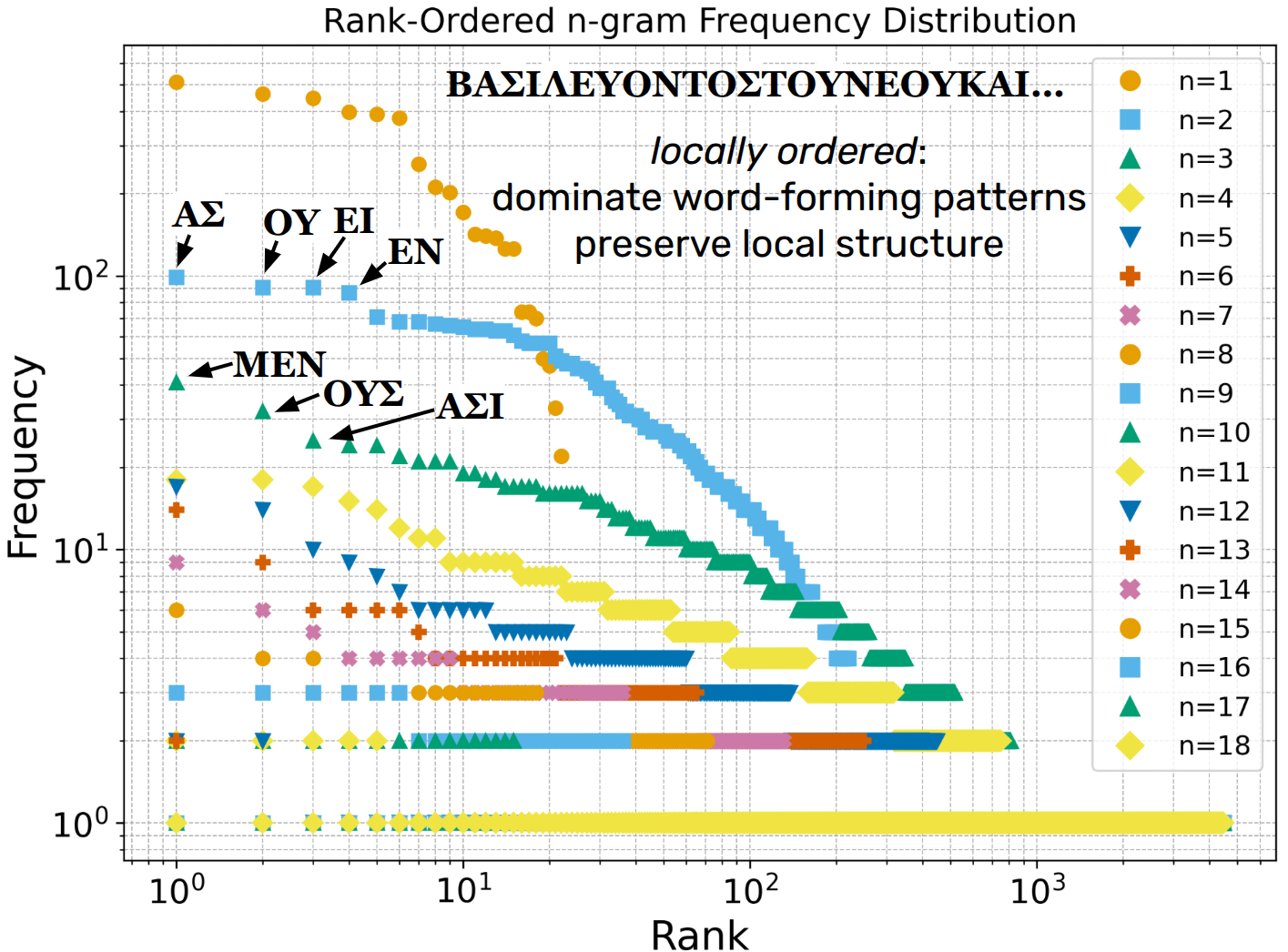}
    \label{Fig:greek_unique_spaceless}
\end{subfigure}
\caption{Greek text on the Rosetta Stone: with \& without spaces \cite{RosettaStoneOnline}.}
\label{Fig:greek_rosetta_stone}
\end{figure}

\section{Retokenizing Memory Grammars (for sub-word patterns)}

We define a hierarchy of neural grammar models built on memory-encoded directed acyclic graphs (DAGs) of overlapping $n$-grams. The three tiers roughly correspond to languages generated by states of memory with: fixed synapses (RTG), smoothly evolving synapses via memory erasure \& regrowth (CRG), and the addition of a global long-term memory (R-CRG).

\subsection{Tier 1: Retokenizing Grammar (RTG)}

A \textit{Retokenizing Grammar} represents a single memory DAG encoding a word through overlapping $n$-grams. It is defined as:
\[
\text{RTG} = (\Sigma, G)
\]
where:
\begin{itemize}
  \item $\Sigma$ is a finite alphabet;
  \item $G = (V, E)$ is a DAG with nodes $\tau \in V \subseteq \Sigma^n$ for fixed $n \geq 2$;
  \item There is an edge $(\tau_1, \tau_2) \in E$ iff $\mathrm{suffix}_{n-1}(\tau_1) = \mathrm{prefix}_{n-1}(\tau_2)$;
  \item The transition $\tau_1 \wedge \tau_2 \mapsto \tau_3$ merges $\tau_1$ and $\tau_2$ into a new string $\tau_3$ by overlapping the shared $(n{-}1)$-gram:
  \[
  \tau_3 = \texttt{merge}(\tau_1, \tau_2)
  \]
\end{itemize}

Each transition moves us deeper into $G$. A complete path through $G$ corresponds to a sequence of such merges ending a terminating node $w$ (the stored word). The language of an RTG is: $L(\text{RTG}) = \left\{ w \in \Sigma^* \right\}$, where $\Sigma^*\subset V$. Note that nodes like {\bf aaa} can be formed via self-mergers \texttt{merge}({\bf aa},{\bf aa}). A terminating node formed from self-mergers can correspond to an infinitely long word. 

\subsubsection*{Token Representations and Retokenization Classes}

Each string $w \in L(\text{RTG})$ corresponds to a terminating node in the DAG and defines a unique symbolic unit. However, its internal structure may admit multiple valid tokenizations via overlapping $n$-grams. We associate to each $w$ a set of smooth tensor representations that describe how it may be composed from lower-order fragments.

Let $\tilde{v}^{(n)}_w$ denote the composed representation of $w$ (an $n$-gram at level $n$), and let $\mathcal{R}(w)$ denote its retokenization class:
\[
\mathcal{R}(w) \subset \mathcal{T} = \text{all valid retokenized expressions over sub-tokens}
\]

\noindent For example, for $w = \textbf{aba}$, we may have:
\begin{eqnarray}
\mathcal{R}(\textbf{aba}) = \left\{
\tilde{v}^{(3)}_{\textbf{aba}},\quad
\tilde{v}^{(2)}_{\textbf{ab}} v_{\textbf{a}},\quad
v_{\textbf{a}} \tilde{v}^{(2)}_{\textbf{ba}},\quad
v_{\textbf{a}} v_{\textbf{b}} v_{\textbf{a}}
\right\}
\end{eqnarray}

Each expression describes a valid smooth composition of $\tilde{v}^{(3)}_{\textbf{aba}}$ based on the available overlaps. These are not distinct outputs of the grammar, but rather different structural decompositions of the same element $w \in L$.

\subsection{Tier 2: Cut-and-Regrow Grammar (CRG)}

A \textit{Cut-and-Regrow Grammar} generalizes the RTG by allowing the underlying memory DAG to evolve through a process of structural erasure and local recomposition. Rather than a fixed instance of memory, a CRG defines a family of potential memory reconstructions constrained by anchor overlap rules.

We define a CRG as a transformation:
\[
\text{CRG}: \texttt{DAG}_{\text{initial}} \rightarrow \texttt{DAG}_{\text{final}}
\]
where:
\begin{itemize}
  \item $\texttt{DAG}_{\text{initial}} = (V_{< n_c}, E_{< n_c})$ is a truncated memory graph containing only $n$-grams of length $n < n_c$ (the cut threshold);
  \item New $n$-grams ($n \geq n_c$) may be regrown into $\texttt{DAG}_{\text{final}}$ by composing overlapping $(n{-}1)$-grams from $\texttt{DAG}_{\text{initial}}$;
  \item A novel $n$-gram $\tau$ is valid iff it is the result of a merge or self-merge from overlapping $(n{-}1)$-grams already present in the graph.
\end{itemize}

Growth proceeds locally and recursively. At each step, a new DAG is created by merging compatible anchors. This process may be terminated arbitrarily or allowed to run until structural collapse (no further growth) or infinite replay (e.g through cyclic self-mergers).

\medskip
\noindent We denote:
\[
\mathcal{G}_{\text{cut}} = \text{all DAGs reachable from } \texttt{DAG}_{\text{initial}} \text{ via valid regrowth}
\]
\[
\texttt{DAG}_{\text{final}} \in \mathcal{G}_{\text{cut}}
\]

As in the RTG, the language $L(\text{CRG})$ consists of terminating nodes (strings) in $\texttt{DAG}_{\text{final}}$, and each string $w \in L$ may admit a corresponding retokenization class $\mathcal{R}(w)$ derived from its construction history.

\subsection{Tier 3: Replay-Enriched Retokenizing Grammar (R-CRG)}

A \textit{Replay-Enriched Retokenizing Grammar} extends the CRG by introducing a global long-term memory of stored word structures. Let:
\[
\mathcal{M}_{\text{initial}} = \left\{ \texttt{DAG}_w \mid w \in \Sigma^* \right\}
\]
be a collection of disentangled RTGs, each encoding a stored word $w$.

At runtime, a subset of these memory DAGs may be reloaded into the retokenizer. The result is an entangled memory structure: 
\[
\texttt{DAG}_{\text{reloaded}} = \bigcup_{w \in \mathcal{M}_{\text{load}}} \texttt{DAG}_w
\]
This composite structure defines a new RTG with potentially mixed anchors. Applying CRG-style cut-and-regrow operations to this DAG yields a modified instance of memory: $\texttt{DAG}_{\text{reloaded}} \rightarrow \texttt{DAG}_{\text{final}}$. We define:
\[
\mathcal{M}_{\text{final}} = \mathcal{M}_{\text{initial}} \cup \text{dis}\big(\left\{\texttt{DAG}_{\text{final}}\right\}\big)
\]
where $\text{dis}\big(\left\{\texttt{DAG}_{\text{final}}\right\}\big)$ are the disentangled (via replay relearning) memory states stored in $\left\{\texttt{DAG}_{\text{final}}\right\}$.

The R-CRG may thus be understood as the system defined by all possible reachable global memory states. The language $L(\text{R-CRG})$ consists of terminating strings in the active memory, or equivalently, all strings recognized by (and replayable via) its key-value memory.

\begin{algorithm}[ht]
\caption{Hierarchical Retokenizer Training Protocol}
\SetAlgoLined

\KwIn{Input sequence $v$, basis size $d=26$, parameters $\{\xi_g, \epsilon_n\}$}
\KwOut{Hierarchical token memory $\{\tilde{v}^{(n)}\}$ and projectors $\{P_n\}$}

Initialize one-hot vectors $v_k$ for each symbol $k$ in alphabet\;
Set context $\tilde{v}^{(1)} \gets v$ (level-1 features)\;

\For{$n = 2$ \KwTo max-depth}{
  Initialize $g^{(n)}_{\mu_{n-1}, k} \gets 0$ for all $\mu_{n-1}, k$\;

  \tcc{Hebbian update from input events}
  \ForEach{training event $(\tilde{v}^{(n-1)}_{\mu_{n-1}}, v_k)$ in data}{
    $g^{(n)}_{\mu_{n-1}, k} \gets (1 - \xi_g)g^{(n)}_{\mu_{n-1}, k} + \xi_g \cdot \tilde{v}^{(n-1)}_{\mu_{n-1}} v_k$\;
  }

  \tcc{Apply DAG smoothness constraint}
  Zero-out elements of $g^{(n)}$ violating DAG compatibility\;

  \tcc{Define merged token vectors}
  \ForEach{entry $g^{(n)}_{\mu_{n-1}, k}$}{
    \If{$g^{(n)}_{\mu_{n-1}, k} > \epsilon_n$}{
      Define new token $\tilde{v}^{(n)}_{\mu_n}$ from $(\mu_{n-1}, k)$\;
      Define projector $P^{\mu_n,\mu_{n-1},k}_n \gets 1$\;
    } \Else{
      $P^{\mu_n,\mu_{n-1},k}_n \gets 0$\;
    }
  }

  Use $\tilde{v}^{(n)}$ as input to next level\;
}

\Return $\{\tilde{v}^{(n)}\}, \{P_n\}$\;
\end{algorithm}

\begin{algorithm}[ht]
\caption{Random Growth Protocol for Hierarchical Retokenizer}
\SetAlgoLined

\KwIn{Alphabet size $d = 26$, depth $N$, thresholds $\{\epsilon_n\}$}
\KwOut{Hierarchical memory $\{g^{(n)}\}$ and token projectors $\{P_n\}$}

Initialize level-1 basis $v_k$: one-hot vectors for each symbol $k$\;
Set $\tilde{v}^{(1)} \gets v$\;

\For{$n = 2$ \KwTo $N$}{
  \tcc{Randomly initialize correlation matrix}
  Sample $g^{(n)}_{\mu_{n-1}, k} \sim \mathcal{U}[0,1]$ i.i.d.\;
  
  \tcc{Apply DAG mask to enforce smoothness constraint}
  Zero-out elements of $g^{(n)}$ violating hierarchical DAG constraints\;

  \tcc{Construct merged token vectors}
  \ForEach{$(\mu_{n-1}, k)$}{
    \If{$g^{(n)}_{\mu_{n-1}, k} > \epsilon_n$}{
      Define new token $\tilde{v}^{(n)}_{\mu_n}$ from $(\mu_{n-1}, k)$\;
      Set projector $P^{\mu_n,\mu_{n-1},k}_n \gets 1$\;
    } \Else{
      $P^{\mu_n,\mu_{n-1},k}_n \gets 0$\;
    }
  }

  Use $\tilde{v}^{(n)}$ to define basis for next level\;
}

\Return $\{g^{(n)}\}, \{\tilde{v}^{(n)}\}, \{P_n\}$\;
\end{algorithm}






  




\end{document}